\let\mypdfximage\pdfximage
\def\pdfximage{\immediate\mypdfximage}

\documentclass[acmtog,screen]{acmart}
\acmSubmissionID{166}

\acmJournal{TOG}
\acmVolume{39}
\acmNumber{6}
\acmArticle{254}
\acmYear{2020}
\acmMonth{12}

\setcopyright{rightsretained}

\acmDOI{10.1145/3414685.3417779}

\usepackage{graphicx,adjustbox}
\usepackage{bm}       
\usepackage{multirow}
\usepackage{xspace}
\usepackage{animate}

\newcommand{\revision}[1]{#1}

\newcommand{\latent}{\bm{u}}
\newcommand{\latentopt}{\bm{u^{*}}}
\newcommand{\params}{\bm{\theta}}
\newcommand{\paramsopt}{\bm{\theta^{*}}}
\newcommand{\img}{\bm{I}}
\newcommand{\albedo}{\bm{a}}
\newcommand{\normal}{\bm{n}}
\newcommand{\rough}{r}
\newcommand{\spec}{\bm{s}}
\newcommand{\loss}{\mathcal{L}}
\newcommand{\lossPix}{\loss_\mathrm{pixel}}
\newcommand{\lossPercp}{\loss_\mathrm{percept}}
\newcommand{\render}{\mathcal{R}}

\newcommand{\generator}{\mathcal{G}}

\newcommand{\light}{L}
\newcommand{\camera}{C}
\newcommand{\Real}{\mathbb{R}}

\newcommand{\bw}{\bm{w}}

\newcommand{\bz}{\bm{z}}

\newcommand{\noise}{\bm{\xi}}
\newcommand{\calW}{\mathcal{W}}
\newcommand{\calZ}{\mathcal{Z}}
\newcommand{\calN}{\mathcal{N}}
\newcommand{\calWN}{\mathcal{W}^+\mathcal{N}}

\newcommand{\commentout}[1]{}

\DeclareMathOperator*{\argmin}{arg\,min}

\newlength{\resLen}
\newlength{\raiseLen}

\newcommand{\one}{}
\newcommand{\two}{}
\newcommand{\thr}{}
\newcommand{\fou}{}
\newcommand{\fiv}{}
\newcommand{\six}{}
\newcommand{\sev}{}
\newcommand{\eit}{}
\newcommand{\nin}{}
\newcommand{\ten}{}

\newcommand{\totReal}{28\xspace}
\newcommand{\totSynthetic}{39\xspace}

\graphicspath{{images/}}

\title{MaterialGAN: Reflectance Capture using a Generative SVBRDF Model}

\author{Yu Guo}
\affiliation{\institution{University of California, Irvine}}
\author{Cameron Smith}
\affiliation{\institution{Adobe Research}}
\author{Milo\v{s} Ha\v{s}an}
\affiliation{\institution{Adobe Research}}
\author{Kalyan Sunkavalli}
\affiliation{\institution{Adobe Research}}
\author{Shuang Zhao}
\affiliation{\institution{University of California, Irvine}}

\begin{document}
\setlength{\resLen}{6.9in}
\begin{teaserfigure}
	\centering
	\addtolength{\tabcolsep}{-4pt}
	\begin{tabular}{cc}
		\raisebox{0.225in}{\rotatebox[origin=c]{90}{\small Input}} &
		\includegraphics[width=\resLen]{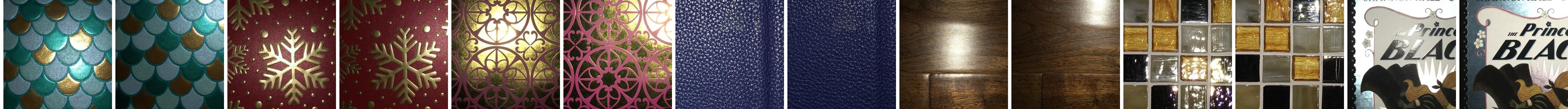}
		\\[-2pt]
		\raisebox{0.45in}{\rotatebox[origin=c]{90}{\small Rendering}} &
		\includegraphics[width=\resLen]{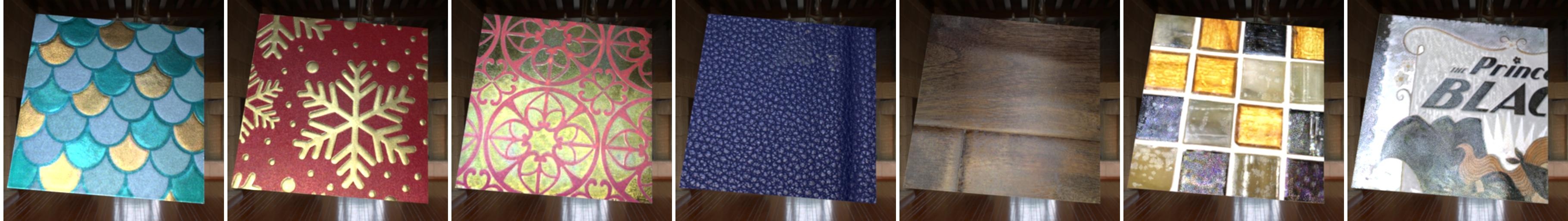}
		\\[-2pt]
		\raisebox{0.45in}{\rotatebox[origin=c]{90}{\small Estimated maps}} &
		\includegraphics[width=\resLen]{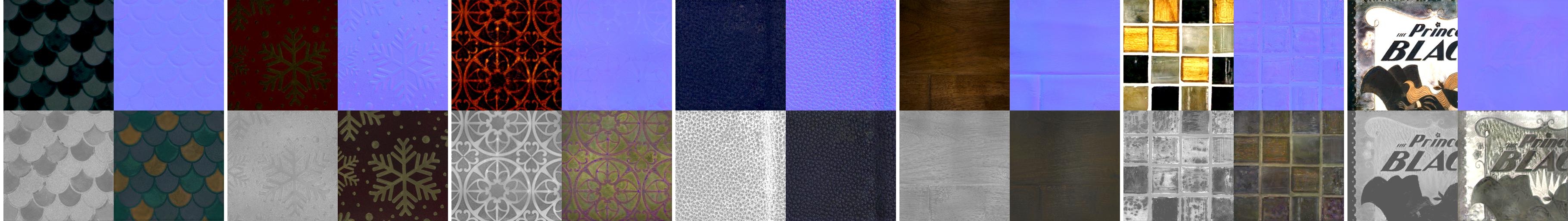}
	\end{tabular}
	\caption{\label{fig:teaser}
		We introduce a method to capture SVBRDF material maps from a small number of mobile flash photographs, achieving high quality results both on original and novel views. Our key innovation is optimization in the latent space of MaterialGAN, a generative model trained to produce plausible material maps; MaterialGAN thus serves as a powerful implicit prior for result realism. Here we show re-rendered views for several different materials under environment illumination. We use 7 inputs for these results (with 2 of them shown).
		(Please use Adobe Acrobat and click the renderings to see them animated.)
	}
\end{teaserfigure}

    \begin{abstract}
    We address the problem of reconstructing spatially-varying BRDFs from a small set of image measurements.
    This is a fundamentally under-constrained problem, and previous work has relied on using various regularization priors or on capturing many images to produce plausible results.
    In this work, we present \emph{MaterialGAN}, a deep generative convolutional network based on StyleGAN2, trained to synthesize realistic SVBRDF parameter maps.
    We show that MaterialGAN can be used as a powerful material prior in an inverse rendering framework: we optimize in its latent representation to generate material maps that match the appearance of the captured images when rendered.
    We demonstrate this framework on the task of reconstructing SVBRDFs from images captured under flash illumination using a hand-held mobile phone. Our method succeeds in producing plausible material maps that accurately reproduce the target images, and outperforms previous state-of-the-art material capture methods in evaluations on both synthetic and real data.
    Furthermore, our GAN-based latent space allows for high-level semantic material editing operations such as generating material variations and material morphing.

\end{abstract}
    %
    \maketitle
    \section{Introduction}
\label{sec:intro}
Despite a few decades of effort in computer graphics and vision, capturing spatially-varying reflectance of real-world materials remains a challenging and actively researched task.
Measurement methods have traditionally used custom hardware systems to densely sample illumination and viewing directions \cite{Marschner1999,Matusik2003}, followed by post-processing such as fitting parametric BRDF models \cite{Ngan2005}. However, such approaches are restricted to laboratory conditions.

Recent work has explored methods for casual capture of spatially-varying BRDFs (SVBRDFs) using commodity hardware and in less constrained environments \cite{Francken2009,Ren2011,Aittala2013,Aittala2015,Hui2017}.
These methods usually follow an \emph{inverse-rendering} approach: they define a forward rendering model and optimize reflectance parameters so that the simulated appearance matches physical measurements under certain image metrics.
With a small number of measured images, this approach is fundamentally under-constrained: there usually exist many material estimates capable of producing renderings that match the measurements, but
 many of these estimates can be unrealistic and may not generalize to novel illumination and viewing conditions.
The solution to this problem has been to \emph{regularize} the optimization using pre-determined \emph{material priors} such as linear low-dimensional BRDF models \cite{Ren2011,Hui2017} or stationary stochastic textures \cite{Aittala2015,Aittala2016}.
However, such hand-crafted priors do not generalize to a wide range of real-world materials.

More recently, learning-based approaches have demonstrated remarkable results for reconstructing SVBRDFs from one \cite{Deschaintre2018,Li2018} or more images \cite{Deschaintre2019}.
While these methods use rendering-based losses (similar to the inverse rendering approaches) during training, at test time they predict SVBRDFs from images using a single feed-forward pass through a deep network.
As a result, the reconstructed material parameters may not accurately reproduce the measured appearance.
In contrast, Gao et al. \shortcite{Gao2019} propose using an optimization-based approach in conjunction with a learned material prior.
Specifically, they train a fully-convolutional auto-encoder on a large material dataset and optimize in the latent space of this auto-encoder. This ensures that the reconstructed SVBRDF parameters both reproduce the measurements and are plausible real-world materials.
However, while this learned material prior is a significant improvement over hand-crafted priors, it still produces a relatively localized and highly flexible latent space that requires a good initialization (for example, from single image methods \cite{Deschaintre2018,Li2018}) and even then can fail to produce good results.

In this paper, we propose a different material prior that builds on the remarkable progress in image synthesis using deep Generative Adversarial Networks (GANs) \cite{Goodfellow2014,Karras2018,StyleGAN}.
We train \emph{MaterialGAN}---a StyleGAN2-based deep convolutional neural network \cite{StyleGAN2}---to generate plausible materials from a large-scale, spatially-varying material dataset \cite{Deschaintre2018}.
MaterialGAN learns \emph{global} correlations in material parameters, both spatially (thus encoding texture patterns) as well as across parameters (for example, relationships between diffuse and specular parameters).
As illustrated in Figure~\ref{fig:material_gan_samples}, sampling from the MaterialGAN latent space produces plausible, realistic materials with complex variations and diverse appearance.

While GANs have traditionally been used to synthesize images, we demonstrate a very different application, using MaterialGAN as a powerful prior in an inverse rendering-based material capture framework.
We append a rendering layer to MaterialGAN, setting up a differentiable pipeline from the learned latent space, through generating material maps, to rendering images under specified views and lighting.
This allows us to optimize the MaterialGAN latent vector(s) to minimize the error between the rendered and measured images and reconstruct the corresponding material maps.
Doing so ensures that the reconstructed SVBRDFs lie on the ``manifold of realistic materials'', while at the same time accurately reproducing the captured images.


We demonstrate that our GAN-based optimization framework produces high-quality SVBRDF reconstructions from a small number (3-7) images captured under flash illumination using hand-held mobile phones, and improves upon previous state-of-the-art methods \cite{Gao2019,Deschaintre2019}.
In particular, it produces cleaner, more realistic material maps that better reproduce the appearance of the captured material under both input \emph{and novel} lighting.
Moreover, as illustrated in Figure~\ref{fig:real}, MaterialGAN adapts to a wide range of SVBRDF samples ranging from diffuse to specular materials and near-stochastic textures to structured patterns with multiple distinct, complex materials.

Furthermore, our GAN-based latent space offers the ability to edit the latent vector in semantically meaningful ways (via operations like interpolation in the latent space) and generate realistic materials that go beyond the captured images.
This is not possible with current material capture methods that do not afford any control over their per-pixel BRDF estimates.

    \section{Related work}
\label{sec:related}

\paragraph{Reflectance capture.}
\revision{Acquiring} material data from physical measurements is the goal of a broad range of methods.
\revision{Please refer to surveys~\cite{weyrich09,Guarnera2016,dong19} for more comprehensive introduction to the related works.}

Most \revision{reflectance capture} approaches 
observe a material sample under \revision{varying viewing} and lighting configurations. They differ in the number of light patterns required and their \revision{types such as} moving linear light \cite{Gardner2003,Ren2011}, Gray code patterns \cite{Francken2009}, spherical harmonic illumination \cite{Ghosh2009}, and Fourier patterns \cite{Aittala2013}.

Methods have also been proposed for material capture ``in the wild'', i.e., under uncontrolled environment conditions with commodity hardware, typically captured with a hand-held mobile phone with flash illumination. Some of these methods impose strong priors on the materials, such as linear combinations of basis BRDFs \cite{Hui2017,Xu2016} (where the basis BRDFs can come from the measured data \cite{Matusik2003}). Later work by Aittala et al. \shortcite{Aittala2015,Aittala2016} estimated per-pixel parameters of stationary spatially-varying SVBRDFs from two-shot and one-shot photographs.
In the latter case, the approach used a neural Gram-matrix texture descriptor based on the texture synthesis and feature transfer work of Gatys \shortcite{Gatys2015,Gatys2016} to compare renderings with similar texture patterns but without pixel alignment.

More recently, deep learning-based approaches have demonstrated remarkable progress in the quality of SVBRDF estimates from single images (usually captured under flash illumination) \cite{Li2017,Deschaintre2018,Li2018}. These methods train deep convolutional neural networks with large datasets of artistically created SVBRDFs, and with a combination of losses that evaluate the difference in material maps and renderings from the dataset ground truth.

Deschaintre \shortcite{Deschaintre2019} extended the single-shot approach to multiple images. The key idea is to extract features from the input images with a shared encoder, max-pooling the features and decoding the final maps from the pooled features. This architecture has the benefit of being independent of the number of inputs, while also not requiring explicit light position information. In our experience, this approach produces smooth, plausible maps with low artifacts; however, re-rendering the maps tends to be not as close to the target measurements because the network cannot ``check'' its results at runtime. Moreover, we find that especially on real data, this method also has strong biases such as dark diffuse albedo maps and exaggerating surface normals (especially along strong image gradients that might be caused by albedo variations). We believe this is not due to any technical flaw; the method may be reaching the limit of what is possible using current feed-forward convolutional architectures and currently available datasets.

Gao et al. \shortcite{Gao2019} introduced an inverse rendering-based material capture approach that optimizes for material maps to minimize error with respect to the captured images. Since this is an under-constrained problem, they propose optimizing over the latent space of a learned material auto-encoder network to minimize rendering error. This approach has the benefit of explicitly matching the appearance of the captured image measurements, while also using the auto-encoder as a material ``prior''. Moreover, the encoder and decoder are fully convolutional, which has the advantage of resolution independence.
However, we find that the convolutional nature of this model also has the disadvantage of only providing local regularization and not capturing global patterns in the material, such as the long-range spatial patterns and correlations between the different material parameter maps. As a result, this method relies on previous methods (for example, Deschaintre et al.~\shortcite{Deschaintre2018}) to provide a good initialization, without which it can converge to poor results.
In contrast, our MaterialGAN is a more globally robust latent space and produces higher quality reconstructions without requiring accurate initializations, though it is no longer resolution-independent.

\paragraph{Generative adversarial networks.}
GANs~\cite{GAN} have become extremely successful in the past few years in various domains, including images~\cite{DCGAN}, video~\cite{Tulyakov18}, audio~\cite{Donahue18}, \revision{and 3D shapes~\cite{Li19}}. A GAN typically consists of two competing networks; a generator, whose goal is to produce results that are indistinguishable from the real data distribution, and a discriminator, whose job is to learn to identify generated results from real ones. For generating realistic images (especially of human faces), there has been a sequence of improved models and training strategies, including ProgressiveGAN~\cite{Karras2018}, StyleGAN~\shortcite{StyleGAN} and StyleGAN2~\shortcite{StyleGAN2}. StyleGAN2 in particular is the state-of-the-art GAN model and our work is based on its architecture, modified to output more channels.

\revision{Recently, GANs have also been used to solve inverse problems \cite{Bora17,Asim19,Malley19}.}
\revision{In computer graphics and vision, this work has focused} on embedding images into the latent space, with the goal of editing the images in semantically meaningful ways via latent vector manipulations \cite{Zhu2016}. This embedding requires solving an optimization problem to find the latent vector.
More recent work such as Image2StyleGAN~\cite{Abdal19a} and Image2StyleGAN++~\cite{Abdal19b} has looked at problem of embedding images specifically into the the StyleGAN latent space. While these methods focus on projecting portrait images into face-specific StyleGAN models, we find their analysis can be adapted to our problem.
We build on this to propose a GAN embedding-based inverse rendering approach.


    \section{MaterialGAN: A Generative SVBRDF Model}
\label{sec:gan}
Generative Adversarial Networks \cite{Goodfellow2014} are trained to map an input from a latent space (often randomly sampled from a multi-variate normal distribution) to a plausible instance of the target distribution.
In recent years, GANs have made remarkable progress in synthesizing high-resolution, photo-realistic images. Inspired by this progress, we propose MaterialGAN, a GAN that is trained to generate plausible materials, thus implicitly learning an SVBRDF manifold. MaterialGAN is based on the architecture of StyleGAN2 \cite{StyleGAN2}.
\subsection{Overview of StyleGAN and its latent spaces}
\label{ssec:latent_space}
StyleGAN2 \cite{StyleGAN2} is an improvement of StyleGAN \cite{StyleGAN} and is the state-of-the-art generative adversarial network (GAN) for image synthesis, especially for human faces.
The architecture has several advantages over previous models like ProgressiveGAN~\cite{Karras2018} and DCGAN~\cite{DCGAN}. For our purposes, the main advantage is that the model is not simply a black-box stack of convolution and upsampling layers, but has additional, more specific structure, allowing for much easier inversion (latent space optimization).
The StyleGAN2 architecture starts with a learned constant $4 \times 4 \times 512$ tensor and progressively upsamples it to the final output target resolution via a sequence of convolutional and upsampling layers (7 in total to end with a final image resolution of $256 \times 256$).
Given an input latent code vector $\bz \in \calZ \subset \Real^{512}$, StyleGAN2 transforms it through a non-linear mapping network of fully-connected layers into an intermediate latent vector $\bw \in \calW \subset \Real^{512}$.
The rationale for the introduction of the space of~$\calW$ is that while $\calZ$ requires (almost) every latent $\bz \in \calZ$ to correspond to a realistic output, vectors $\bw \in \calW$ are free from this overly stringent constraint, which leads to a less ``entangled'' mapping, with more meaningful dimensions (see \cite{StyleGAN,StyleGAN2} for more discussion).
In the original StyleGAN, the vector $\bw \in \calW$ is mapped via a learned affine transformation to mean and variance ``style'' vectors that control adaptive instance normalizations (AdaIN) \cite{Huang2017} that are applied before and after every convolution in the generation process (thus $7 \times 2 = 14$ times for a model of resolution $256 \times 256$). The statistics of the AdaIN normalizations caused the feature maps and output images of StyleGAN to suffer from droplet artifacts. StyleGAN2 removes the droplet artifacts entirely by replacing the AdaIn normalization layers with a demodulation operation which bakes the entire style block into a single layer while maintaining the same scale-specific control as StyleGAN.
We construct a matrix $\bw^+ \in \calW^+ \subset \Real^{512 \times 14}$ by replicating $\bw$ 14 times.
During training and standard synthesis, the columns of $\bw^+$ are identical, and correspond to $\bw$.
However, as we will discuss later (and similar to Abdal et al. \shortcite{Abdal19a}), we relax this constraint when optimizing for an embedding; $\calW^+$ thus becomes an extended latent space, more powerful than $\calW$ or $\calZ$.
Additionally, StyleGAN2 injects Gaussian noise, $\noise$, into each of the 14 layers of the generator.
This noise gives StyleGAN2 the ability to synthesize stochastic details at multiple resolutions.
Abdal et al. \shortcite{Abdal19b} make the observation that one can also treat these noise inputs $\noise$ as a latent space $\calN$. Thus, combining these two spaces defines yet another latent space $\calWN$.
\setlength{\resLen}{.13\columnwidth}
\begin{figure}[t]
	\addtolength{\tabcolsep}{-4pt}
	\begin{tabular}{ccccccc}
		\includegraphics[width=\resLen]{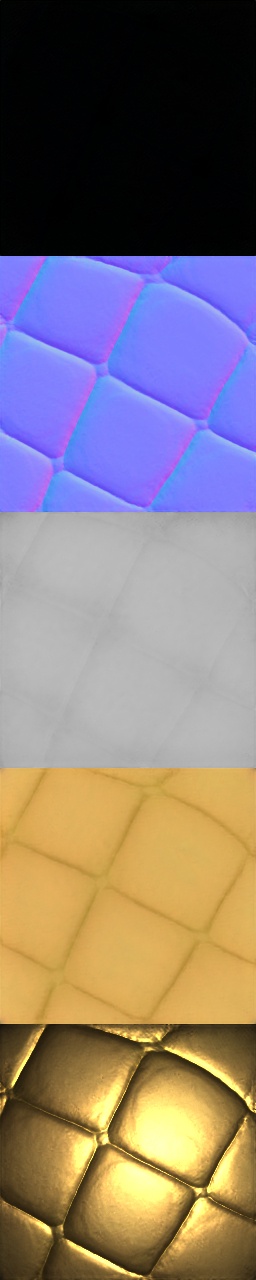} &
		\includegraphics[width=\resLen]{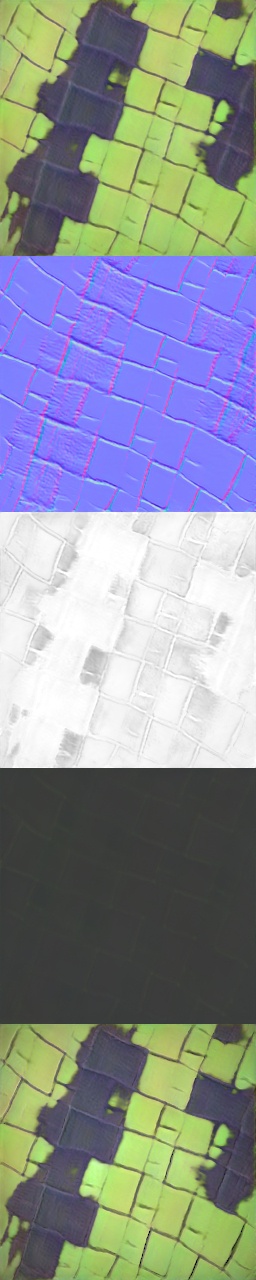} &
		\includegraphics[width=\resLen]{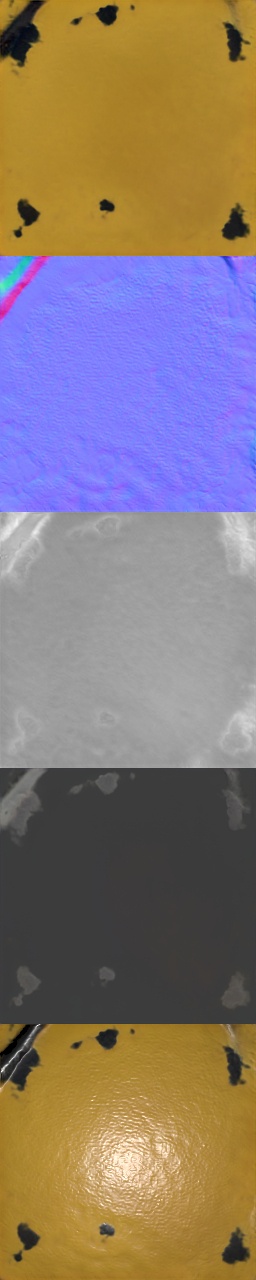} &
		\includegraphics[width=\resLen]{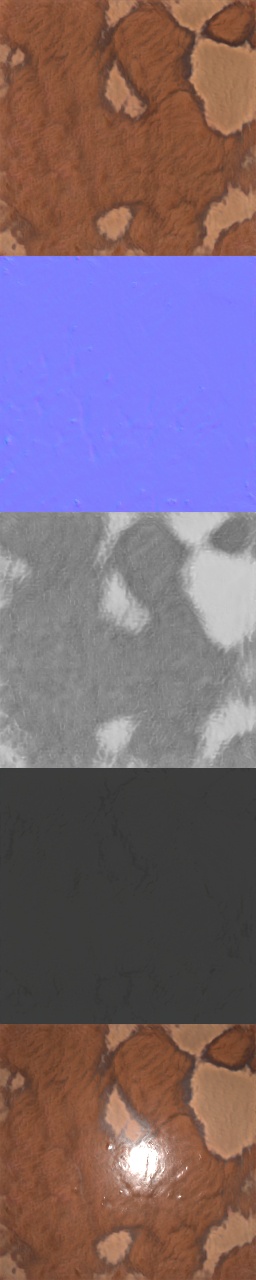} &
		\includegraphics[width=\resLen]{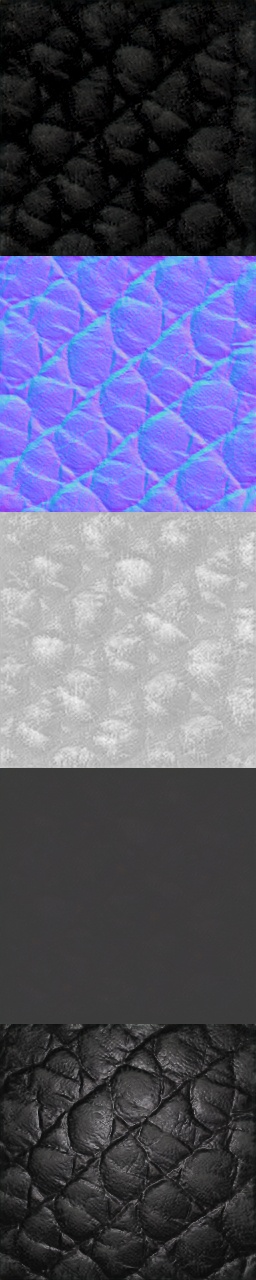} &
		\includegraphics[width=\resLen]{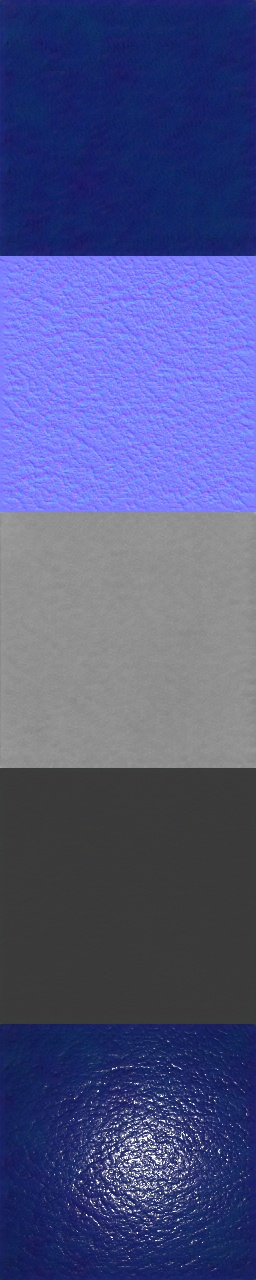} &
		\includegraphics[width=\resLen]{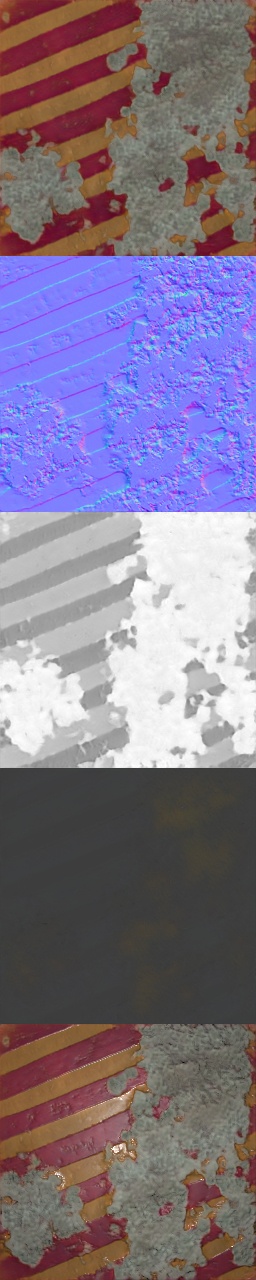}
	\end{tabular}
	\caption{Seven materials generated by randomly sampling MaterialGAN. Top to bottom: diffuse albedo, normal, roughness, specular albedo and renderings under flash illumination. As can be seen, the material maps are high-quality with meaningful correlations both spatially and across materials parameters, and visually look like plausible real-world materials.
	}
	\label{fig:material_gan_samples}
\end{figure}
\subsection{MaterialGAN training}
\label{ssec:training}
MaterialGAN was trained with the dataset provided by Deschaintre et al.~\shortcite{Deschaintre2018} (and also used in Gao et al.~\shortcite{Gao2019}). They generated this dataset by sampling the parameters of procedural material graphs from Allegorithmic Substance Share to create an initial set of 155 high-quality SVBRDFs at resolution $4096 \times 4096$. The dataset was augmented by blending multiple SVBRDFs and generating $256 \times 256$ resolution crops at random positions, scales and rotations. The final dataset consists of around 200,000 SVBRDFs. For detailed information about the curation of dataset we refer the reader to \cite{Deschaintre2018}. Since pairs of SVBRDFs in the dataset were the same with only a slight variation, we selected 100,000 SVBRDFs.
The maps for each SVBRDF are stacked in 9 channels (3 for albedo, 2 for normals, 1 for roughness, and 3 for specular albedo). We account for this by adapting the MaterialGAN architecture to output 9-channel outputs. MaterialGAN is trained in TensorFlow (version 1.15) with the same loss functions and similar hyper-parameters from StyleGAN2 \cite{StyleGAN2}. StyleGAN2 configuration F was used for all experiments. The generator and discriminator were trained using Adam optimizers. The learning rate was increased per resolution from 0.001 to 0.0025 for both the generator and the discriminator. The discriminator was shown 25 million images. Training on 8$\times$ Nvidia Tesla V100 takes about 5 days.
Figure~\ref{fig:material_gan_samples} shows materials generated by randomly sampling the MaterialGAN latent space and images rendered from them. As can be seen here, MaterialGAN generates a wide variety of nearly photorealistic materials ranging from structured to stochastic, diffuse to specular, and with large-scale variations to fine detail. Furthermore, Figure~\ref{fig:fakemorph} \revision{and the accompanying video show} example interpolations between pairs of generated materials in the latent space, producing plausible non-linear morphing results.
\setlength{\resLen}{.95\columnwidth}
\begin{figure}[tb]
	\centering
	\includegraphics[width=\resLen]{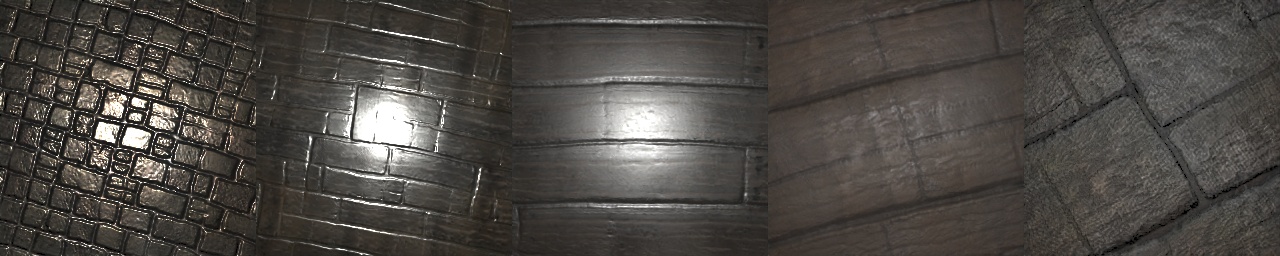}
	\includegraphics[width=\resLen]{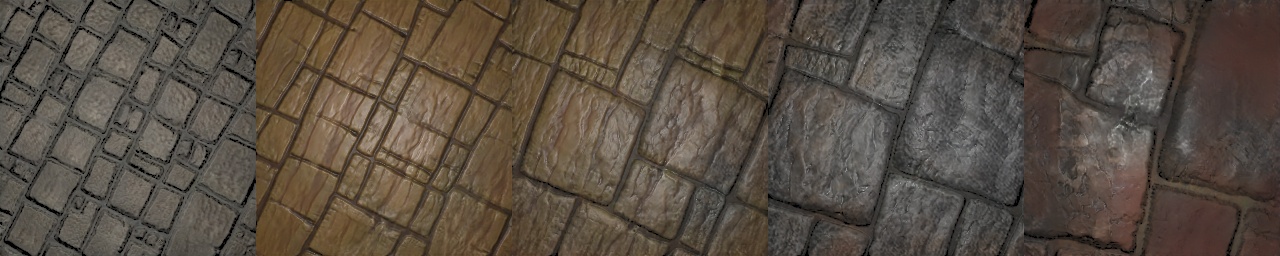}
	\includegraphics[width=\resLen]{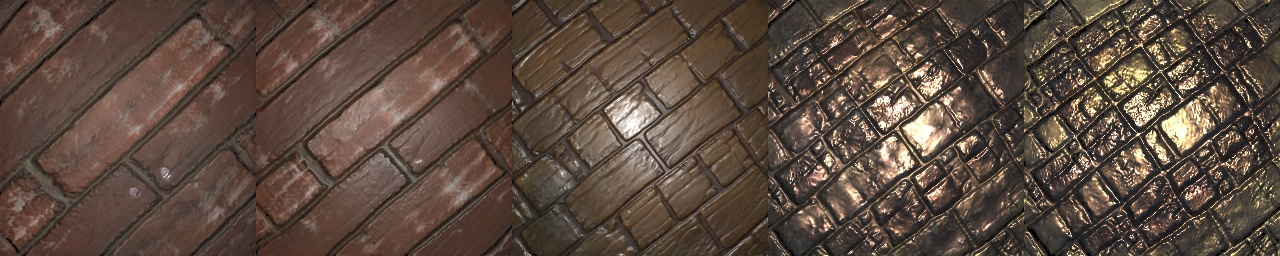}
	\includegraphics[width=\resLen]{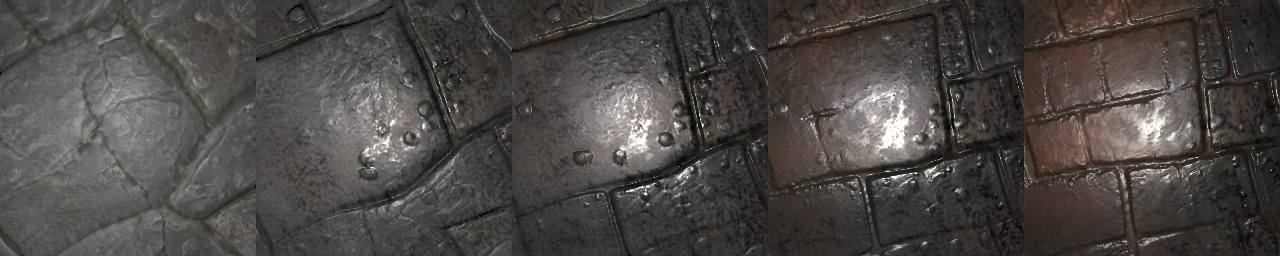}
	\caption{\label{fig:fakemorph}
		\textbf{Interpolation in MaterialGAN latent space.} Each row shows an example of interpolation between two randomly generated materials, demonstrating non-linear morphing behavior.
	}
\end{figure}

    \section{SVBRDF Capture using MaterialGAN}
\label{sec:framework}
We utilize MaterialGAN, the powerful generative model described in the previous section, in a fundamentally new fashion: to \emph{capture} SVBRDF maps.
Specifically, we use MaterialGAN as a \emph{material prior} for SVBRDF acquisition via an inverse rendering framework.
Our goal is to estimate the SVBRDF parameter maps from one or a small number of photographs of a near-planar material sample.
We utilize a common BRDF model that involves a diffuse and a specular component using the microfacet BRDF with the GGX normal distributions~\cite{Walter07}.
Our unknown parameter vectors $\params := (\albedo,\normal,\rough,\spec)$ encode the four per-pixel parameter maps: diffuse albedo~$\albedo$, surface normal~$\normal$, roughness~$\rough$, and specular albedo~$\spec$.
To recover the unknown parameter maps, we capture $k$ images $\img_1, \cdots, \img_k$.
We assume known viewing and lighting configurations for each image, which we denote as $(\light_i, \camera_i)$.
Further, we assume that the material is lit by a single point source, collocated with the camera.%
\footnote{%
	In theory, non-collocated lights, area lights or projection patterns (e.g. on an LCD or similar screen) can be used as well, and would require a straightforward modification to our forward rendering process.
}
The images can be reprojected into a common frontal view (which is straightforward with a known viewing configuration).
We introduce a differentiable rendering operator $\render$ that takes as input the parameter maps as well as the viewing and lighting configurations, and synthesizes corresponding images of the material.
Under this setup, our goal is to find values of the unknown parameters $\params$ so that renderings with these parameters match the measurements $\img_i$.
In other words, we focus on solving the following optimization problem:
\begin{equation}
	\label{eq:opt1}
	\paramsopt = \textstyle\argmin_{\albedo, \normal, \rough} \sum_{i=1}^k \loss(\render(\params; \,\light_i, \camera_i), \img_i),
\end{equation}
where $\loss$ is a loss function that measures the difference between the captured images, $\img_i$ and the renderings generated from the estimated SVBRDF parameters, $\render(\params; \,\light_i, \camera_i)$.
\begin{figure}[t]
	\centering
	\includegraphics[width=1.0\columnwidth]{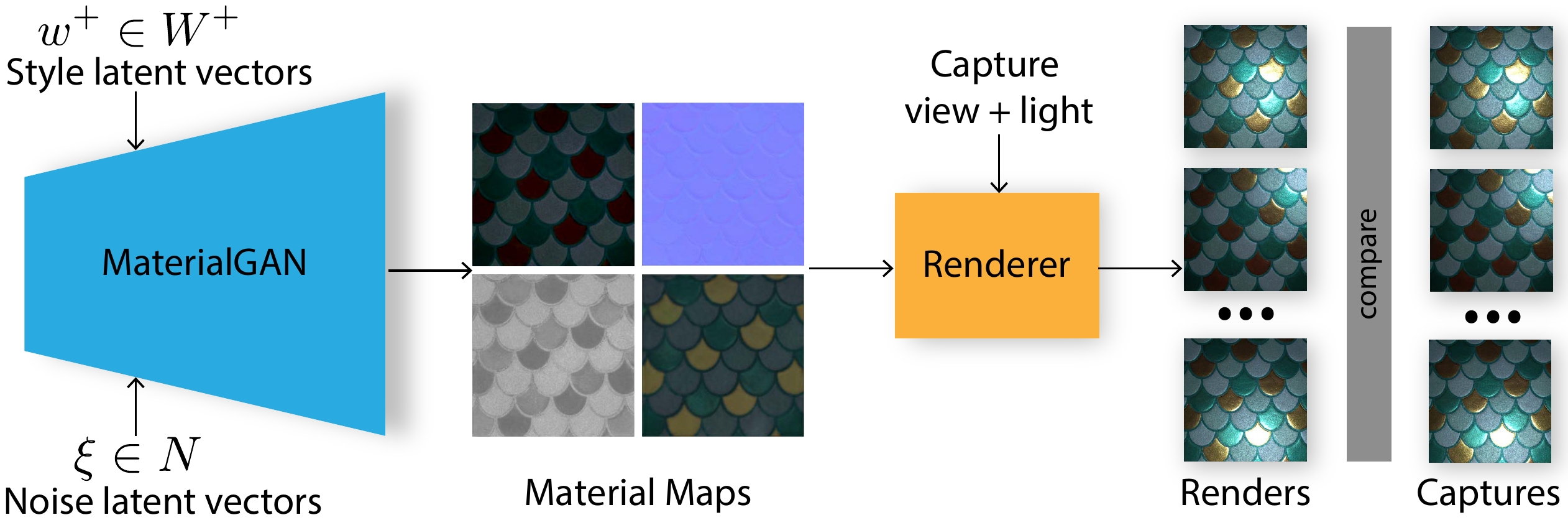}
	\caption{Our inverse rendering pipeline. We optimize for latent vectors $\bw^+$ and $\noise$, that feed into the layers of the StyleGAN2-based MaterialGAN model. The MaterialGAN generator produces material maps (diffuse albedo, normal, roughness and specular albedo), that are rendered under the captured view/light settings. Finally, the renderings and measurements are compared using a combination of L2 and perceptual losses.}
	\label{fig:system}
\end{figure}

\subsection{Incorporating the MaterialGAN prior}
Eq.~\eqref{eq:opt1} is, in general, a challenging optimization to solve due to its under-constrained nature.
Given a small number of input measurements, the optimization can overfit to the input, producing implausible maps that do not generalize to novel views and lighting.
To overcome this challenge, we leverage the MaterialGAN prior: instead of directly optimizing for the parameter maps~$\params$,  we can optimize for a vector $\latent$ in the MaterialGAN \emph{latent space} and map (decode) this latent vector back into material maps $\params$.
The optimization problem then becomes:
\begin{equation}
	\label{eq:opt_gan}
	\latentopt = \textstyle\argmin_{\latent} \sum_{i=1}^k \loss(\render(\generator(\latent); \,\light_i, \camera_i), \img_i),
\end{equation}
where $\generator$ is the learned MaterialGAN generator.
Given that both $\generator$ and $\render$ are differentiable operations, Eq.~\eqref{eq:opt_gan} can be optimized via gradient-based methods to estimate $\latentopt$ and the corresponding SVBRDF maps $\generator(\latentopt)$.
The above operation is similar to recent work on embedding images in the StyleGAN latent space \cite{Abdal19a,Abdal19b}.
The key difference is that we do not match material parameters directly, but evaluate their error through the rendering operator $\render(\cdot)$.
To our knowledge, ours is the first approach to use a GAN latent space in combination with a rendering operator.
\paragraph{Loss function.}
We optimize Eq.~\ref{eq:opt_gan} using a combination of a standard per-pixel L2 loss and a ``perceptual loss'' \cite{Johnson2016} that has been shown to produce sharper results in image synthesis tasks:
\begin{equation}
	\label{eq:loss}
	\loss(\img, \img') = \lambda_1 \lossPix + \lambda_2 \lossPercp,
\end{equation}
The perceptual loss is defined as:
\begin{equation}
	\lossPercp(\img, \img') = \textstyle\sum_{j=1}^4 w^\mathrm{percept}_j \left\|F_j(\img) - F_j(\img')\right\|_2^2,
\end{equation}
where $F_1, \cdots, F_4$ are the flattened feature maps corresponding to the outputs of VGG-19 layers  \texttt{conv1\_1}, \texttt{conv1\_2}, \texttt{conv3\_2}, and \texttt{conv4\_2} from a pre-trained VGG network \cite{VGG}. See section \ref{ssec:optim} for more details.
\paragraph{Optimization details.}
We convert the TensorFlow-trained MaterialGAN model to PyTorch, in which our optimization framework is implemented. We optimize Eq.~\ref{eq:opt_gan} using the Adam optimizer in PyTorch, with a learning rate of $0.01$. We set all other hyper-parameters to default values.
Now that our basic optimization framework is set up, there remain two key ingredients to implement our GAN-based optimization framework (Eq.~\eqref{eq:opt_gan}): (i)~the choice of \emph{latent space} that we optimize $\latent$ over, and (ii)~our optimization strategy to minimize the objective function. In the following sections, we describe our approach, along with an empirical analysis of these design choices.
\setlength{\resLen}{.47\columnwidth}
\begin{figure}[t]
	\addtolength{\tabcolsep}{-4pt}
	\begin{tabular}{c @{\hspace{2\tabcolsep}} ccc}
	\raisebox{.15in}{\rotatebox[origin=c]{90}{\footnotesize{GT}}} &
	\includegraphics[width=\resLen]{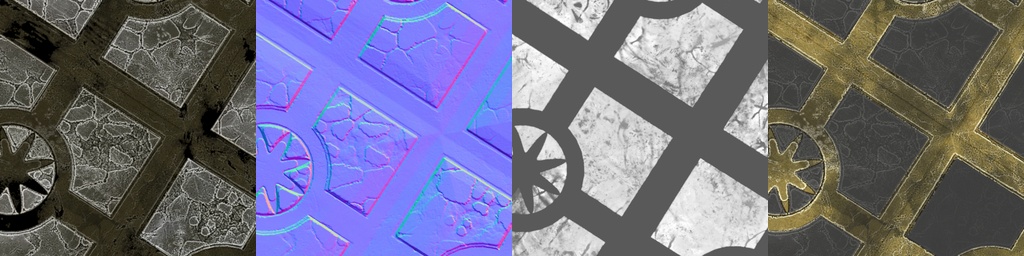} & &
	\includegraphics[width=\resLen]{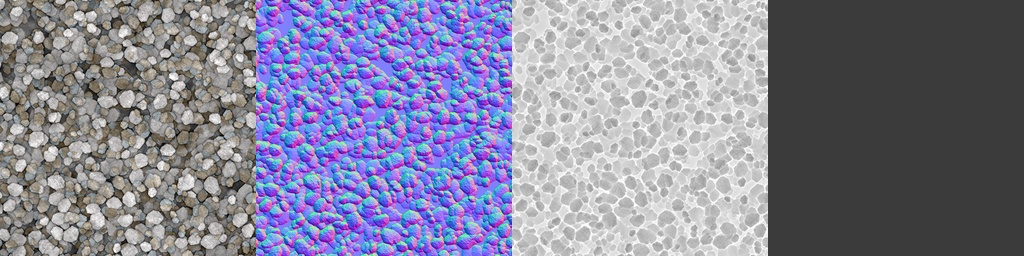}
	\\
	\raisebox{.15in}{\rotatebox[origin=c]{90}{\footnotesize{$\calW$}}} & 
	\includegraphics[width=\resLen]{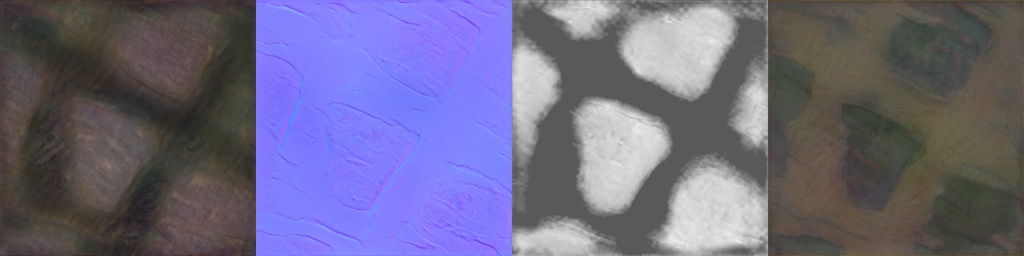} & &
	\includegraphics[width=\resLen]{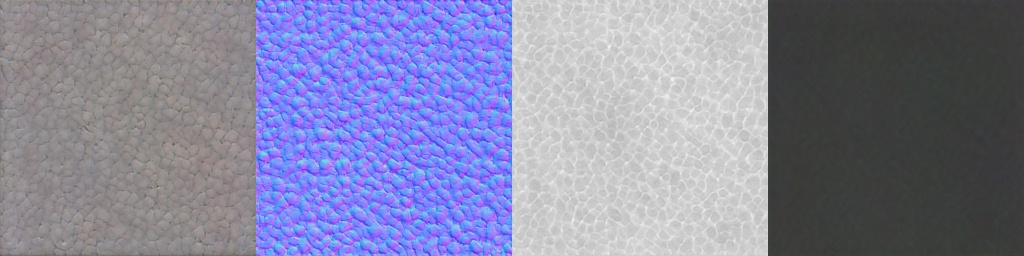}
	\\
	\raisebox{.15in}{\rotatebox[origin=c]{90}{\footnotesize{$\calW^+$}}} &
	\includegraphics[width=\resLen]{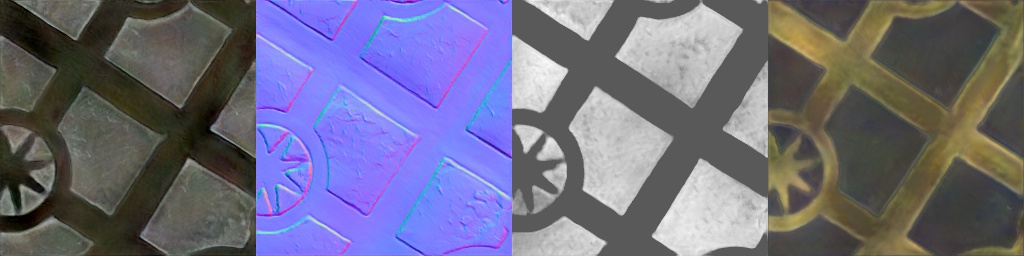} & &
	\includegraphics[width=\resLen]{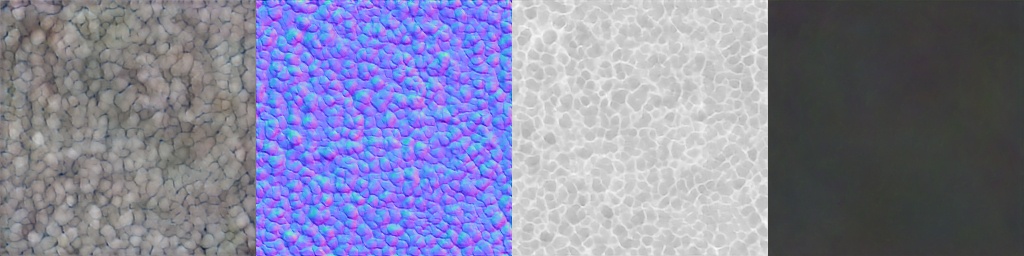}
	\\
	\raisebox{.15in}{\rotatebox[origin=c]{90}{\footnotesize{$\calW^+\calN$}}} &
	\includegraphics[width=\resLen]{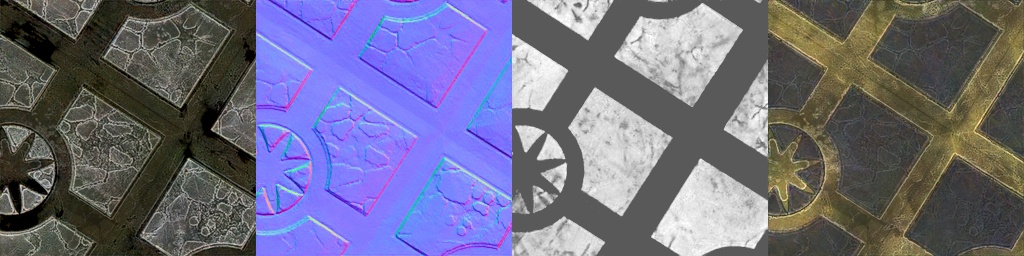} & &
	\includegraphics[width=\resLen]{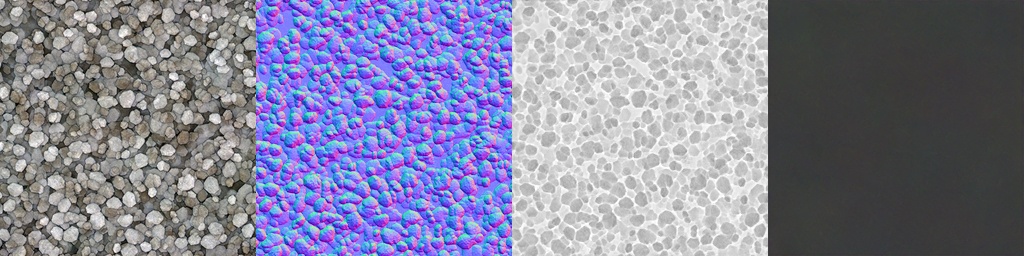}
	\end{tabular}
	\caption{\label{fig:embed}
		\textbf{Embedding SVBRDFs into different latent spaces.} We take two synthetic SVBRDF material maps (top) and embed them into different latent spaces with and without the noise space (second--fourth rows). For illustration, we also embed the maps into a pure-noise space \emph{only}; this is unable to recover the color at all.
	}
\end{figure}
\subsection{Latent space}
\label{ssec:latent}
As discussed in Sec. \ref{ssec:latent_space}, StyleGAN2 (and consequently, MaterialGAN) has a number of potential latent spaces.
In particular, MaterialGAN uses three different \emph{style} latent spaces: the input latent code $\bz \in \calZ$ , the intermediate latent code $\bw \in \calW$ and per-layer styles $\bw^+ \in \calW^+$.
StyleGAN2 also injects noise $\noise \in \calN$ into every layer of the network to generate stochastic variations.
The typical forward generation process of the GAN only uses $\bz$, with $\bw$ being generated from $\bz$ via a mapping network, and $\bw^+$ being generated from $\bw$ via affine transformations.
However, Abdal~et~al.~\shortcite{Abdal19a} note that the space of $\calZ$ is too restrictive for accurate embedding of faces or other content into the GAN space.
In other words, given the image of a human face, it is generally impossible to find a single $\bz \in \calZ$ such that the generated image closely matches the target.
This remains the case even when extending the space to $\calW$, i.e., when searching for a $\bw$ instead of a $\bz$.
The space $\calW^+$, on the other hand, offers much stronger representative power.
Our experiments on embedding material maps into MaterialGAN demonstrate that optimizing for $\calW^+$ is also needed for MaterialGAN to accurately reproduce input maps.
We demonstrate this in Figure \ref{fig:embed}, via an experiment where we embed a given material (with known material maps) into MaterialGAN.
As shown in rows (2) and (3), maps generated by optimizing $\bw^+ \in \calW^+$ contain more detail compared to those using $\bw \in \calW$.
\renewcommand{\one}{fake_038}

\setlength{\resLen}{.51in}
\begin{figure}[t]
	\addtolength{\tabcolsep}{-4pt}
	\begin{tabular}{cccc}
		  & \textbf{\footnotesize{SVBRDF maps}} & \textbf{\small Opt.} & \textbf{\small Novel}
		\\
		\raisebox{.2in}{\rotatebox[origin=c]{90}{\footnotesize{GT}}} &
		\includegraphics[height=\resLen]{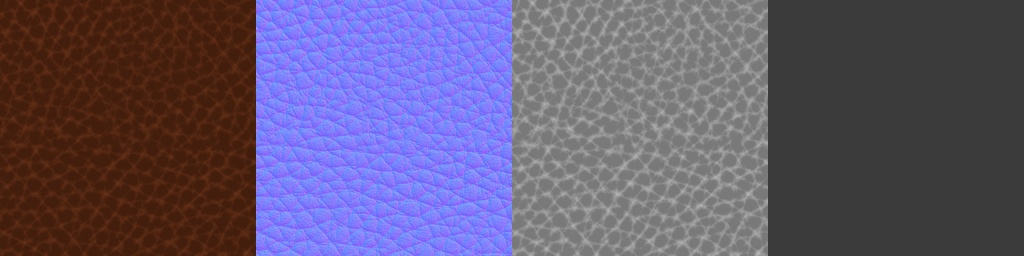} &
		\includegraphics[height=\resLen]{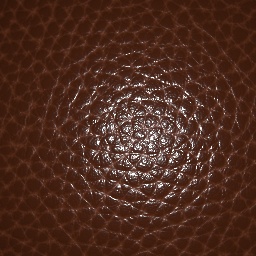} &
		\includegraphics[height=\resLen]{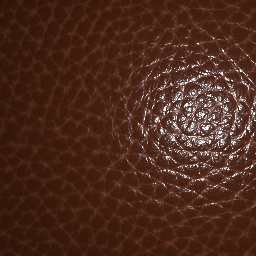}
		\\
		\raisebox{.2in}{\rotatebox[origin=c]{90}{\footnotesize{(1)}}} &
		\includegraphics[height=\resLen]{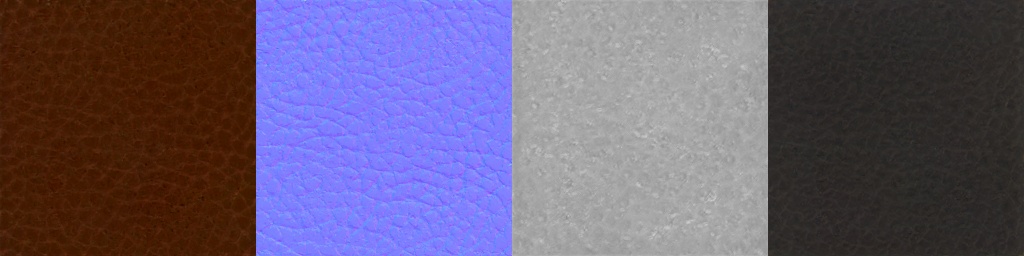} &
		\includegraphics[height=\resLen]{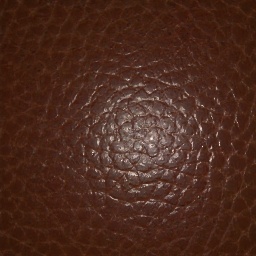} &
		\includegraphics[height=\resLen]{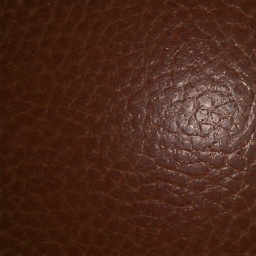}
		\\
		\raisebox{.2in}{\rotatebox[origin=c]{90}{\footnotesize{(2)}}} &
		\includegraphics[height=\resLen]{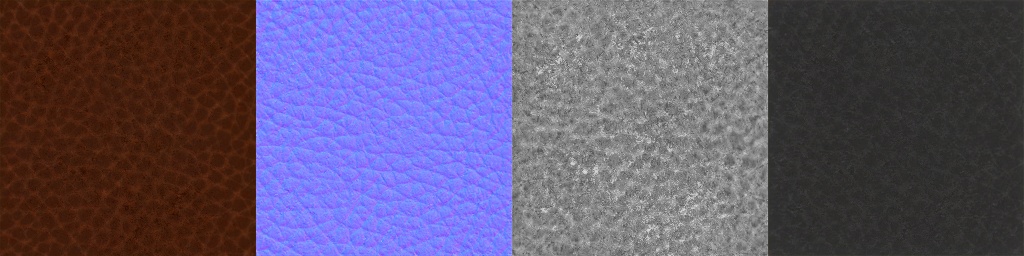} &
		\includegraphics[height=\resLen]{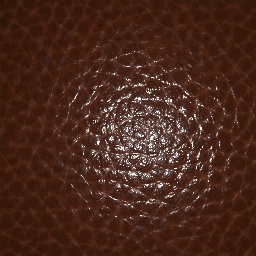} &
		\includegraphics[height=\resLen]{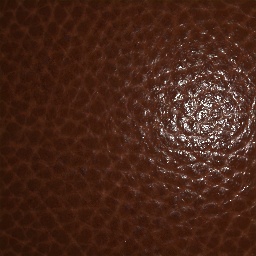}
		\\
		\raisebox{.2in}{\rotatebox[origin=c]{90}{\footnotesize{(3)}}} &
		\includegraphics[height=\resLen]{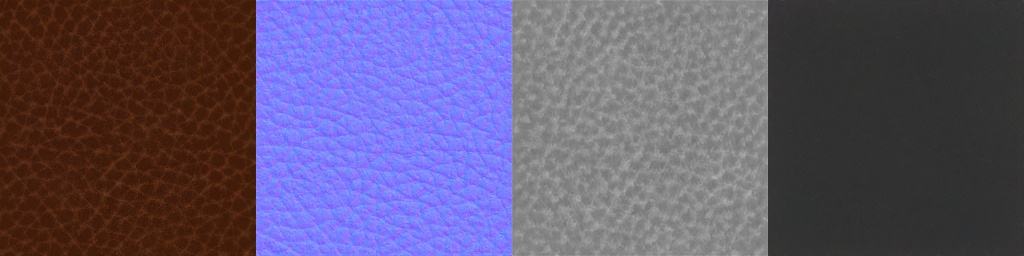} &
		\includegraphics[height=\resLen]{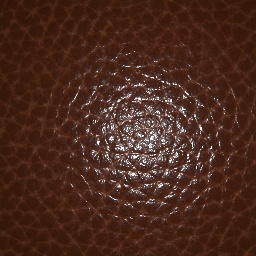} &
		\includegraphics[height=\resLen]{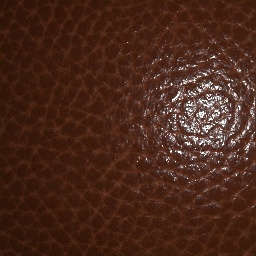}
	\end{tabular}
	\caption{\label{fig:optimize}
		\textbf{Optimization strategy.} We evaluated three optimization strategies: (1) optimize $\bw^+$ first, then $\noise$; (2) jointly optimize $\bw^+$ and $\noise$; (3) alternate between $\bw^+$ and $\noise$ every 10 iterations. Strategy (1) causes artifacts during the optimization, and (2) brings more noise into the maps. Particularly, for textures with small features, (1) and (2) may drive the optimization to bad local minima while the per-pixel loss could still be very low. Strategy (3) appears to be a good compromise, giving us better results in most cases. Note: ``Opt.'' means an  optimized input view or its re-rendering, i.e. not a novel view.
	}
\end{figure}

On the other hand, some small-scale details are still missing.
In fact, according to our experiments, only colors and large-scale features can be captured by the $\calW^+$ space.
For depicting high-frequency patterns, as demonstrated in rows (4) and (5) of Figure \ref{fig:embed}, we need to go even further and optimize the noise vector $\noise$ (instead of drawing it from multi-variate normal distributions).
We note that optimizing for the noise component is even more important in MaterialGAN, compared to embedding faces in StyleGAN or StyleGAN2.
We suspect that this is because with human faces, the distinction between large-scale features (e.g., eyes, noise, and mouth) and small-scale features (e.g., winkles) is very prominent, allowing the $\calW^+$ space to focus mostly on the large-scale features while leaving the small-scale ones to the noise vector $\noise \in \calN$.
In our case, the boundary between large-scale and small-scale material features is much less distinct.
The physical scales of real-world materials varies in a continuous fashion, making it virtually impossible to assign them to only one of the $\calW^+$ and $\calN$ spaces.
We hypothesize that for this reason, we need to focus on both $\calW^+$ and $\calN$ to achieve high-quality reconstruction of SVBRDF maps.
Based on these empirical observations, estimating SVBRDF parameter maps from photographs using our pre-trained MaterialGAN boils down to solving the following optimization:
\begin{equation}
	\latentopt = \argmin_{\bw^+ \in \calW^+,\, \noise \in \calN} \sum_{i=1}^k \loss(\render(\generator(\bw^+, \noise); \,\light_i, \camera_i), \img_i).
\end{equation}
Since there are two variables $\bw^+$ and $\noise$ that behave in a correlated fashion, a proper optimization strategy is crucial to achieve high-quality results. We now discuss our alternating two-step optimization method.
\subsection{Optimization strategy}
\label{ssec:optim}
Abdal~et~al.~\shortcite{Abdal19a,Abdal19b} recommended using a two-stage setting by first optimizing $\bw^+$ (with $\noise$ fixed) and then $\noise$ (with $\bw^+$ fixed).
In our case, this approach does work in some cases but is not always the top-performing option.
In addition to this strategy, we propose two alternatives, leading to three different optimization schemes:
\begin{enumerate}
	\item \textbf{Strategy 1}: Optimize $\bw^+$ first, then optimize $\noise$;
	\item \textbf{Strategy 2}: Jointly optimize both $\bw^+$ and $\noise$;
	\item \textbf{Strategy 3}: Alternatively optimize $\bw^+$ and $\noise$ for a small number (for example, 10) of iterations each.
\end{enumerate}
Figure~\ref{fig:optimize} shows a comparison of these strategies.
All of them give reasonable results, but Strategy 1 is better suited for materials with strong large-scale features.
Strategy 2 provides the fastest convergence because it allows the noise vector $\noise$ to be modified from the very beginning.
This, however, generally causes the optimization to use $\noise$ for encoding higher-level features and is prone to overfitting.
Finally, Strategy 3---a hybrid of Strategies 1 and 2---behaves in a more robust fashion than either of the previous strategies in most cases.
We use Strategy 3 for all the results in our paper.
Additionally, our experiments indicate that it is desirable to use different VGG layer weights for the optimization of $\bw^+$ and $\noise$. The weights we are using are, for $\bw^+$: [1/512,1/512,1/128,1/64]; for $\noise$: [1/64,1/64,1/256,1/512].
\renewcommand{\one}{fake_037}
\renewcommand{\two}{real_book1}

\setlength{\resLen}{.51in}
\begin{figure}[t]
	\addtolength{\tabcolsep}{-4pt}
	\begin{tabular}{cccc}
		& \textbf{\footnotesize{SVBRDF maps}} & \textbf{\small Opt.} & \textbf{\small Novel}
		\\
		\raisebox{.2in}{\rotatebox[origin=c]{90}{\footnotesize{GT}}} &
		\includegraphics[height=\resLen]{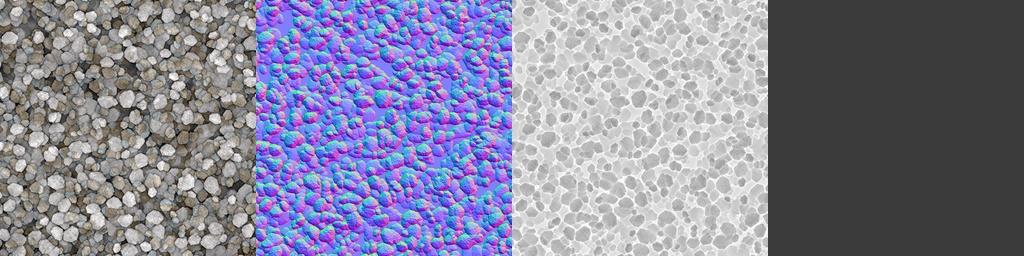} &
		\includegraphics[height=\resLen]{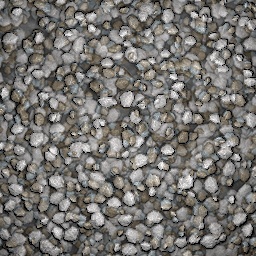} &
		\includegraphics[height=\resLen]{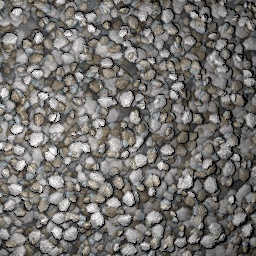}
		\\
		\raisebox{.2in}{\rotatebox[origin=c]{90}{\footnotesize{(1)}}} &
		\includegraphics[height=\resLen]{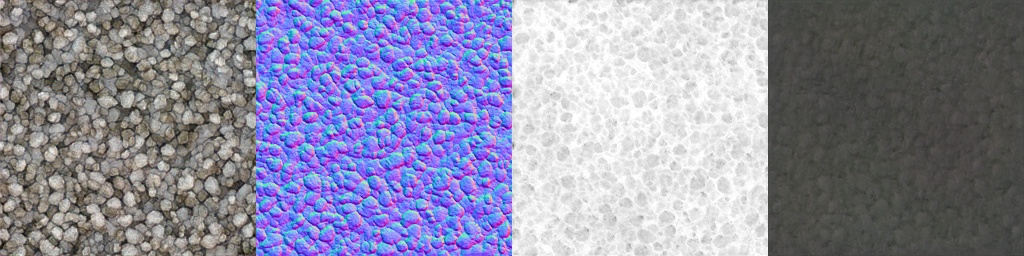} &
		\includegraphics[height=\resLen]{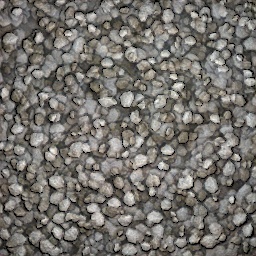} &
		\includegraphics[height=\resLen]{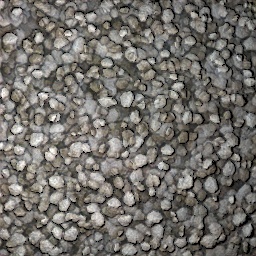}
		\\
		\raisebox{.2in}{\rotatebox[origin=c]{90}{\footnotesize{(2)}}} &
		\includegraphics[height=\resLen]{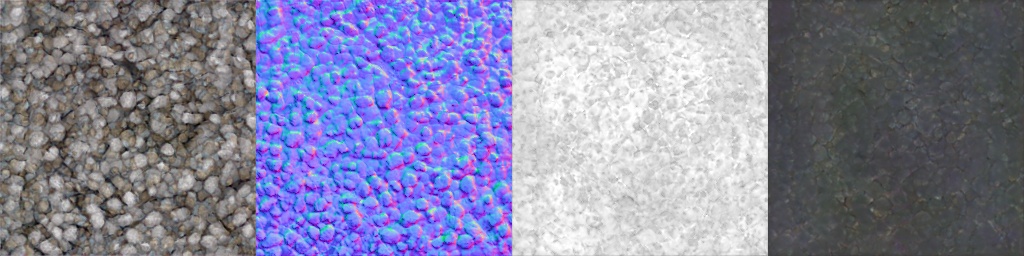} &
		\includegraphics[height=\resLen]{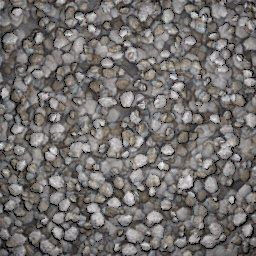} &
		\includegraphics[height=\resLen]{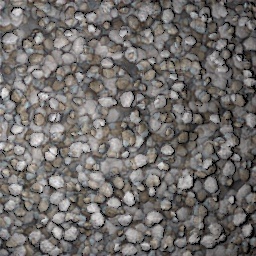}
		\\
		\raisebox{.2in}{\rotatebox[origin=c]{90}{\footnotesize{GT}}} &
		 &
		\includegraphics[height=\resLen]{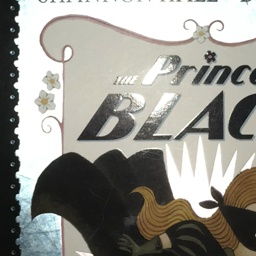} &
		\includegraphics[height=\resLen]{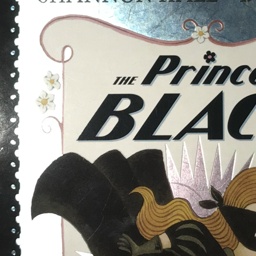}
		\\
		\raisebox{.2in}{\rotatebox[origin=c]{90}{\footnotesize{(1)}}} &
		\includegraphics[height=\resLen]{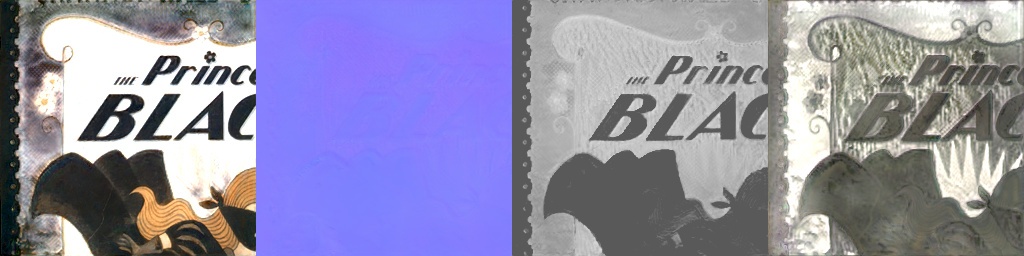} &
		\includegraphics[height=\resLen]{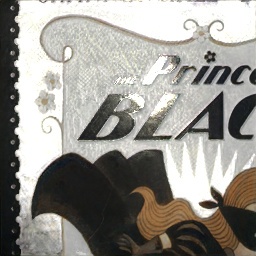} &
		\includegraphics[height=\resLen]{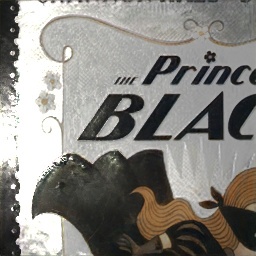}
		\\
		\raisebox{.2in}{\rotatebox[origin=c]{90}{\footnotesize{(2)}}} &
		\includegraphics[height=\resLen]{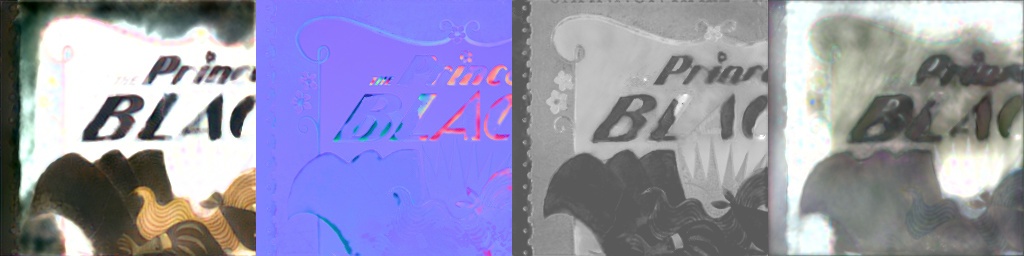} &
		\includegraphics[height=\resLen]{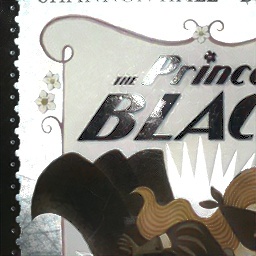} &
		\includegraphics[height=\resLen]{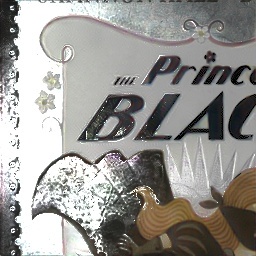}
	\end{tabular}
	\caption{\label{fig:noiseVSrefine}
		\revision{\textbf{Noise optimization vs. post-refinement.} (1) Optimize $\bw^+$ and $\noise$ but no post-refinement; (2) Optimize $\bw^+$ only but with post-refinement. This shows that $\noise$ takes an important role; optimizing only $\bw^+$ has too little expressive power and converges to suboptimal solutions, which post-refinement cannot fix (see especially normal maps in (2)).}
	}
\end{figure}
\paragraph{Noise optimization vs. post-refinement.}
\label{ssec:post-refine}
Instead of optimizing latent space $\bw^+$ with noise $\noise$, another option is to apply post-refinement (that is, pixel-space optimization without any latent space) after optimizing $\bw^+$ only. However, the space $\bw^+$ is too small to realistically match per-pixel detail: if optimizing $\bw^+$ only, the resulting maps have significant artifacts. Adding post-refinement to such a result essentially becomes per-pixel optimization (with little regularization), which tends to to work poorly with a small number of inputs. Optimizing $\noise$ offers more powerful regularization, as the noise is inserted into all layers of the generator, rather than just appended at the end (like post-refinement). We show two failure examples in Figure \ref{fig:noiseVSrefine}, where optimizing $\bw^+$ leads to unsatisfactory texture maps.
\begin{figure}[t]
	\addtolength{\tabcolsep}{-4pt}
	\begin{tabular}{ll@{\hspace{8\tabcolsep}}ll}
		\includegraphics[width=0.23\columnwidth]{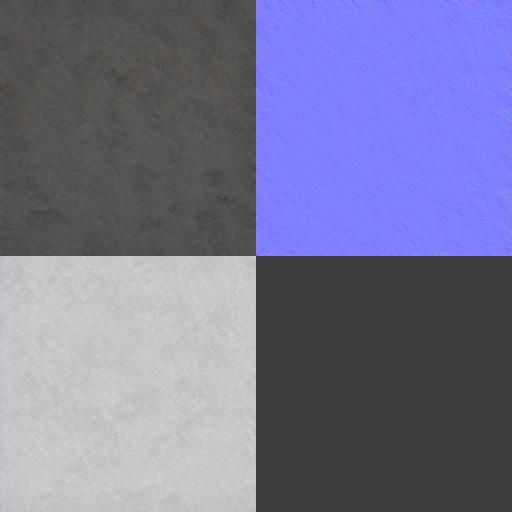} &
		\includegraphics[width=0.23\columnwidth]{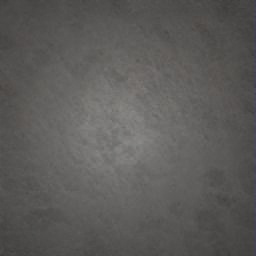} &
		\includegraphics[width=0.23\columnwidth]{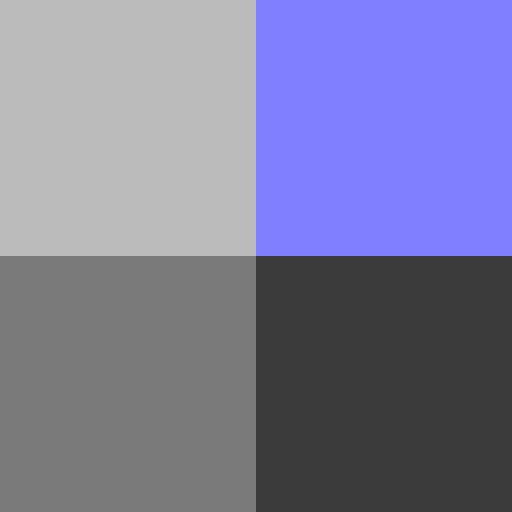} &
		\includegraphics[width=0.23\columnwidth]{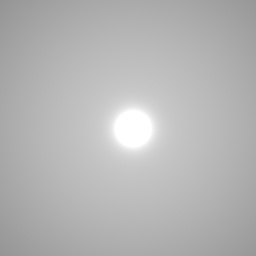}
		\\
		\multicolumn{2}{c}{\small (a) Mean $\bw$ initialization} & 
		\multicolumn{2}{c}{\small (b) Low roughness initialization} \\
	\end{tabular}
	\caption{\label{fig:init}
		\textbf{Visualization of our constant initializations.} We initialize our optimization with the two materials shown here and pick the result with the lowest final loss. (This applies in cases where we do not use the result from Deschaintre et al. as initialization, as detailed in the results section and supplementary materials.) Left: Material maps generated from the mean latent vector $\bw$. Right: An additional low roughness, specular initialization.
	}
\end{figure}
\subsection{Initialization}
\label{ssec:init}
We find that our method is robust to the initialization of the latent vectors. We experimented with using the same initial configuration---represented by the material produced by the mean $\bw$ of our GAN training data (see Figure \ref{fig:init}(a))---and found that is works well for most of the materials we tried (both synthetic and real).
However, this initialization represents a material with a high roughness (reflecting a bias in our training data) and sometimes leads to errors when fitting highly specular / low roughness materials.
Therefore, we add an additional low roughness initialization (see Figure \ref{fig:init}(b)).
In practice, given the captured images, we run our MaterialGAN optimization starting from both initializations and retain the result with the lowest optimization error of Eq.~\eqref{eq:loss}.
All of our results in this paper followed this scheme.
    \section{Results}
\label{sec:results}
\textbf{Only a small subset of our results fits into the paper. Please see our supplemental material and video for more results.}
\paragraph{Test data.}
For synthetic tests, we use several examples from the test set of Deschaintre~\shortcite{Deschaintre2018}, as well as some from the Adobe Stock dataset~\cite{Li2018}. This gives a total of \totSynthetic synthetic results.
For our real results, we use a hand-held mobile phone to capture images with flash, resulting in a collocated camera and point light illumination.
Similar to previous work \cite{Hui2017,Deschaintre2019}, we use a paper frame to register the multiple images.
We add markers to the frame to improve camera pose estimation.
Using this process, we capture \totReal physical samples with nine images per material, roughly covering the sample with $3 \times 3$ specular highlights. 
Unless otherwise specified, all our results use seven images for inverse-rendering optimizations and the remaining two (under novel lighting) for evaluating the results.
\paragraph{Inverse-rendering performance.}
Our optimization takes about 2 minutes to complete 2000 iterations on a Titan RTX GPU. In many cases, the results converge after ~500 iterations, but we use 2000 everywhere for simplicity.
\renewcommand{\one}{fake_022}
\renewcommand{\two}{fake_039}

\setlength{\resLen}{.45in}
\begin{figure}[t]
	\addtolength{\tabcolsep}{-4pt}
	\begin{tabular}{ccccc}
		& \textbf{\footnotesize{SVBRDF maps}} & \textbf{\small Opt.} & \multicolumn{2}{c}{\textbf{\small Novel views}}
		\\
		\raisebox{.2in}{\rotatebox[origin=c]{90}{\footnotesize{GT}}} &
		\includegraphics[height=\resLen]{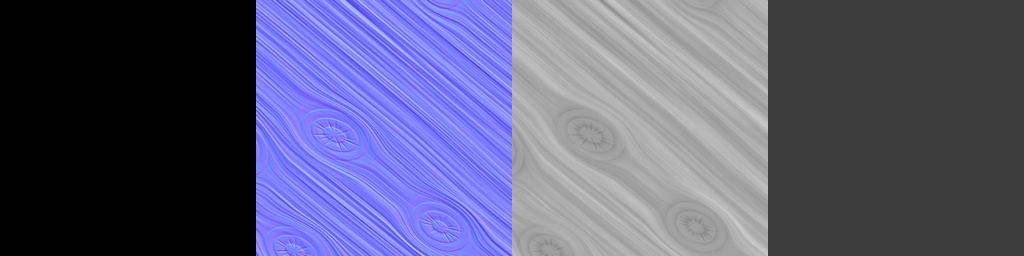} &
		\includegraphics[height=\resLen]{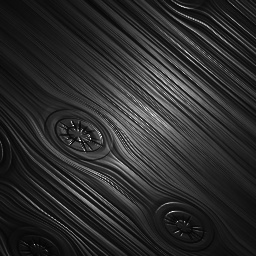} &
		\includegraphics[height=\resLen]{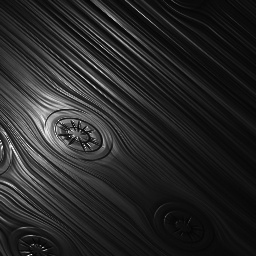} &
		\includegraphics[height=\resLen]{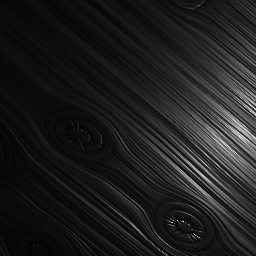}
		\\
		\raisebox{.2in}{\rotatebox[origin=c]{90}{\footnotesize{Ours}}} &
		\includegraphics[height=\resLen]{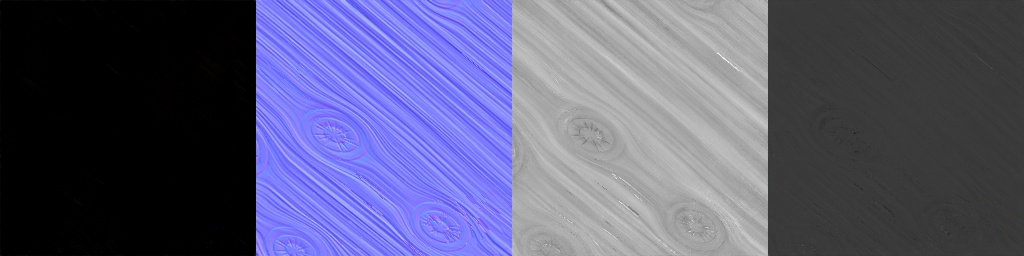} &
		\includegraphics[height=\resLen]{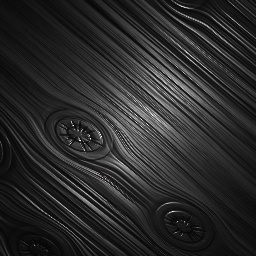} &
		\includegraphics[height=\resLen]{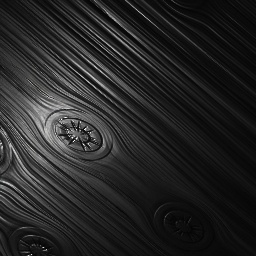} &
		\includegraphics[height=\resLen]{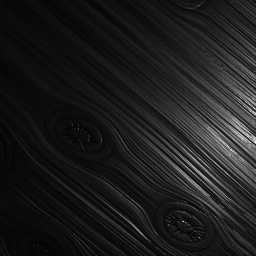}
		\\
		\raisebox{.2in}{\rotatebox[origin=c]{90}{\footnotesize{GT}}} &
		\includegraphics[height=\resLen]{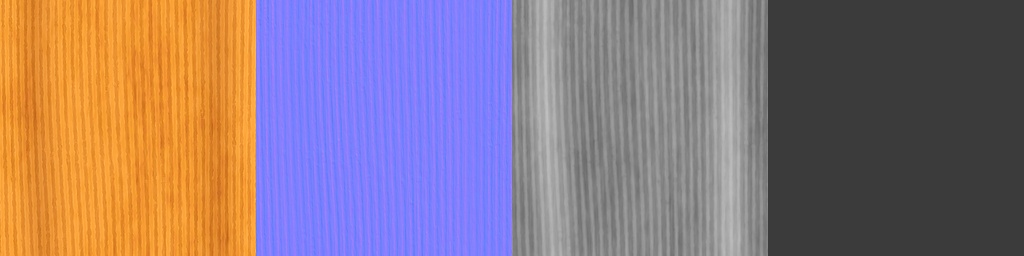} &
		\includegraphics[height=\resLen]{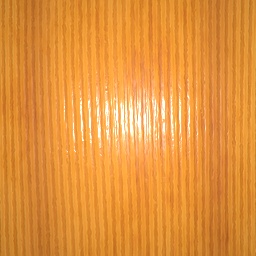} &
		\includegraphics[height=\resLen]{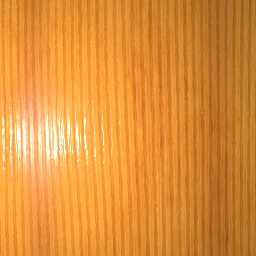} &
		\includegraphics[height=\resLen]{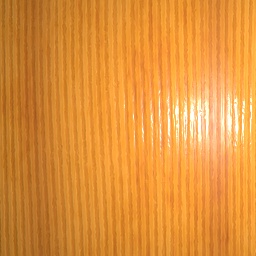}
		\\
		\raisebox{.2in}{\rotatebox[origin=c]{90}{\footnotesize{Ours}}} &
		\includegraphics[height=\resLen]{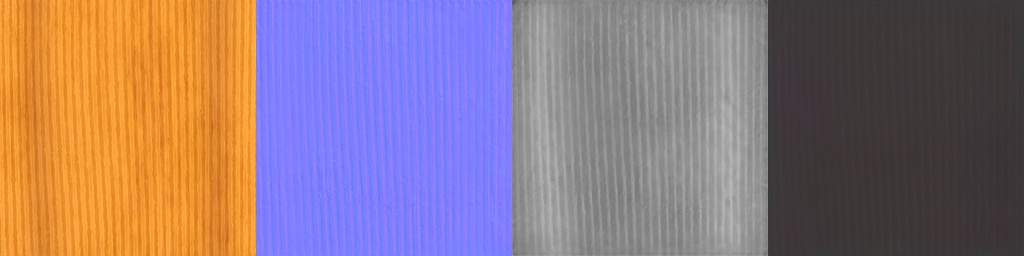} &
		\includegraphics[height=\resLen]{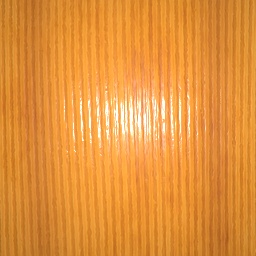} &
		\includegraphics[height=\resLen]{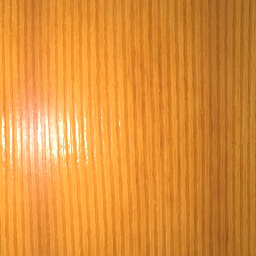} &
		\includegraphics[height=\resLen]{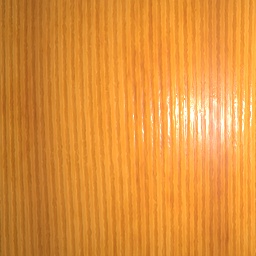}
	\end{tabular}
	\caption{\label{fig:synthetic}
		\textbf{SVBRDF reconstruction on synthetic data.} We demonstrate results on synthetic SVBRDFs, one from \cite{Deschaintre2019} (top) and one from the Adobe Stock Material dataset (bottom). We are able to accurately reconstruct these materials from 7 input images (one input shown). Many more synthetic results are available in supplementary materials.
	}
\end{figure}
\paragraph{Testing on synthetic data.}
Figure \ref{fig:synthetic} contains two synthetic results using our method, showing a close match both in maps and in novel view renderings. For more results, please refer to supplemental materials. \revision{We} note that all methods perform \revision{better} on synthetic data \revision{than on} real data, possibly because of the exact BRDF model match and \revision{perfect} calibration, and also
because the synthetic test set, while distinct from the training set, is relatively similar in style.
%
%
\renewcommand{\one}{real_wall-plaster-white}
\renewcommand{\two}{real_plastic-red-carton}
\renewcommand{\thr}{real_leather-blue}
\renewcommand{\fou}{real_bathroomtile2}
\renewcommand{\fiv}{real_wood-walnut}
\renewcommand{\six}{real_wood-tile}
\renewcommand{\sev}{real_book1}
\renewcommand{\eit}{real_book2}
\renewcommand{\nin}{real_giftbag1}
\renewcommand{\ten}{real_cards-red}

\setlength{\resLen}{.48in}
\begin{figure*}[htbp]
    \centering
    \small
    \addtolength{\tabcolsep}{-4pt}
    \begin{tabular}{rlrccc@{\hspace{2\tabcolsep}}lrccc}
        & \multicolumn{2}{c}{\textbf{SVBRDF maps}} & \textbf{Opt.} & \multicolumn{2}{c}{\textbf{Novel views}}
        & \multicolumn{2}{c}{\textbf{SVBRDF maps}} & \textbf{Opt.} & \multicolumn{2}{c}{\textbf{Novel views}}
        \\[1pt]
        &
        \raisebox{3pt}{\textit{~~wall-plaster-white}} & \raisebox{0.40\resLen}{\rotatebox[origin=c]{90}{\scriptsize GT}}&
        \includegraphics[height=\resLen]{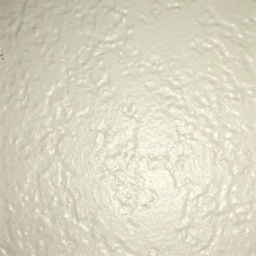} &
        \includegraphics[height=\resLen]{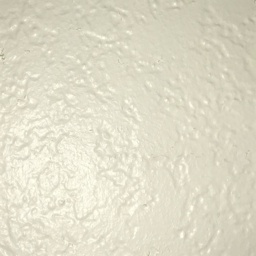} &
        \includegraphics[height=\resLen]{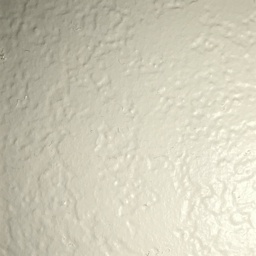} &
        \raisebox{3pt}{\textit{~~plastic-red-carton}} & \raisebox{0.40\resLen}{\rotatebox[origin=c]{90}{\scriptsize GT}}&
        \includegraphics[height=\resLen]{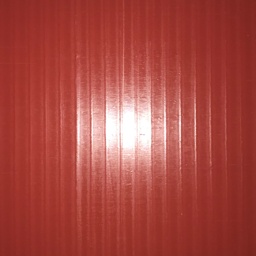} &
        \includegraphics[height=\resLen]{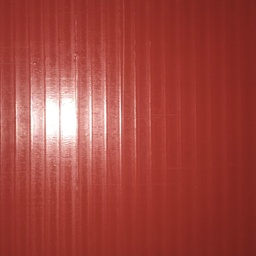} &
        \includegraphics[height=\resLen]{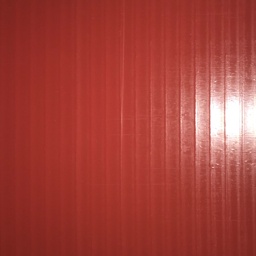}
        \\
        \raisebox{0.40\resLen}{\rotatebox[origin=c]{90}{\scriptsize Ours}} &
        \multicolumn{2}{c}{\includegraphics[height=\resLen]{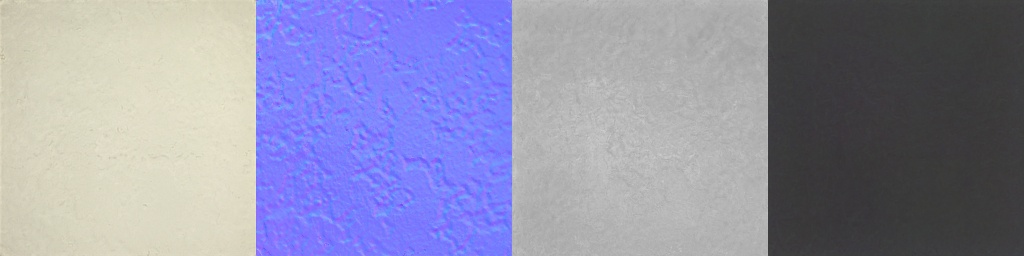}} &
        \includegraphics[height=\resLen]{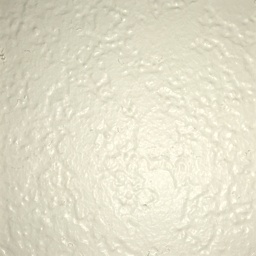} &
        \includegraphics[height=\resLen]{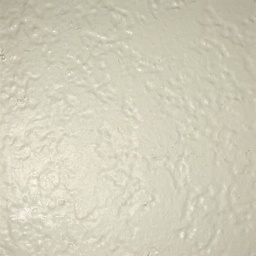} &
        \includegraphics[height=\resLen]{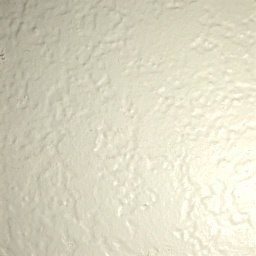} &
        \multicolumn{2}{c}{\includegraphics[height=\resLen]{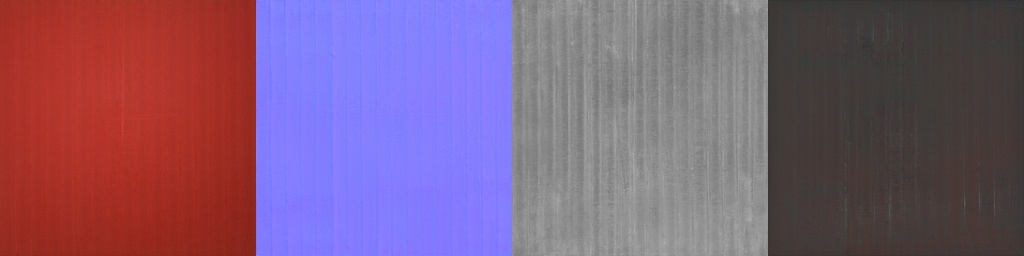}} &
        \includegraphics[height=\resLen]{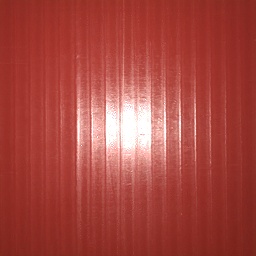} &
        \includegraphics[height=\resLen]{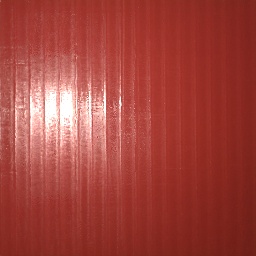} &
        \includegraphics[height=\resLen]{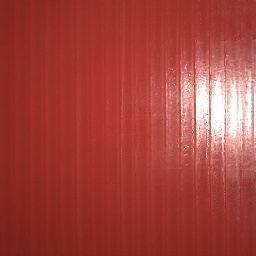}
        \\
        \raisebox{0.40\resLen}{\rotatebox[origin=c]{90}{\scriptsize [Gao19]+}} &
        \multicolumn{2}{c}{\includegraphics[height=\resLen]{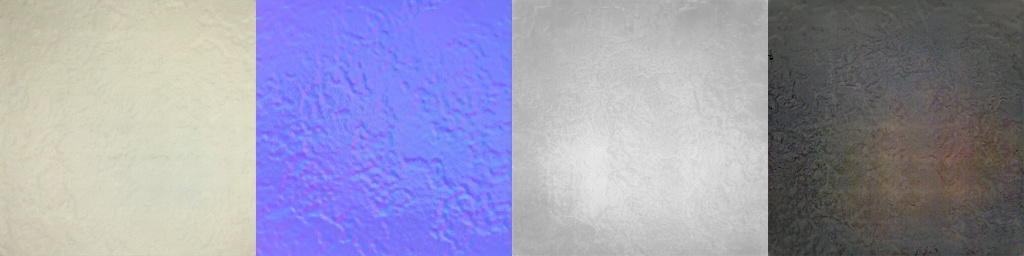}} &
        \includegraphics[height=\resLen]{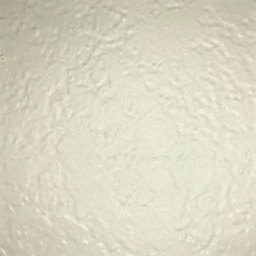} &
        \includegraphics[height=\resLen]{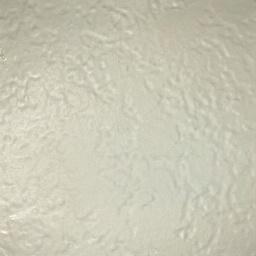} &
        \includegraphics[height=\resLen]{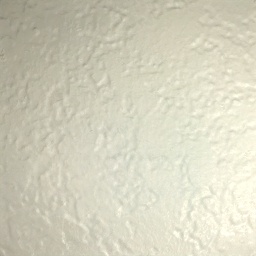} &
        \multicolumn{2}{c}{\includegraphics[height=\resLen]{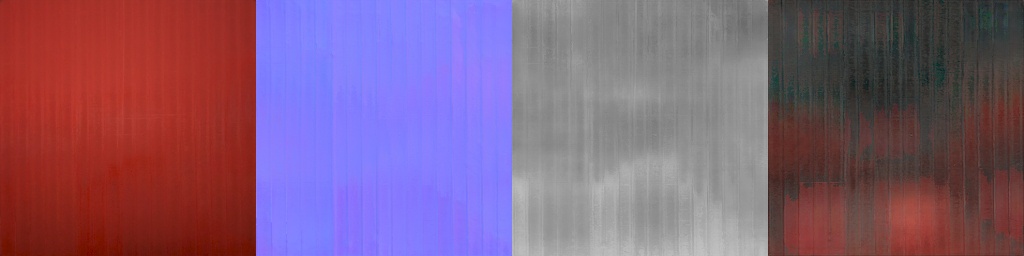}} &
        \includegraphics[height=\resLen]{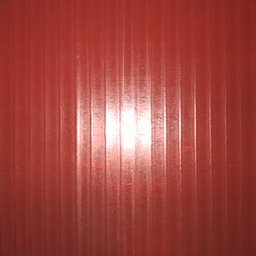} &
        \includegraphics[height=\resLen]{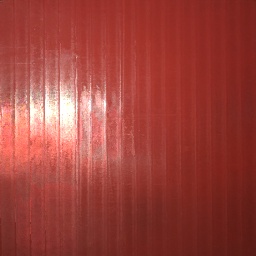} &
        \includegraphics[height=\resLen]{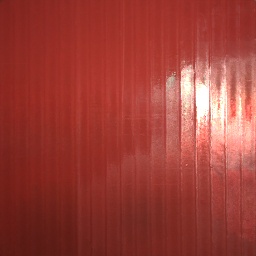}
        \\[1pt]
        &
        \raisebox{3pt}{\textit{~~leather-blue}} & \raisebox{0.40\resLen}{\rotatebox[origin=c]{90}{\scriptsize GT}}&
        \includegraphics[height=\resLen]{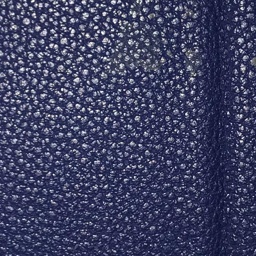} &
        \includegraphics[height=\resLen]{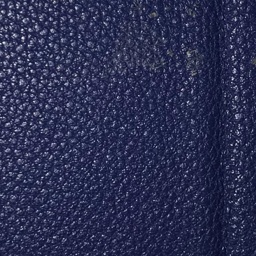} &
        \includegraphics[height=\resLen]{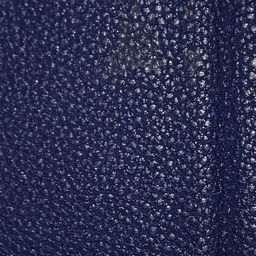} &
        \raisebox{3pt}{\textit{~~bathroomtile2}} & \raisebox{0.40\resLen}{\rotatebox[origin=c]{90}{\scriptsize GT}}&
        \includegraphics[height=\resLen]{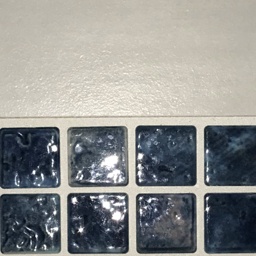} &
        \includegraphics[height=\resLen]{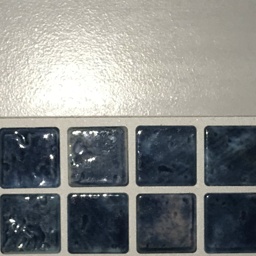} &
        \includegraphics[height=\resLen]{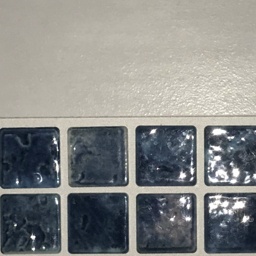}
        \\
        \raisebox{0.40\resLen}{\rotatebox[origin=c]{90}{\scriptsize Ours}} &
        \multicolumn{2}{c}{\includegraphics[height=\resLen]{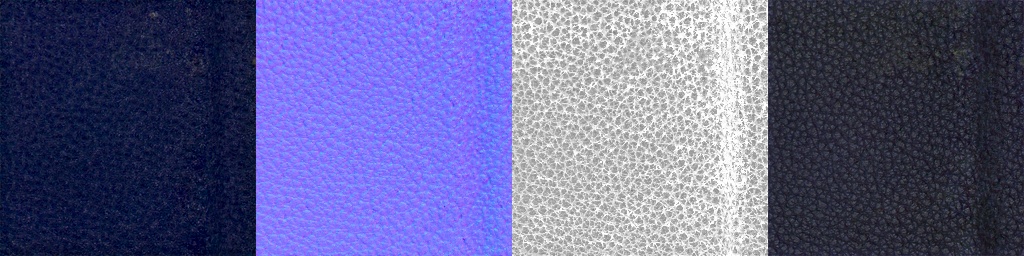}} &
        \includegraphics[height=\resLen]{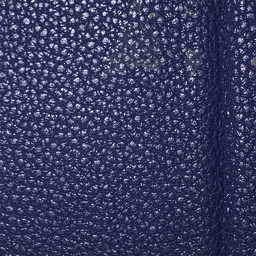} &
        \includegraphics[height=\resLen]{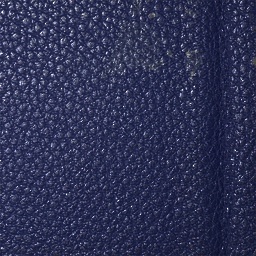} &
        \includegraphics[height=\resLen]{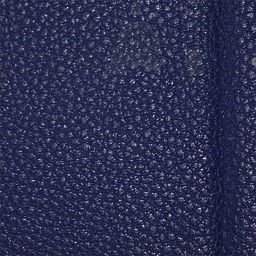} &
        \multicolumn{2}{c}{\includegraphics[height=\resLen]{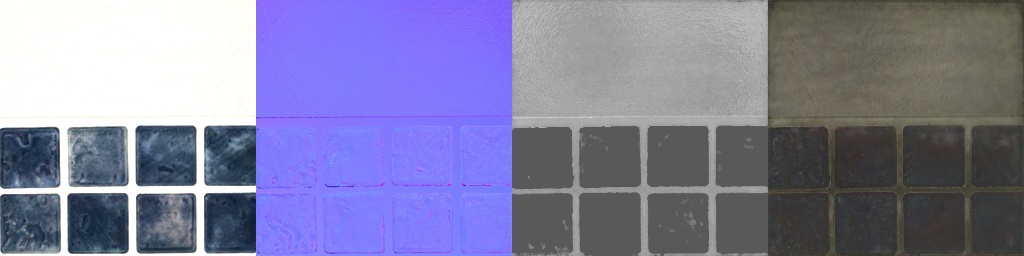}} &
        \includegraphics[height=\resLen]{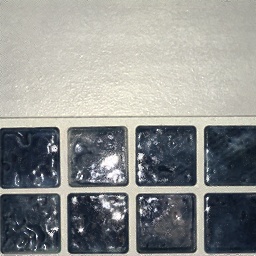} &
        \includegraphics[height=\resLen]{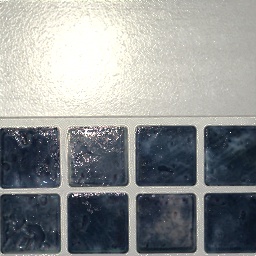} &
        \includegraphics[height=\resLen]{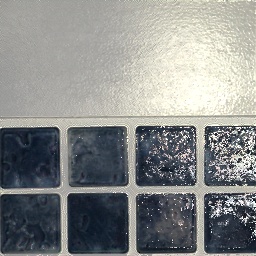}
        \\
        \raisebox{0.40\resLen}{\rotatebox[origin=c]{90}{\scriptsize [Gao19]+}} &
        \multicolumn{2}{c}{\includegraphics[height=\resLen]{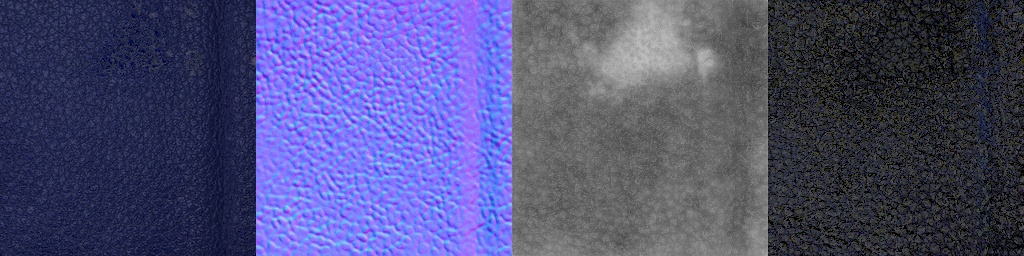}} &
        \includegraphics[height=\resLen]{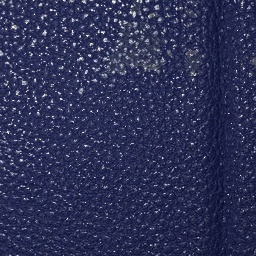} &
        \includegraphics[height=\resLen]{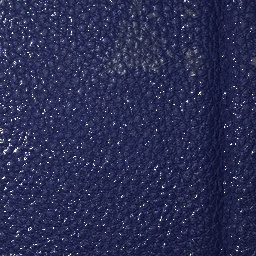} &
        \includegraphics[height=\resLen]{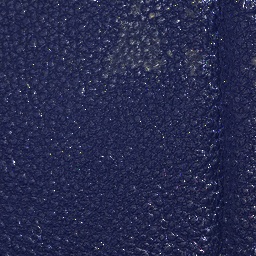} &
        \multicolumn{2}{c}{\includegraphics[height=\resLen]{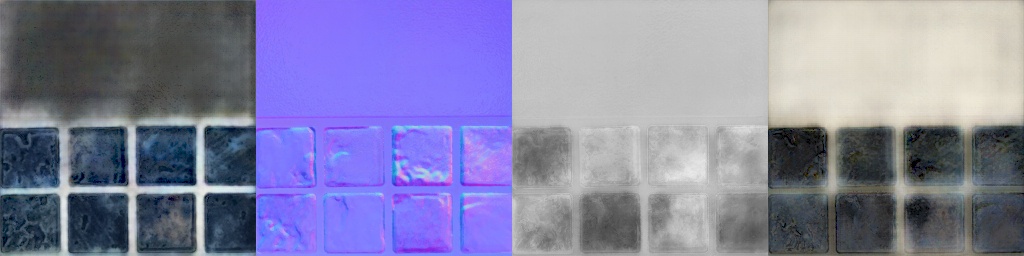}} &
        \includegraphics[height=\resLen]{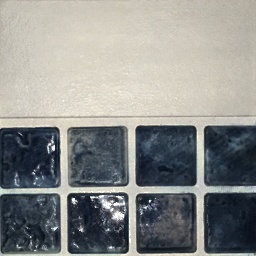} &
        \includegraphics[height=\resLen]{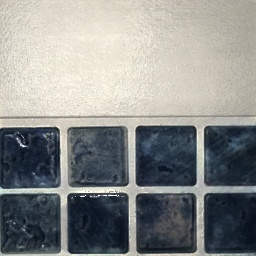} &
        \includegraphics[height=\resLen]{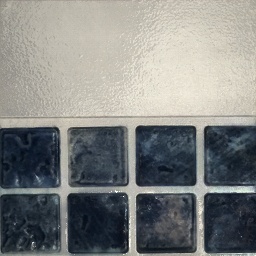}
        \\[1pt]
        &
        \raisebox{3pt}{\textit{~~wood-walnut}} & \raisebox{0.40\resLen}{\rotatebox[origin=c]{90}{\scriptsize GT}}&
        \includegraphics[height=\resLen]{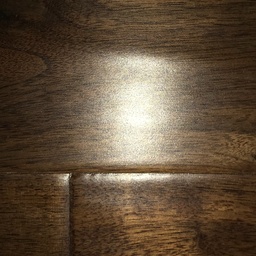} &
        \includegraphics[height=\resLen]{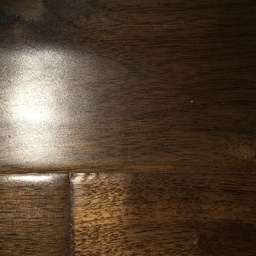} &
        \includegraphics[height=\resLen]{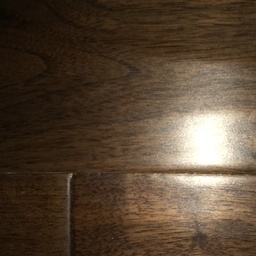} &
        \raisebox{3pt}{\textit{~~wood-tile}} & \raisebox{0.40\resLen}{\rotatebox[origin=c]{90}{\scriptsize GT}}&
        \includegraphics[height=\resLen]{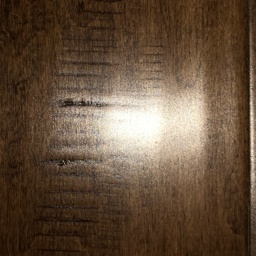} &
        \includegraphics[height=\resLen]{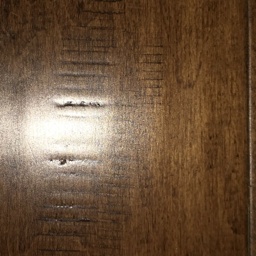} &
        \includegraphics[height=\resLen]{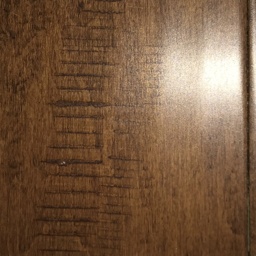}
        \\
        \raisebox{0.40\resLen}{\rotatebox[origin=c]{90}{\scriptsize Ours}} &
        \multicolumn{2}{c}{\includegraphics[height=\resLen]{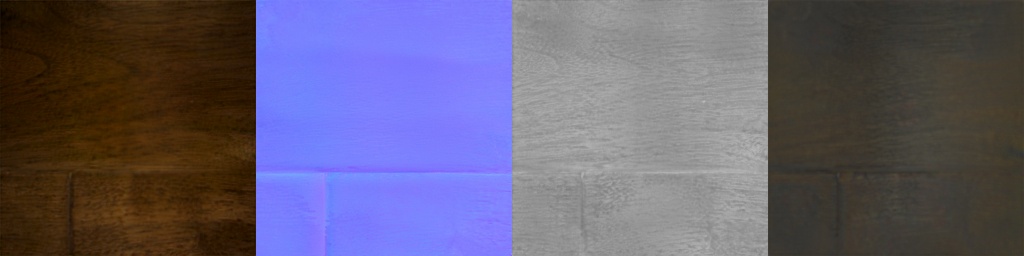}} &
        \includegraphics[height=\resLen]{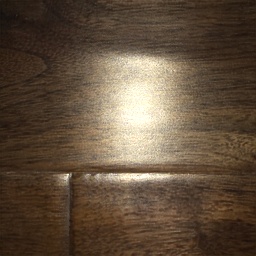} &
        \includegraphics[height=\resLen]{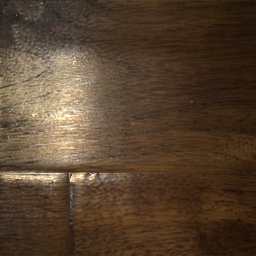} &
        \includegraphics[height=\resLen]{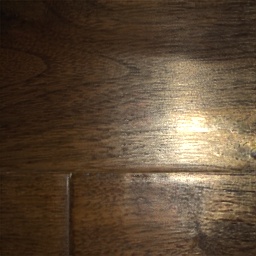} &
        \multicolumn{2}{c}{\includegraphics[height=\resLen]{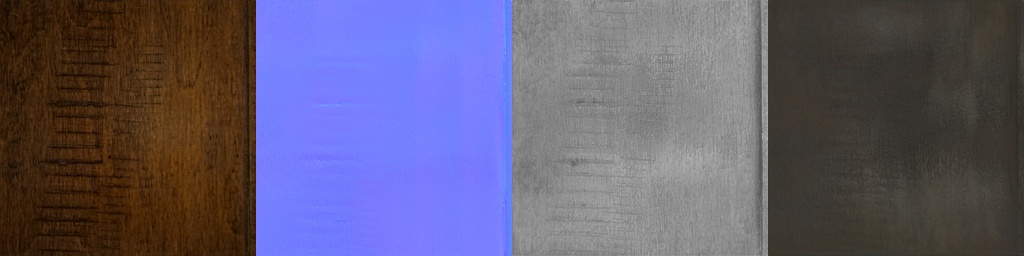}} &
        \includegraphics[height=\resLen]{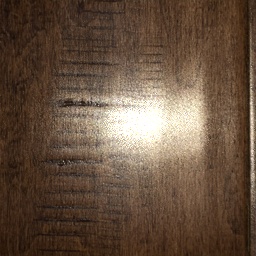} &
        \includegraphics[height=\resLen]{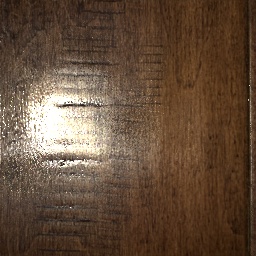} &
        \includegraphics[height=\resLen]{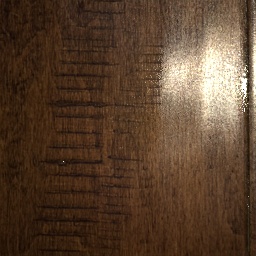}
        \\
        \raisebox{0.40\resLen}{\rotatebox[origin=c]{90}{\scriptsize [Gao19]+}} &
        \multicolumn{2}{c}{\includegraphics[height=\resLen]{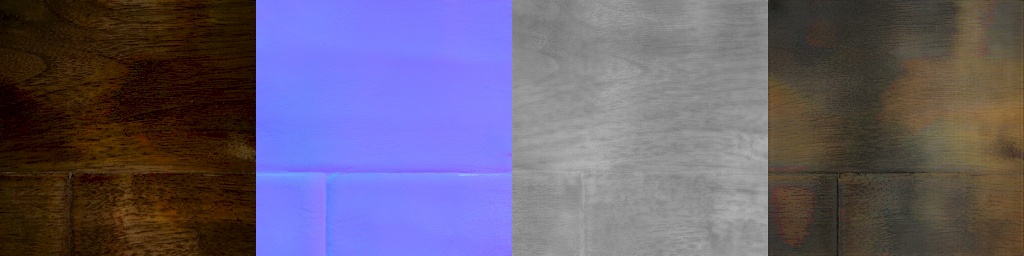}} &
        \includegraphics[height=\resLen]{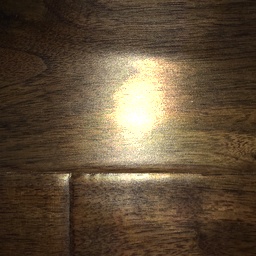} &
        \includegraphics[height=\resLen]{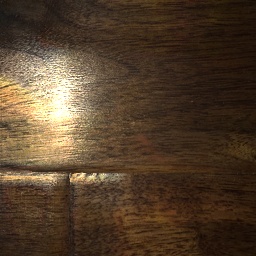} &
        \includegraphics[height=\resLen]{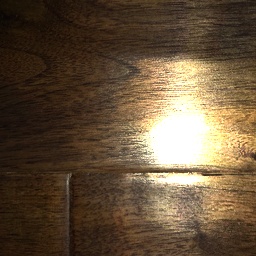} &
        \multicolumn{2}{c}{\includegraphics[height=\resLen]{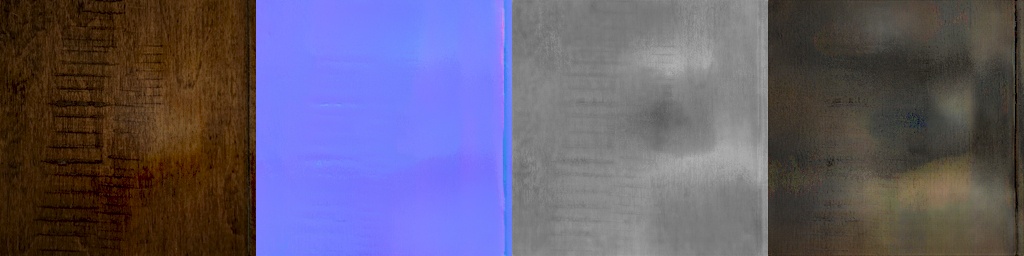}} &
        \includegraphics[height=\resLen]{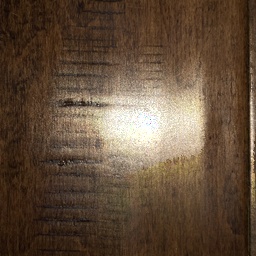} &
        \includegraphics[height=\resLen]{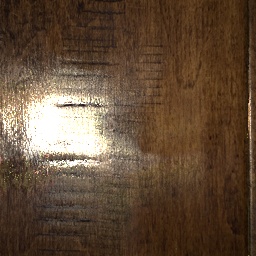} &
        \includegraphics[height=\resLen]{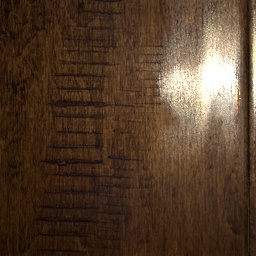}
        \\[1pt]
        &
        \raisebox{3pt}{\textit{~~book1}} & \raisebox{0.40\resLen}{\rotatebox[origin=c]{90}{\scriptsize GT}}&
        \includegraphics[height=\resLen]{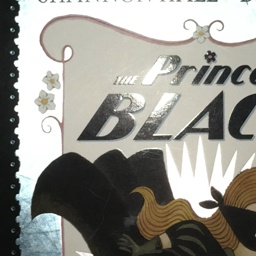} &
        \includegraphics[height=\resLen]{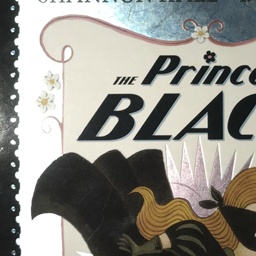} &
        \includegraphics[height=\resLen]{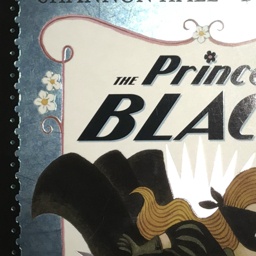} &
        \raisebox{3pt}{\textit{~~book2}} & \raisebox{0.40\resLen}{\rotatebox[origin=c]{90}{\scriptsize GT}}&
        \includegraphics[height=\resLen]{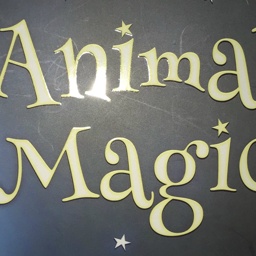} &
        \includegraphics[height=\resLen]{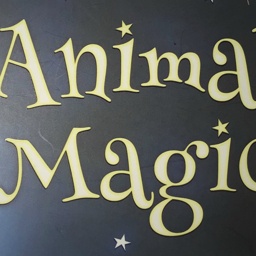} &
        \includegraphics[height=\resLen]{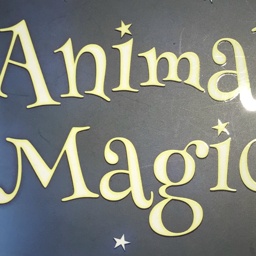}
        \\
        \raisebox{0.40\resLen}{\rotatebox[origin=c]{90}{\scriptsize Ours}} &
        \multicolumn{2}{c}{\includegraphics[height=\resLen]{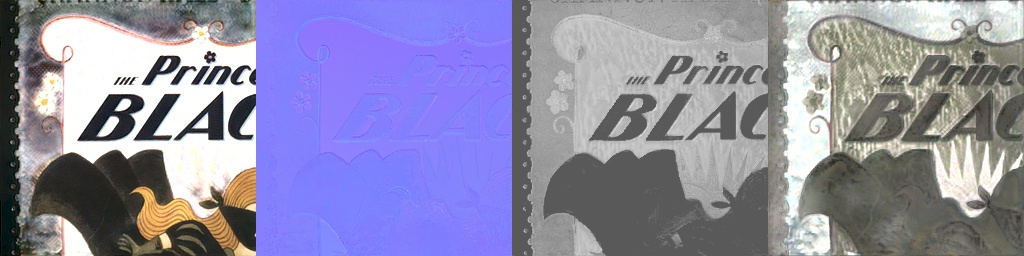}} &
        \includegraphics[height=\resLen]{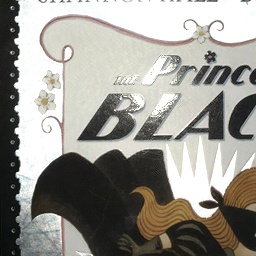} &
        \includegraphics[height=\resLen]{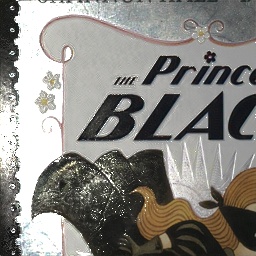} &
        \includegraphics[height=\resLen]{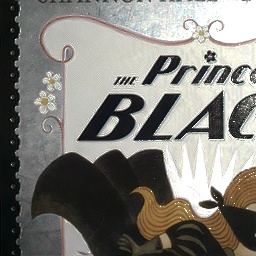} &
        \multicolumn{2}{c}{\includegraphics[height=\resLen]{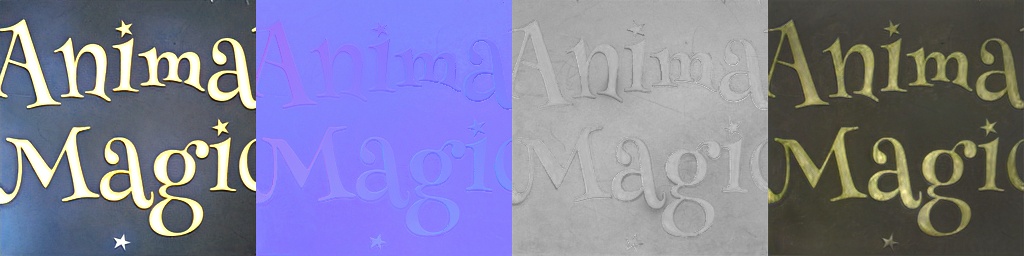}} &
        \includegraphics[height=\resLen]{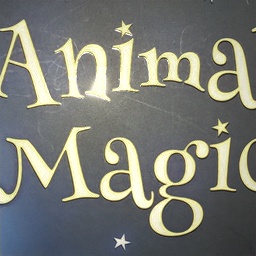} &
        \includegraphics[height=\resLen]{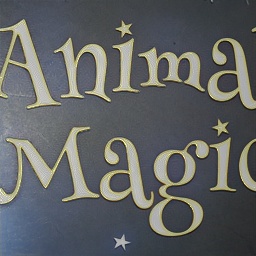} &
        \includegraphics[height=\resLen]{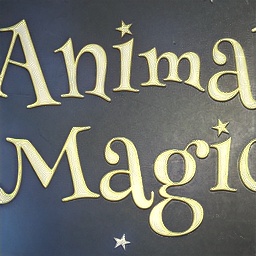}
        \\
        \raisebox{0.40\resLen}{\rotatebox[origin=c]{90}{\scriptsize [Gao19]+}} &
        \multicolumn{2}{c}{\includegraphics[height=\resLen]{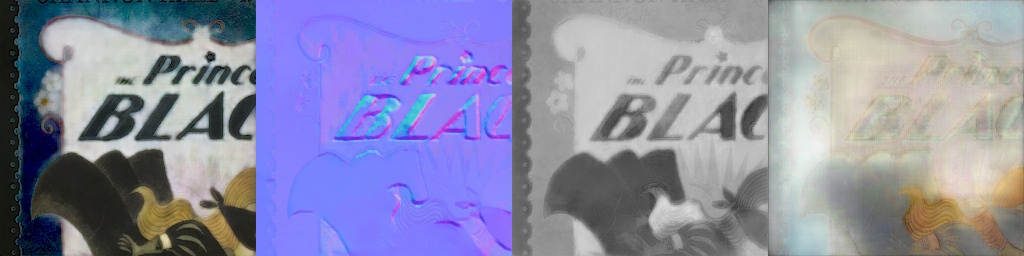}} &
        \includegraphics[height=\resLen]{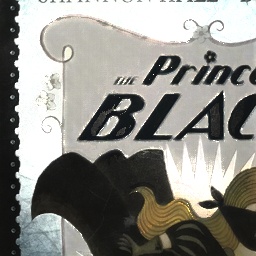} &
        \includegraphics[height=\resLen]{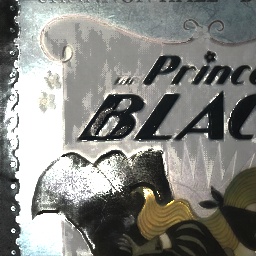} &
        \includegraphics[height=\resLen]{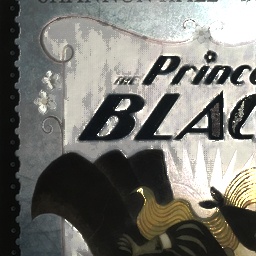} &
        \multicolumn{2}{c}{\includegraphics[height=\resLen]{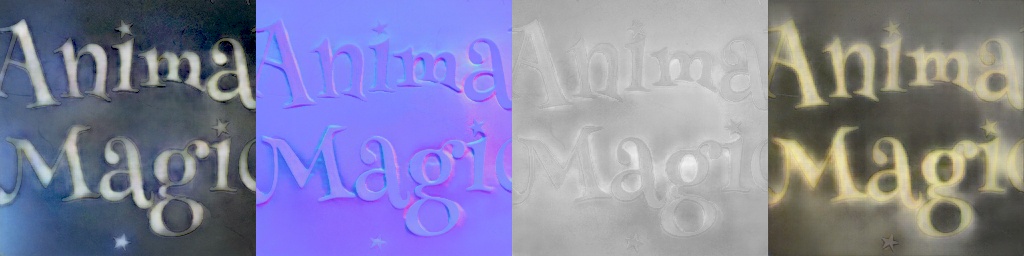}} &
        \includegraphics[height=\resLen]{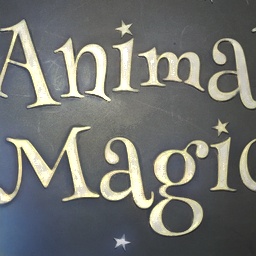} &
        \includegraphics[height=\resLen]{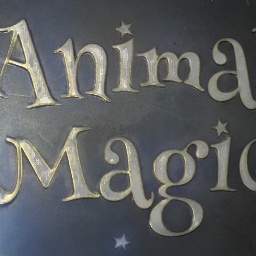} &
        \includegraphics[height=\resLen]{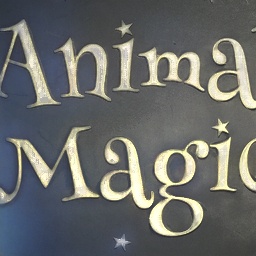}
        \\[1pt]
        &
        \raisebox{3pt}{\textit{~~giftbag1}} & \raisebox{0.40\resLen}{\rotatebox[origin=c]{90}{\scriptsize GT}}&
        \includegraphics[height=\resLen]{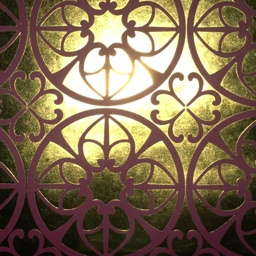} &
        \includegraphics[height=\resLen]{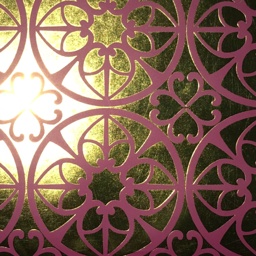} &
        \includegraphics[height=\resLen]{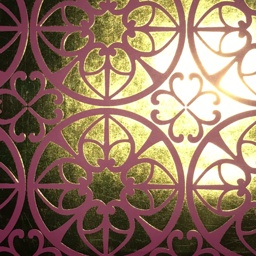} &
        \raisebox{3pt}{\textit{~~cards-red}} & \raisebox{0.40\resLen}{\rotatebox[origin=c]{90}{\scriptsize GT}}&
        \includegraphics[height=\resLen]{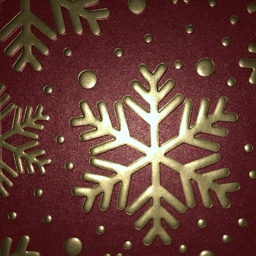} &
        \includegraphics[height=\resLen]{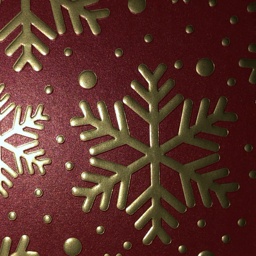} &
        \includegraphics[height=\resLen]{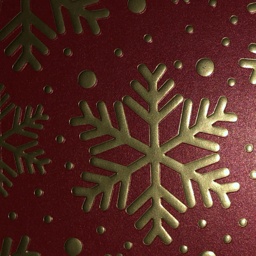}
        \\
        \raisebox{0.40\resLen}{\rotatebox[origin=c]{90}{\scriptsize Ours}} &
        \multicolumn{2}{c}{\includegraphics[height=\resLen]{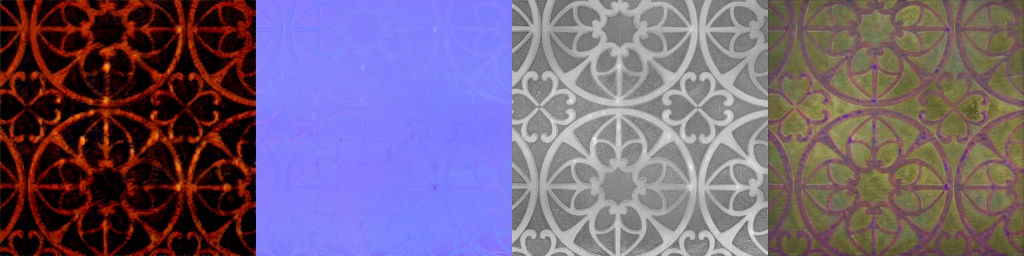}} &
        \includegraphics[height=\resLen]{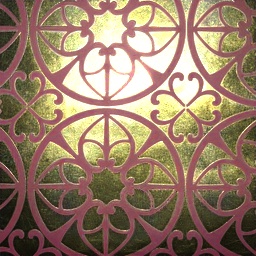} &
        \includegraphics[height=\resLen]{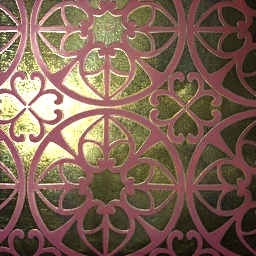} &
        \includegraphics[height=\resLen]{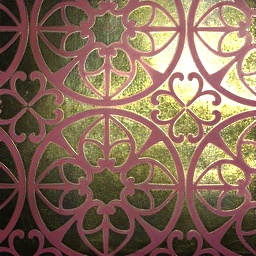} &
        \multicolumn{2}{c}{\includegraphics[height=\resLen]{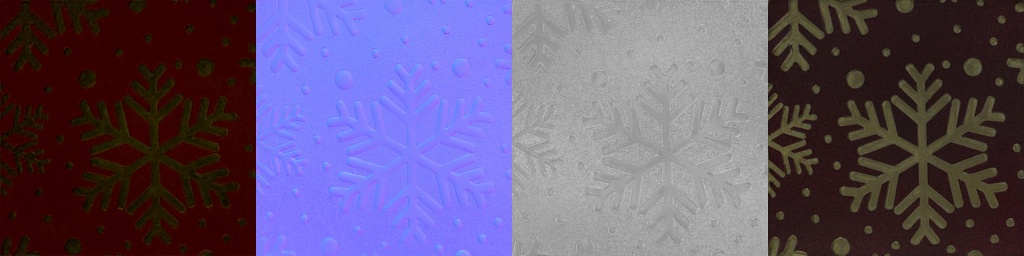}} &
        \includegraphics[height=\resLen]{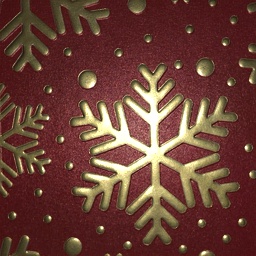} &
        \includegraphics[height=\resLen]{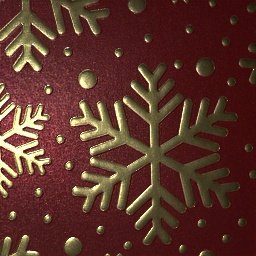} &
        \includegraphics[height=\resLen]{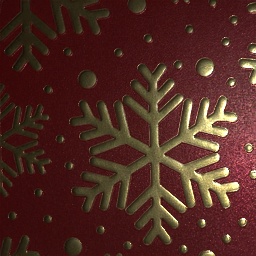}
        \\
        \raisebox{0.40\resLen}{\rotatebox[origin=c]{90}{\scriptsize [Gao19]+}} &
        \multicolumn{2}{c}{\includegraphics[height=\resLen]{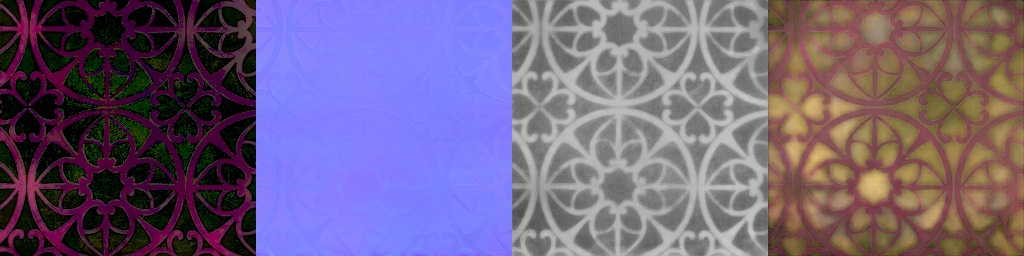}} &
        \includegraphics[height=\resLen]{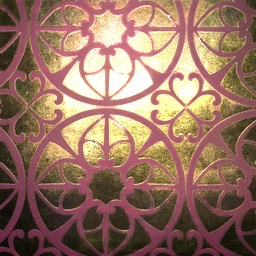} &
        \includegraphics[height=\resLen]{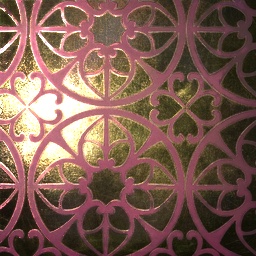} &
        \includegraphics[height=\resLen]{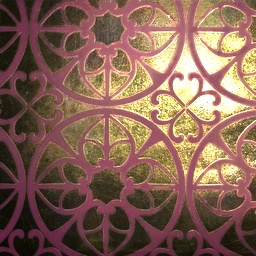} &
        \multicolumn{2}{c}{\includegraphics[height=\resLen]{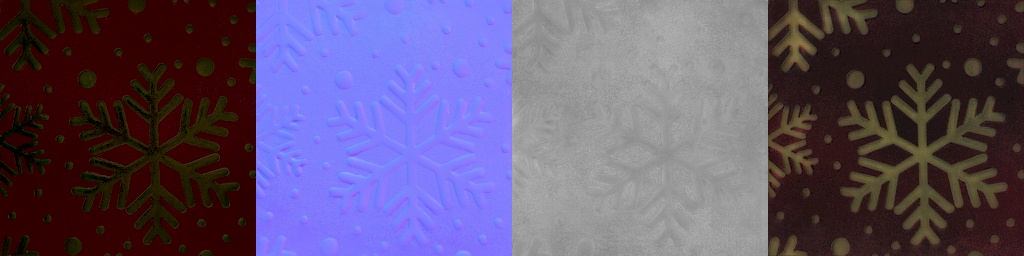}} &
        \includegraphics[height=\resLen]{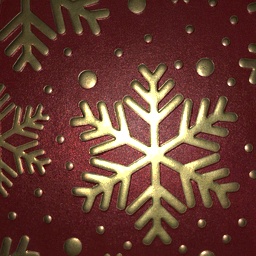} &
        \includegraphics[height=\resLen]{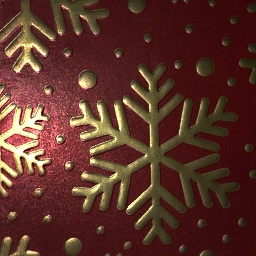} &
        \includegraphics[height=\resLen]{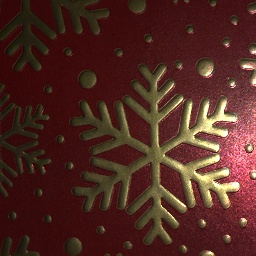}
    \end{tabular}
    \caption{\label{fig:real}
        \textbf{SVBRDF reconstruction on real data.}
        We reconstruct SVBRDF maps from 7 inputs, and compare the resulting maps and images rendered under 2 novel views.
        Gao's method~\shortcite{Gao2019} initialized with Deschaintre's~\shortcite{Deschaintre2019} direct predictions (denoted as ``[Gao19]+'') tends to have complex reflectance burnt into the specular albedo map, leading to inaccurate predictions under novel views.
        Our method with simple initializations, in contrast, is less prone to such burn-ins and generally produces more accurate renderings under novel views.
        Please refer to Table~\ref{tab:main} for more information on the quality of these renderings.
    }
\end{figure*}

\begin{figure*}[!t]
	\setlength{\resLen}{1.76in}
	\small
	\addtolength{\tabcolsep}{-4pt}
	\begin{tabular}{c|c}
		\textbf{Synthetic data} & \textbf{Real data}\\
		\includegraphics[height=\resLen]{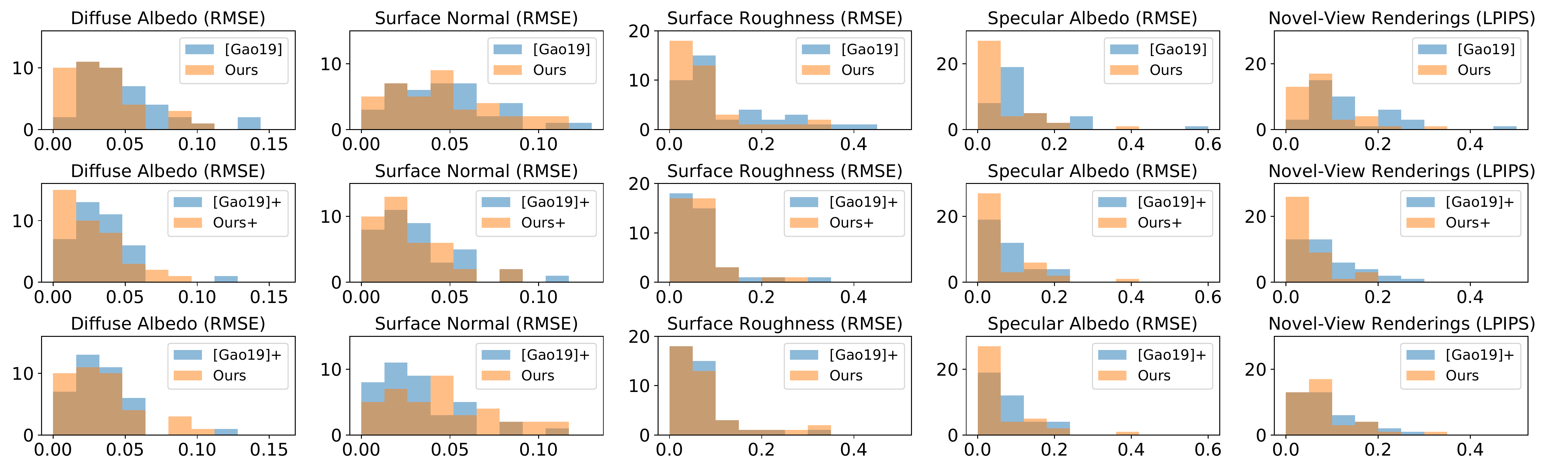} &
		\includegraphics[height=\resLen]{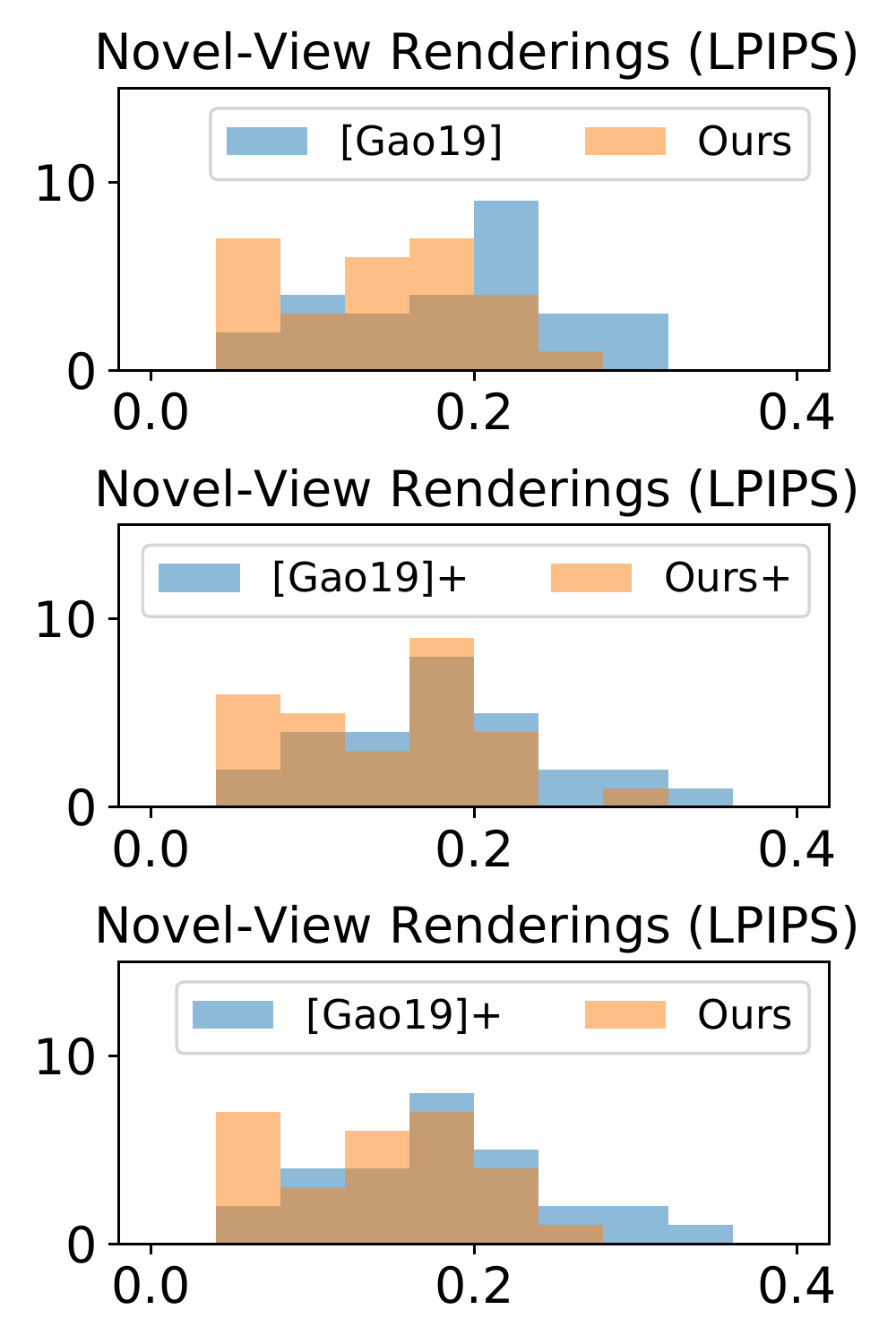}\\[-4pt]
	\end{tabular}
	\caption{\label{fig:rmse}
		\textbf{Performance statistics} of Gao~\shortcite{Gao2019} and our method.
		For each technique, we compute (i)~the Learned Perceptual Image Patch Similarity (LPIPS) metric between renderings of the output SVBRDF maps and the reference images for \totReal \emph{real} and \totSynthetic \emph{synthetic} examples; and (ii)~the root-measure-square error (RMSE) of the inferred maps for the \emph{synthetic} examples.
		For both metrics, a lower score indicates a better accuracy.
		Using identical initializations, our technique (``Ours'' and ``Ours+'') outperforms Gao's (``[Gao19]'' and ``[Gao19]+'') consistently for both real and synthetic examples, as demonstrated in the top and the middle row.
		Furthermore, our technique with constant initializations (``Ours'') has a similar performance with Gao's method initialized using Deschaintre's~\shortcite{Deschaintre2019} direct predictions (``[Gao19]+'') on the synthetic examples and outperforms the latter on the real examples, as shown on the bottom.
	}
\end{figure*}

\subsection{Comparison with prior work on real data}
\label{ssec:real}
Here we compare our method and Gao et al. \shortcite{Gao2019}. For more results and comparisons, including with Deschaintre et al. \shortcite{Deschaintre2019}, and including with and without initialization for ours and Gao's method, please refer to supplemental materials.
We show 10 real examples from our cell phone capture pipeline in Figure \ref{fig:real}. Note that Gao's method is significantly dependent on initialization, while the same is not true for our method. Therefore, in this figure, we show Gao's result \emph{with initialization} by Deschaintre et al. \shortcite{Deschaintre2019}, while our result is shown \emph{without initialization}.
Furthermore, note that we are initializing Gao's method with the 2019 multi-input method by Deschaintre, which is a better initialization than the 2018 single-input method. Thus the baseline we are comparing against is, strictly speaking, even higher than what is published in Gao et al., and combines the two best methods published at this time.
Generally, we find that our method produces cleaner maps and is less prone to overfitting (burn-in) than Gao's, while producing more accurate re-renderings under original and novel lighting. Table~\ref{tab:main} shows a quantitative evaluation of the re-rendering quality on novel lighting. As these novel views would be hard to match pixel-wise using any method, as they have never been observed, we use a perceptual method, specifically the Learned Perceptual Image Patch Similarity (LPIPS) metric \cite{LPIPS} (lower is better). Note that our method (without initialization by Deschaintre's method) produces better scores for novel views than Gao's method (with initialization) for most images; even in the case where our LPIPS score is worse, our maps still look more plausible overall.
We also report quantitative evaluations (histograms) for our entire set of results (see Figure \ref{fig:rmse}). For synthetic data, we compare the RMSE of all predicted maps (diffuse albedo, normal, roughness, specular albedo), as we do know the ground truth for them. For both synthetic and real data, we compare the LPIPS scores on novel lighting. We use a + sign to indicate initialization by Deschaintre et al. In the top row, we compare both methods without initialization by Deschaintre's method, while in the middle row, both methods are initialized, and in the bottom row, we compare our method without initialization to Gao's with initialization. Generally, we find that if both methods are initialized the same way, our method outperforms Gao's. Even in the last row, our performance is comparable on synthetic data (worse on normal map and better on diffuse/specular maps) and still better on real data overall.
%
%
\begin{table}[t]
    \centering
    \small
    \addtolength{\tabcolsep}{-3pt}
    \caption{\label{tab:main}
    	Accuracy of the novel-view renderings shown in Figure~\ref{fig:real} measured using the Learned Perceptual Image Patch Similarity (LPIPS) metric where our method produces better predictions than Gao's~\shortcite{Gao2019} in most cases.
    }
    \begin{tabular}{rccrcc}
        \textbf{Material} & \textbf{Ours} & \textbf{[Gao19]+} & \textbf{Material} & \textbf{Ours} & \textbf{[Gao19]+}\\
        \toprule
        \textit{wall-plaster-white} & \textbf{0.071} & 0.132 & \textit{plastic-red-carton} & \textbf{0.095} & 0.166 \\
        \textit{leather-blue} & \textbf{0.146} & 0.356 & \textit{bathroomtile2} & \textbf{0.225} & 0.231 \\
        \textit{wood-walnut} & \textbf{0.226} & 0.252 & \textit{wood-tile} & 0.202 & \textbf{0.192} \\
        \textit{book1} & \textbf{0.147} & 0.318 & \textit{book2} & \textbf{0.042} & 0.122 \\
        \textit{giftbag1} & \textbf{0.183} & 0.218 & \textit{cards-red} & \textbf{0.059} & 0.092 \\
        \bottomrule
    \end{tabular}
\end{table}

\paragraph{Note about Deschaintre et al.} We find that the results from \cite{Deschaintre2019} have much less accurate re-rendering than either ours or Gao's method, as they are not doing any optimization to precisely fit the target images. The mismatches we observe are definitely not due to simple scaling or gamma correction issues, as that would be consistent across examples; rather, we find that the method performs much better on synthetic examples that match the visual style of its training set. On the other hand, their method is fast and results tend to be clean and artifact-free, so they are very suitable for initialization of optimization methods.
\subsection{Additional comparisons}
\renewcommand{\one}{fake_030}
\renewcommand{\two}{real_other-bamboo-veawe}

\setlength{\resLen}{0.556in}
\begin{figure*}[tbhp]
	\addtolength{\tabcolsep}{-4pt}
	\begin{tabular}{lcccc@{\hspace{2\tabcolsep}}ccc}
		& & \textbf{\small SVBRDF maps} &
		\multicolumn{2}{c}{\textbf{\small Novel views}}
		& \textbf{\small SVBRDF maps} & 
		\multicolumn{2}{c}{\textbf{\small Novel views}}
		\\
		& \raisebox{.25in}{\rotatebox[origin=c]{90}{\footnotesize{GT}}} &
		\includegraphics[height=\resLen]{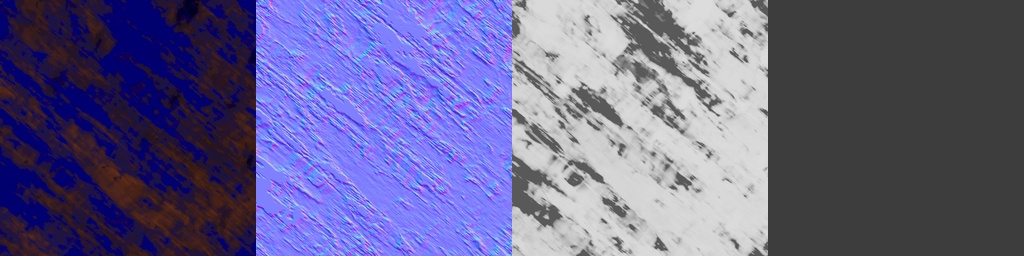} &
		\includegraphics[height=\resLen]{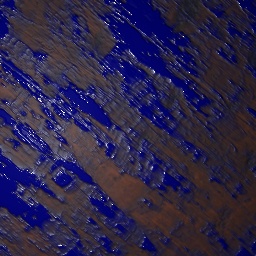} &
		\includegraphics[height=\resLen]{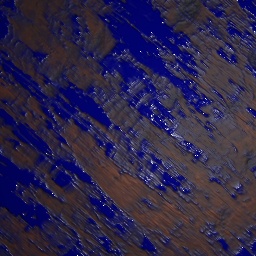} &
		 &
		\includegraphics[height=\resLen]{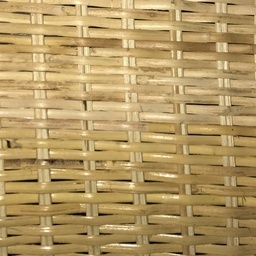} &
		\includegraphics[height=\resLen]{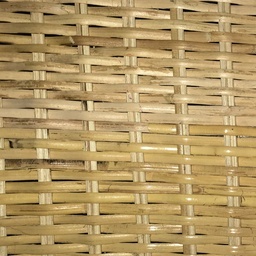}
		\\
		& \raisebox{.25in}{\rotatebox[origin=c]{90}{\footnotesize{[Deschaintre]}}} &
		\includegraphics[height=\resLen]{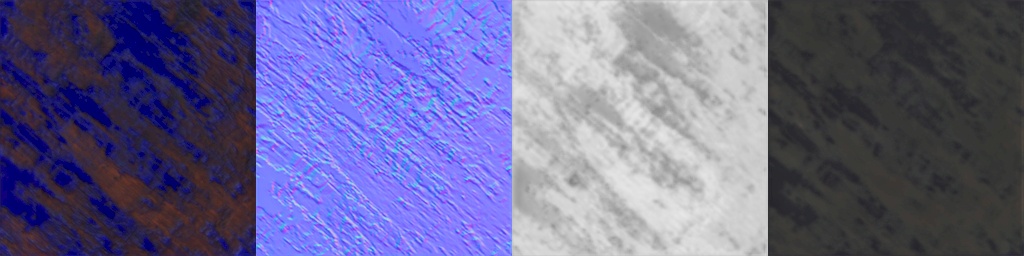} &
		\includegraphics[height=\resLen]{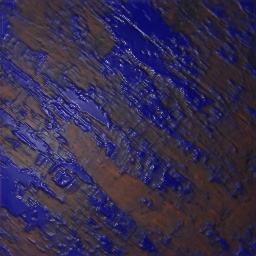} &
		\includegraphics[height=\resLen]{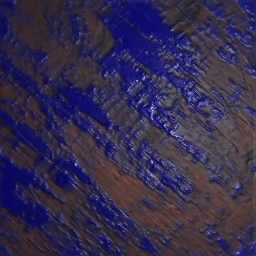} &
		\includegraphics[height=\resLen]{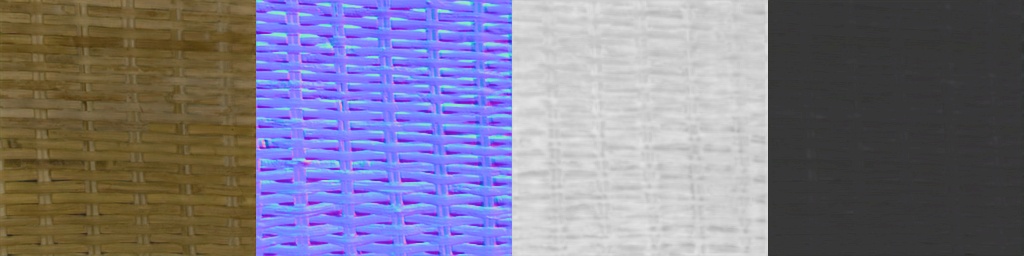} &
		\includegraphics[height=\resLen]{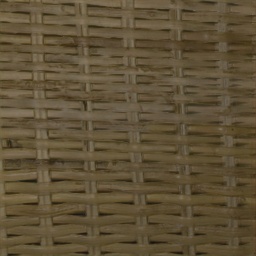} &
		\includegraphics[height=\resLen]{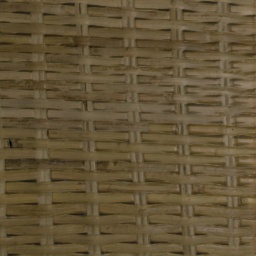}
		\\
		\hline\\[-8pt]
		\multirow{2}{*}[1em]{\rotatebox[origin=c]{90}{\footnotesize\bfseries Constant init.}} &
		\raisebox{.25in}{\rotatebox[origin=c]{90}{\footnotesize{Ours}}} &
		\includegraphics[height=\resLen]{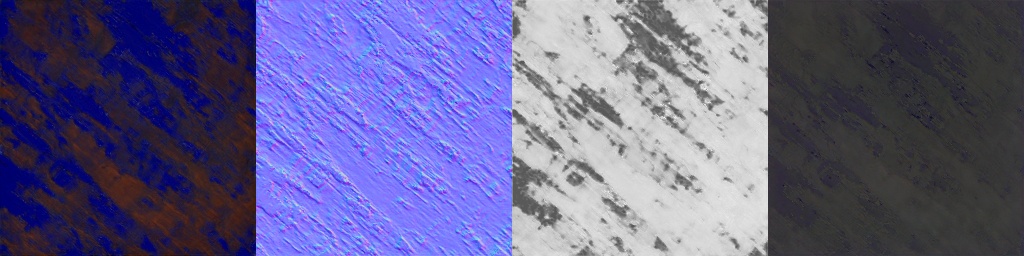} &
		\includegraphics[height=\resLen]{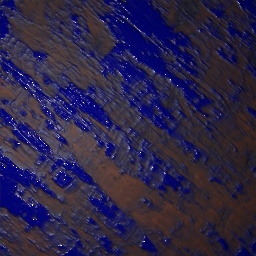} &
		\includegraphics[height=\resLen]{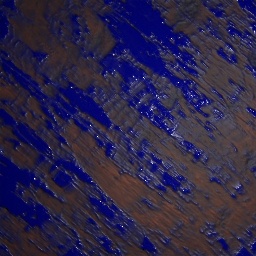} &
		\includegraphics[height=\resLen]{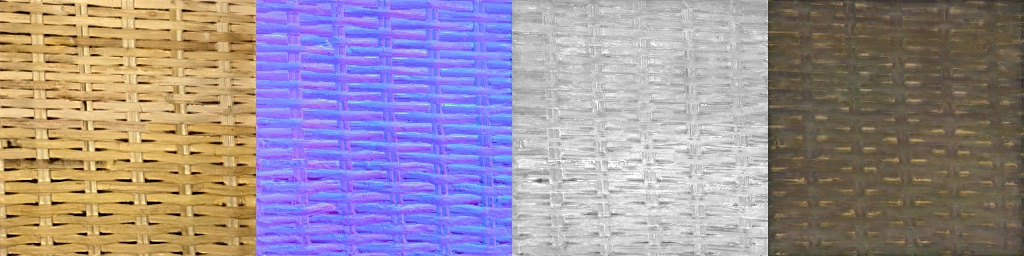} &
		\includegraphics[height=\resLen]{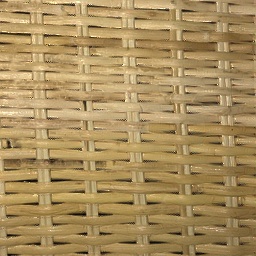} &
		\includegraphics[height=\resLen]{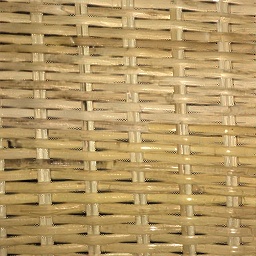}
		\\
		& \raisebox{.25in}{\rotatebox[origin=c]{90}{\footnotesize{[Gao19]}}} &
		\includegraphics[height=\resLen]{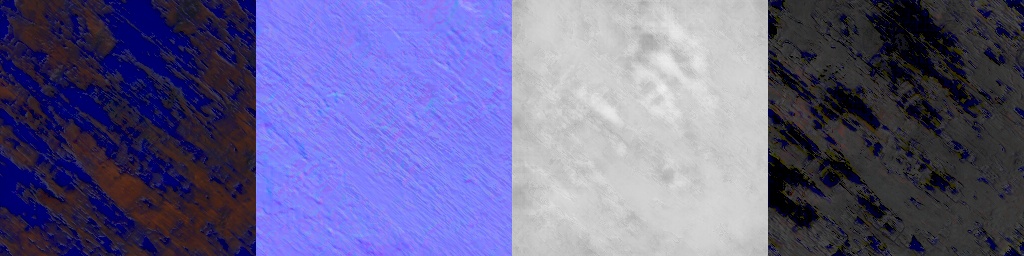} &
		\includegraphics[height=\resLen]{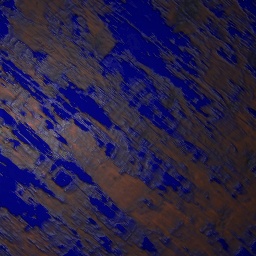} &
		\includegraphics[height=\resLen]{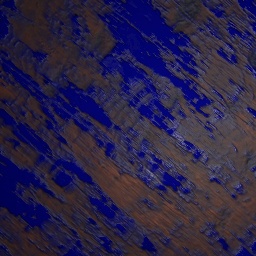} &
		\includegraphics[height=\resLen]{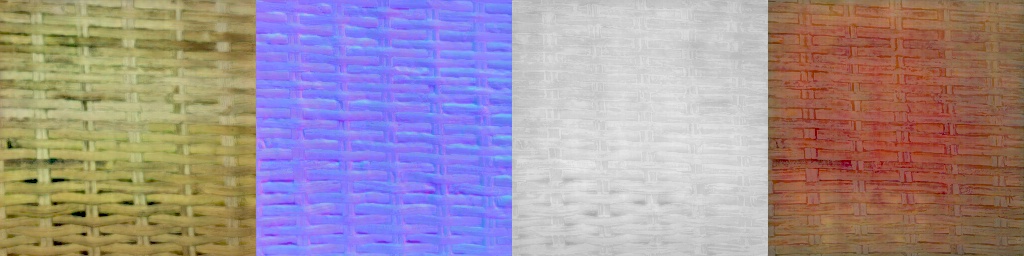} &
		\includegraphics[height=\resLen]{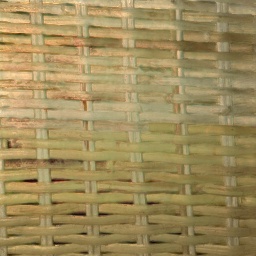} &
		\includegraphics[height=\resLen]{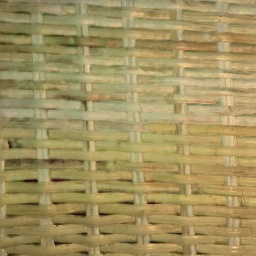}
		\\
		\hline\\[-8pt]
		\multirow{2}{*}[3.15em]{\rotatebox{90}{\footnotesize\bfseries [Deschaintre]-based init.}} &
		\raisebox{.25in}{\rotatebox[origin=c]{90}{\footnotesize{Ours+}}} &
		\includegraphics[height=\resLen]{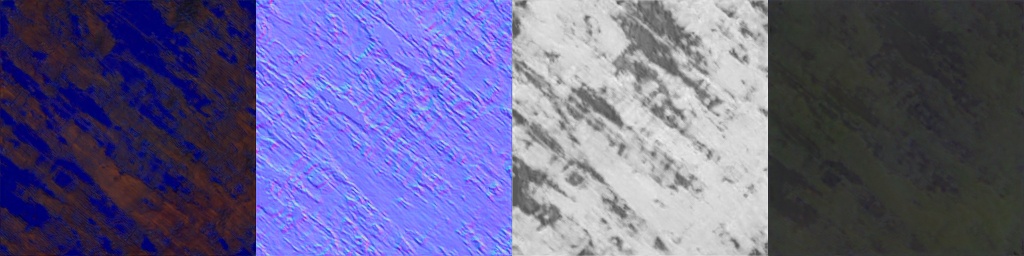} &
		\includegraphics[height=\resLen]{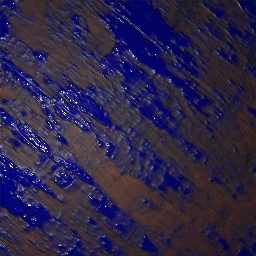} &
		\includegraphics[height=\resLen]{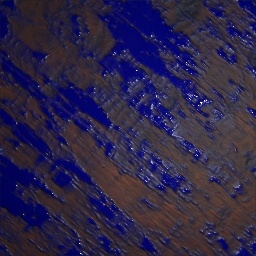} &
		\includegraphics[height=\resLen]{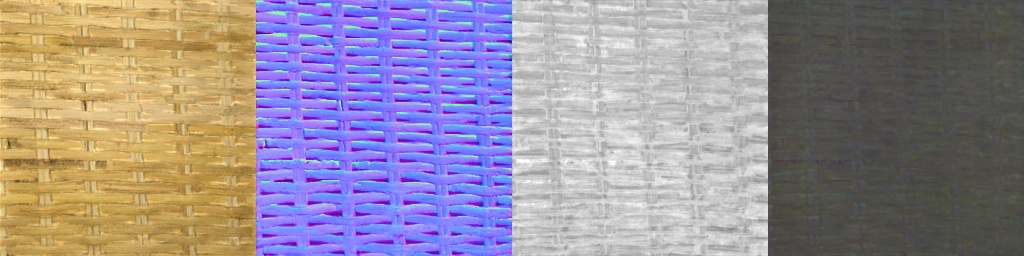} &
		\includegraphics[height=\resLen]{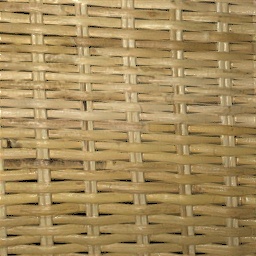} &
		\includegraphics[height=\resLen]{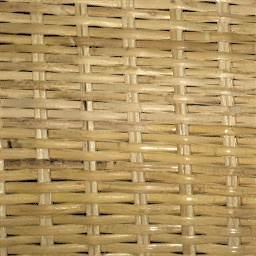}
		\\
		& \raisebox{.25in}{\rotatebox[origin=c]{90}{\footnotesize{[Gao19]+}}} &
		\includegraphics[height=\resLen]{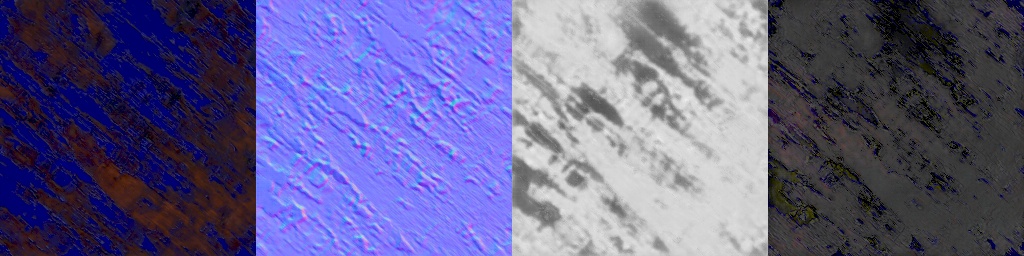} &
		\includegraphics[height=\resLen]{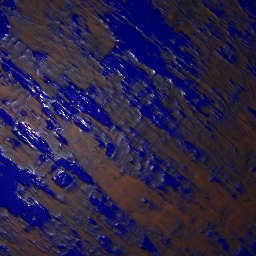} &
		\includegraphics[height=\resLen]{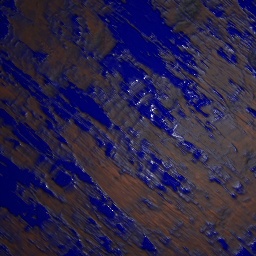} &
		\includegraphics[height=\resLen]{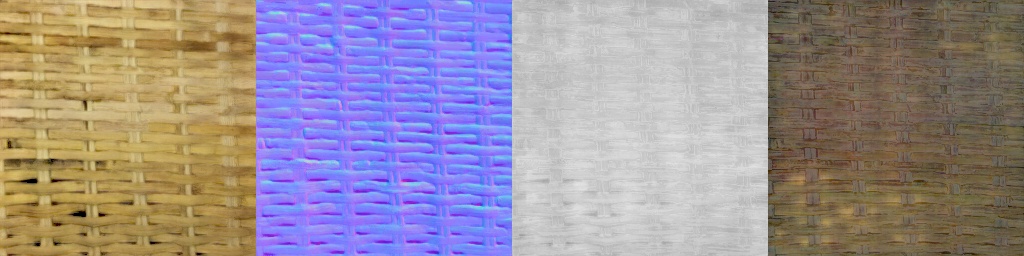} &
		\includegraphics[height=\resLen]{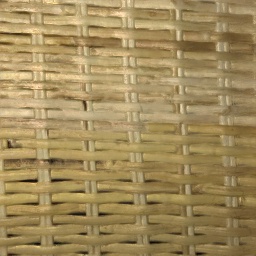} &
		\includegraphics[height=\resLen]{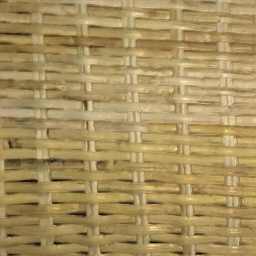}
		\\
	\end{tabular}
	\caption{\label{fig:result_init}
		\textbf{SVBRDF results with different initialization} Unlike Gao's method, ours is less strongly dependent on a good initialization from Deschaintre's method~\shortcite{Deschaintre2019}. In most of cases, starting from simple texture maps (given by our constant initializations) is already good enough to converge to a clean solution. We show all combinations (with and without good initializations) for both methods, for one synthetic and one real example, where techniques initialized with [Deschaintre] are denoted with the suffix ``+'' (i.e., ``Ours+'' and ``[Gao19]+''). Note the failure of Gao's method without good initializations (i.e., ``Gao19'').}
\end{figure*}

\paragraph{Optimization with different initializations.}
In Figure \ref{fig:result_init}, we compare our method to Gao's with and without initialization by Deschaintre's method in all 4 combinations, on a synthetic and a real example. This shows that Gao's method more significantly dependent on good initialization that ours (even though our method can still occasionally benefit).
%
%
\renewcommand{\one}{real_wood-knotty}
\renewcommand{\two}{real_cards-blue}

\setlength{\resLen}{0.558in}
\begin{figure*}[t]
    \centering
    \small
    \addtolength{\tabcolsep}{-4pt}
    \begin{tabular}{rrlrcc@{\hspace{2\tabcolsep}}lrcc}
    	&
        & \multicolumn{2}{c}{\textbf{SVBRDF maps}} & \multicolumn{2}{c}{\textbf{Novel views}}
        & \multicolumn{2}{c}{\textbf{SVBRDF maps}} & \multicolumn{2}{c}{\textbf{Novel views}}
        \\[2pt]
        & &
        & \raisebox{0.40\resLen}{\rotatebox[origin=c]{90}{\scriptsize GT}} &
        \includegraphics[height=\resLen]{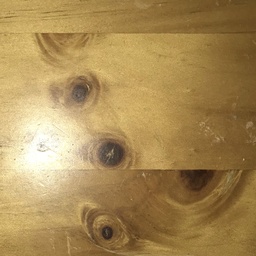} &
        \includegraphics[height=\resLen]{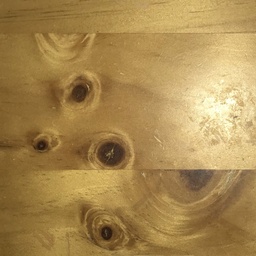} &
        & \raisebox{0.40\resLen}{\rotatebox[origin=c]{90}{\scriptsize GT}} &
        \includegraphics[height=\resLen]{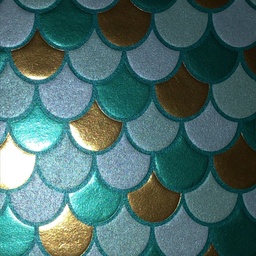} &
        \includegraphics[height=\resLen]{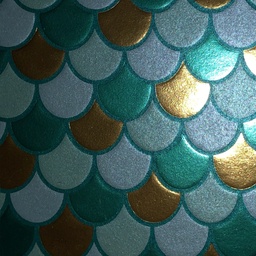}
        \\[-1pt]
        & &
        \textit{~~wood-knotty} & &
        \includegraphics[height=0.5\resLen]{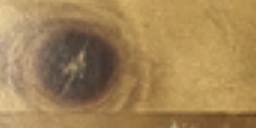} &
        \includegraphics[height=0.5\resLen]{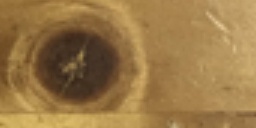} &
        \textit{~~cards-blue} & &
        \includegraphics[height=0.5\resLen]{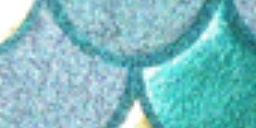} &
        \includegraphics[height=0.5\resLen]{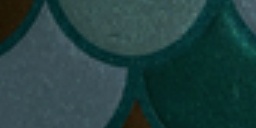}
        \\[2pt]
        \multirow{4}{*}[-1.4em]{\rotatebox[origin=c]{90}{\scriptsize\bfseries No refinement}} &
        \raisebox{0.40\resLen}{\rotatebox[origin=c]{90}{\scriptsize Ours (NR)}} &
        \multicolumn{2}{c}{\includegraphics[height=\resLen]{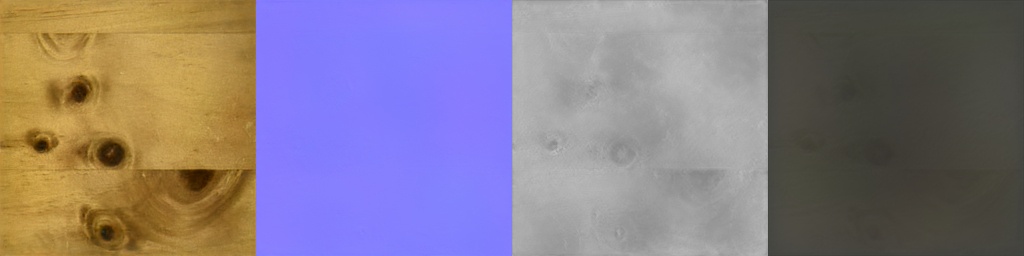}} &
        \includegraphics[height=\resLen]{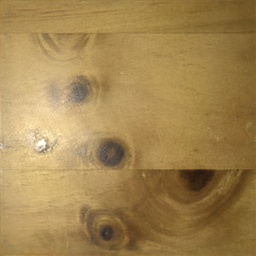} &
        \includegraphics[height=\resLen]{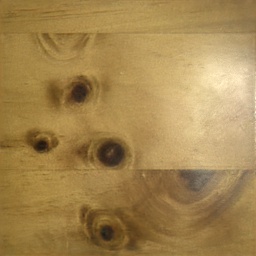} &
        \multicolumn{2}{c}{\includegraphics[height=\resLen]{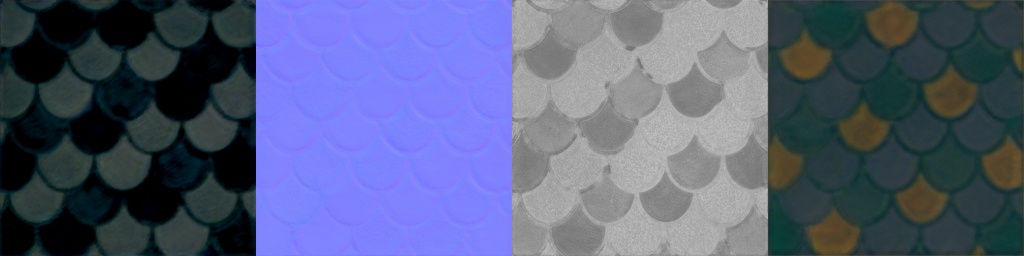}} &
        \includegraphics[height=\resLen]{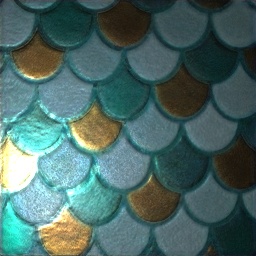} &
        \includegraphics[height=\resLen]{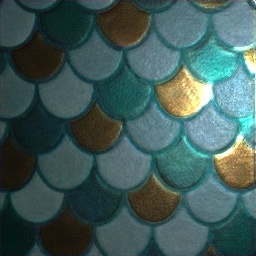}
        \\[-1pt]
        & &
        \multicolumn{2}{c}{\includegraphics[height=0.5\resLen]{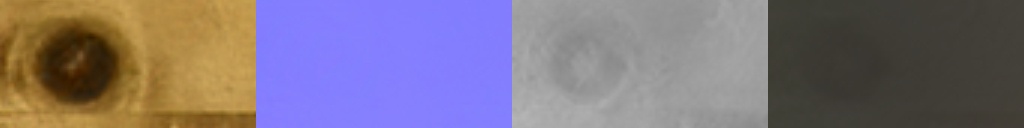}} &
        \includegraphics[height=0.5\resLen]{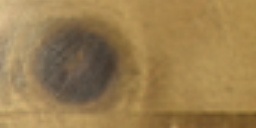} &
        \includegraphics[height=0.5\resLen]{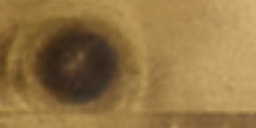} &
        \multicolumn{2}{c}{\includegraphics[height=0.5\resLen]{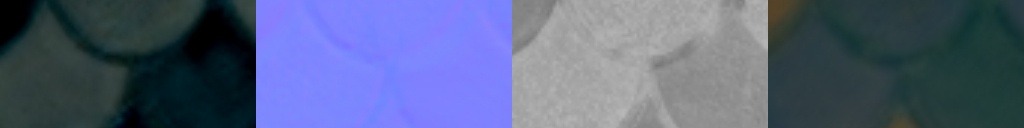}} &
        \includegraphics[height=0.5\resLen]{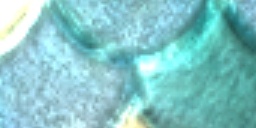} &
        \includegraphics[height=0.5\resLen]{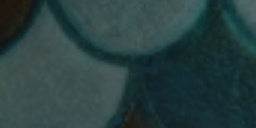}
        \\[2pt]
        & \raisebox{0.40\resLen}{\rotatebox[origin=c]{90}{\scriptsize [Gao19]+ (NR)}} &
        \multicolumn{2}{c}{\includegraphics[height=\resLen]{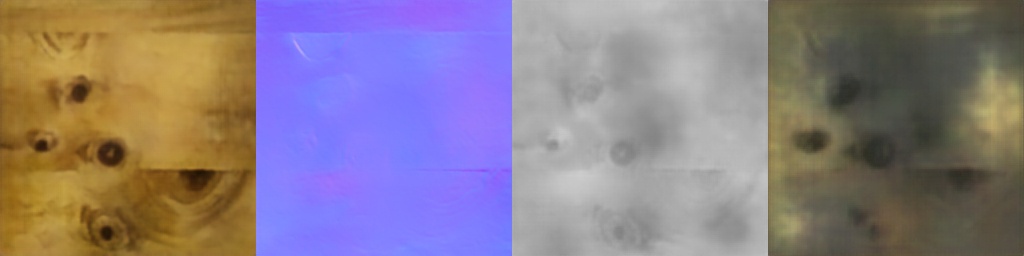}} &
        \includegraphics[height=\resLen]{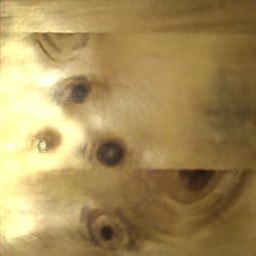} &
        \includegraphics[height=\resLen]{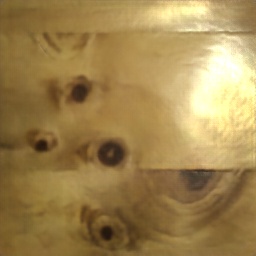} &
        \multicolumn{2}{c}{\includegraphics[height=\resLen]{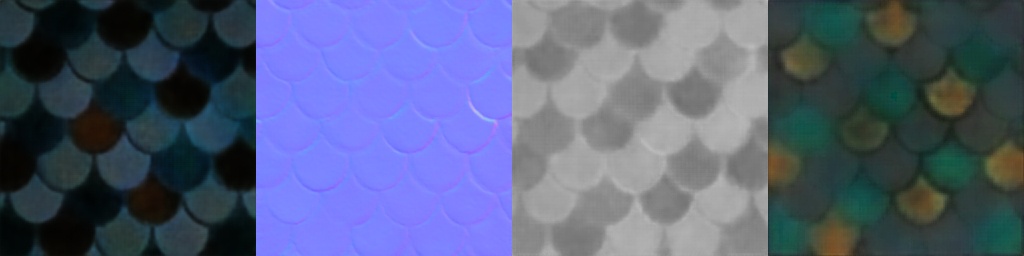}} &
        \includegraphics[height=\resLen]{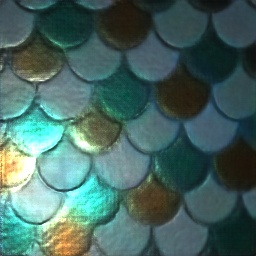} &
        \includegraphics[height=\resLen]{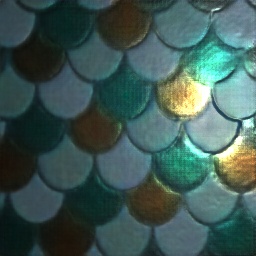}
        \\[-1pt]
        & &
        \multicolumn{2}{c}{\includegraphics[height=0.5\resLen]{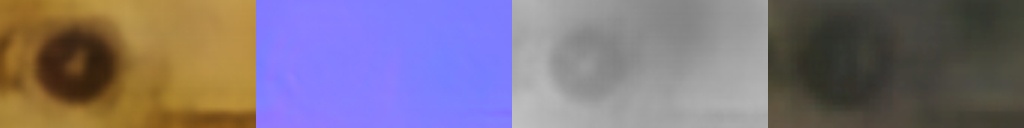}} &
        \includegraphics[height=0.5\resLen]{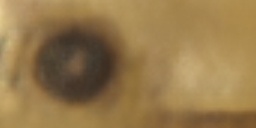} &
        \includegraphics[height=0.5\resLen]{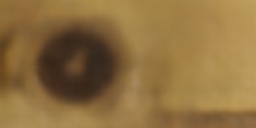} &
        \multicolumn{2}{c}{\includegraphics[height=0.5\resLen]{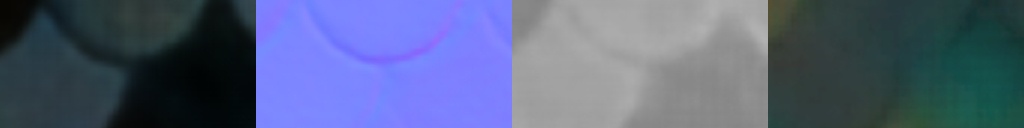}} &
        \includegraphics[height=0.5\resLen]{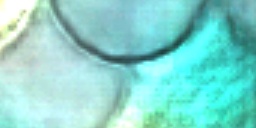} &
        \includegraphics[height=0.5\resLen]{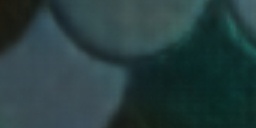}
        \\
        \hline\\[-8pt]
        \multirow{4}{*}[-1em]{\rotatebox[origin=c]{90}{\scriptsize\bfseries With refinement}} &
        \raisebox{0.40\resLen}{\rotatebox[origin=c]{90}{\scriptsize Ours}} &
		\multicolumn{2}{c}{\includegraphics[height=\resLen]{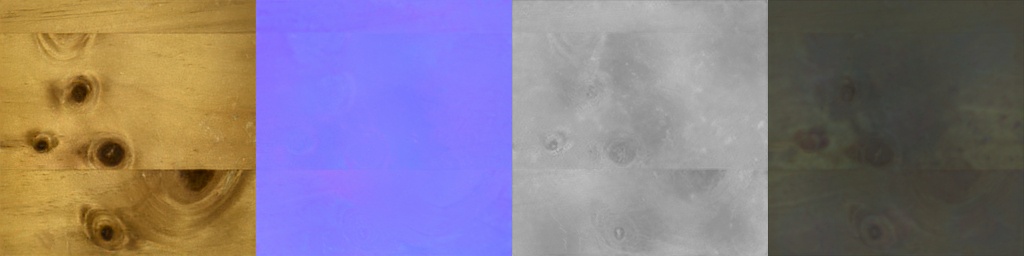}} &
		\includegraphics[height=\resLen]{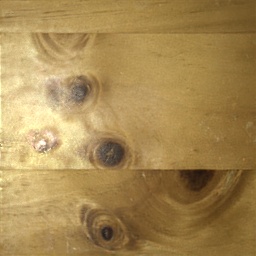} &
		\includegraphics[height=\resLen]{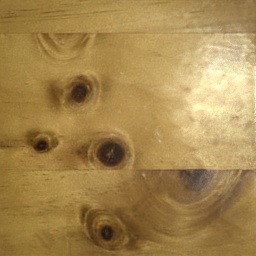} &
		\multicolumn{2}{c}{\includegraphics[height=\resLen]{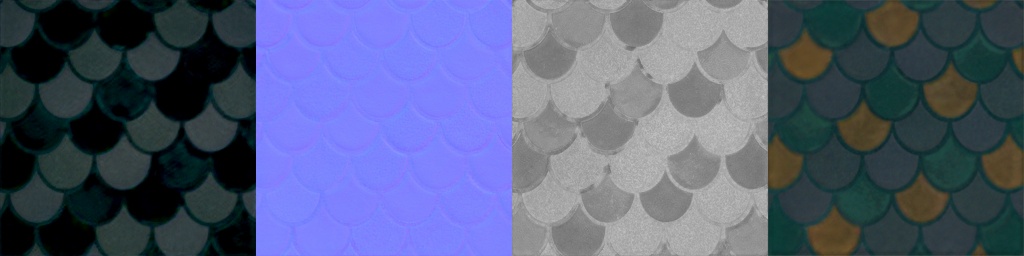}} &
		\includegraphics[height=\resLen]{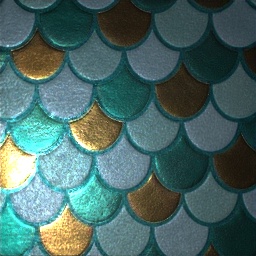} &
		\includegraphics[height=\resLen]{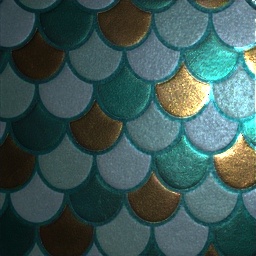}
		\\[-1pt]
		& &
		\multicolumn{2}{c}{\includegraphics[height=0.5\resLen]{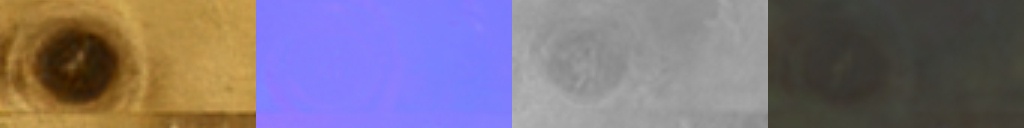}} &
		\includegraphics[height=0.5\resLen]{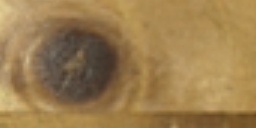} &
		\includegraphics[height=0.5\resLen]{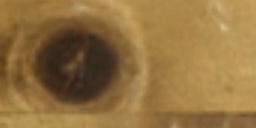} &
		\multicolumn{2}{c}{\includegraphics[height=0.5\resLen]{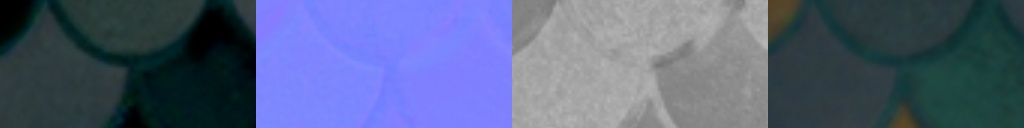}} &
		\includegraphics[height=0.5\resLen]{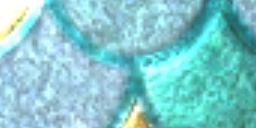} &
		\includegraphics[height=0.5\resLen]{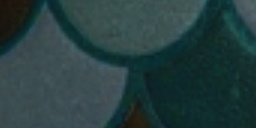}
		\\[2pt]
        & \raisebox{0.40\resLen}{\rotatebox[origin=c]{90}{\scriptsize [Gao19]+}} &
        \multicolumn{2}{c}{\includegraphics[height=\resLen]{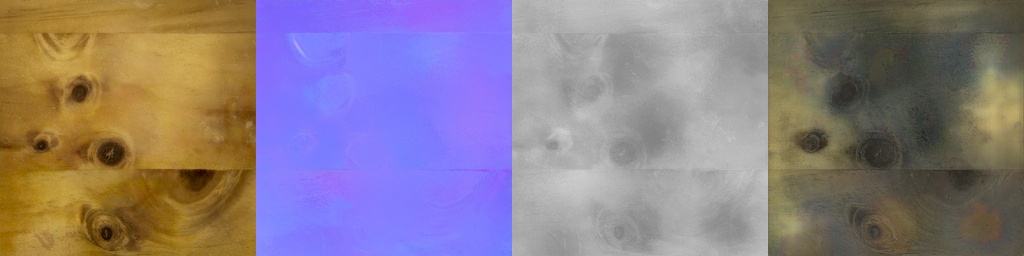}} &
        \includegraphics[height=\resLen]{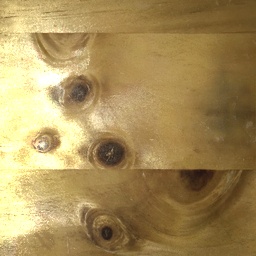} &
        \includegraphics[height=\resLen]{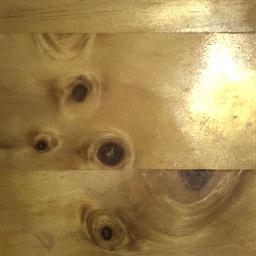} &
        \multicolumn{2}{c}{\includegraphics[height=\resLen]{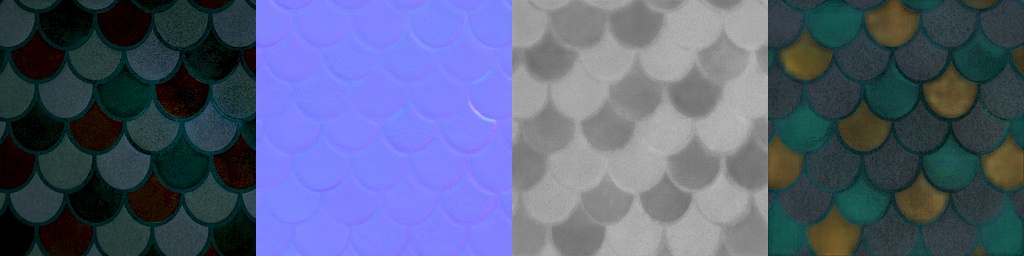}} &
        \includegraphics[height=\resLen]{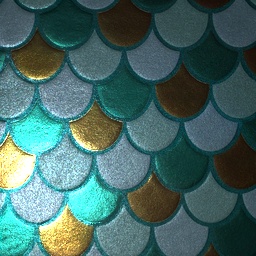} &
        \includegraphics[height=\resLen]{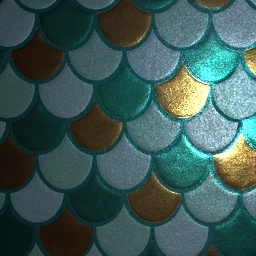}
        \\[-1pt]
        & &
        \multicolumn{2}{c}{\includegraphics[height=0.5\resLen]{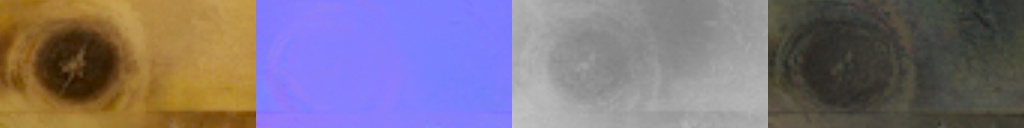}} &
        \includegraphics[height=0.5\resLen]{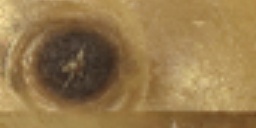} &
        \includegraphics[height=0.5\resLen]{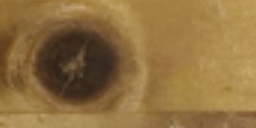} &
        \multicolumn{2}{c}{\includegraphics[height=0.5\resLen]{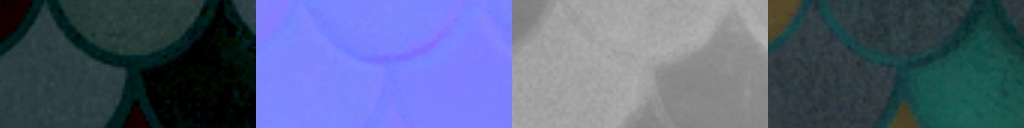}} &
        \includegraphics[height=0.5\resLen]{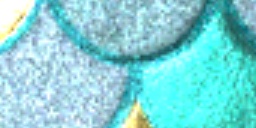} &
        \includegraphics[height=0.5\resLen]{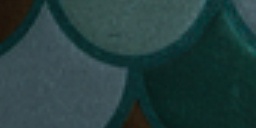}
    \end{tabular}
    \caption{\label{fig:refine}
        \textbf{Per-pixel post-refinement.} Unlike Gao's method, post-refinement via per-pixel optimization makes less of a difference in our method.
        Without post-refinement, [Gao19]+ (i.e., Gao's method initialized with Deschaintre's~\shortcite{Deschaintre2019} direct predictions) usually produces blurry results, as shown in the row marked as ``[Gao19]+ (NR)''.
        Our method, on the contrary, does not rely nearly as heavily on post-refinement: Without it, our results are already quite sharp (see ``Ours (NR)''), thanks to the generative power of our MaterialGAN.
        A zoomed-in version is attached below each SVBRDF map and novel-view image.
    }
\end{figure*}

\paragraph{Post-refinement.}
In general, the quality of our maps is sufficient after using our MaterialGAN-based optimization. However, Gao's method introduced a post-refinement step, where the maps are further optimized without any latent space, and with at most minor regularization. Therefore, we also implement a similar post-refinement step. However, like good initialization, this post-refinement makes less of a difference in our method, and Gao's method is more dependent on it, as it produces significantly blurry maps without it. This is shown in Figure \ref{fig:refine}; note the difference in sharpness of the maps.
\renewcommand{\one}{fake_010}
\renewcommand{\two}{fake_006}
\renewcommand{\thr}{fake_015}
\newcommand{\IDone}{27}
\newcommand{\IDtwo}{27}
\newcommand{\IDthr}{27}

\setlength{\resLen}{.545in}
\begin{figure*}[t]
	\addtolength{\tabcolsep}{-4pt}
	\begin{tabular}{ccccc@{\hspace{4\tabcolsep}}cccc@{\hspace{4\tabcolsep}}cccc}
		\raisebox{.31in}{\rotatebox[origin=c]{90}{\footnotesize{GT Novel}}} &
		\raisebox{0.1in}{\includegraphics[height=\resLen]{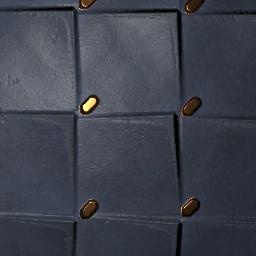}} &
		\multicolumn{3}{l}{\includegraphics[height=1.7\resLen]{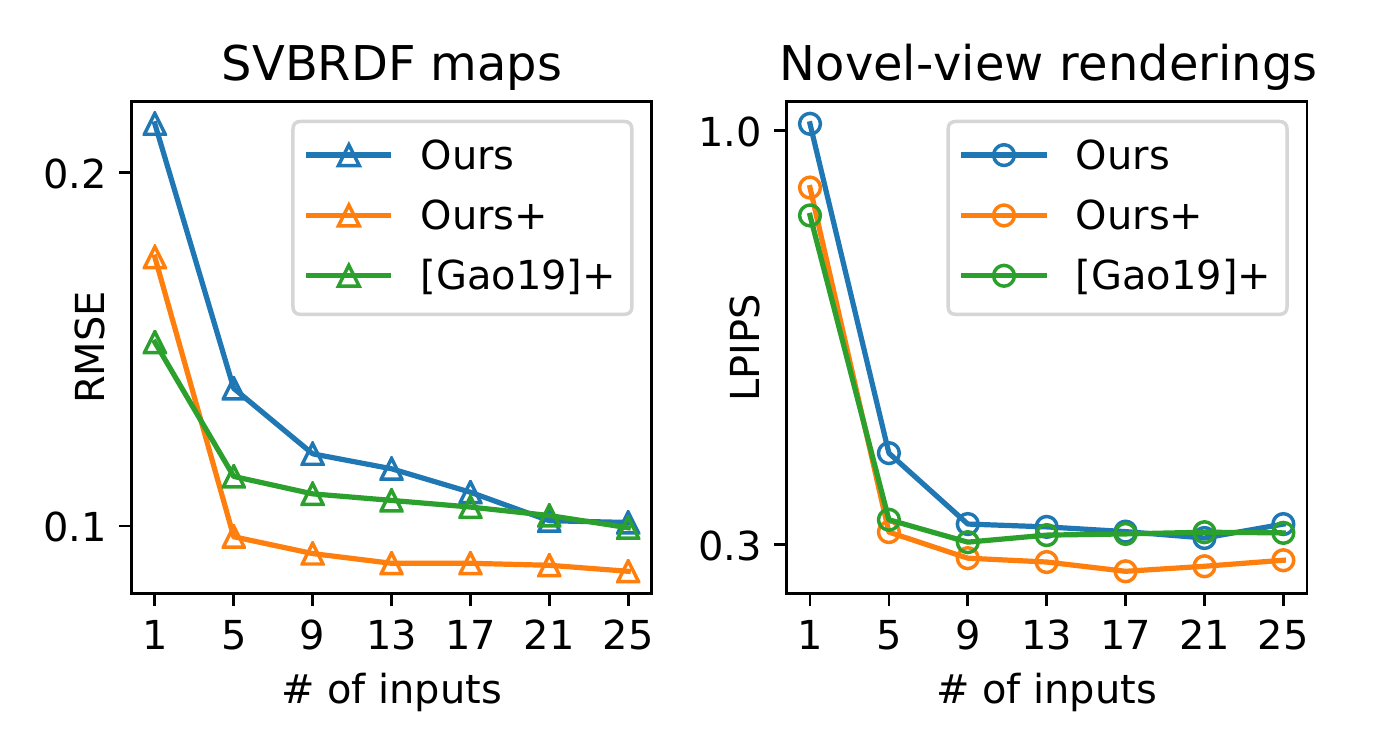}} &
		\raisebox{0.1in}{\includegraphics[height=\resLen]{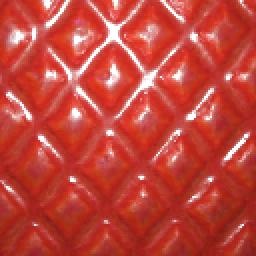}} &
		\multicolumn{3}{l}{\includegraphics[height=1.7\resLen]{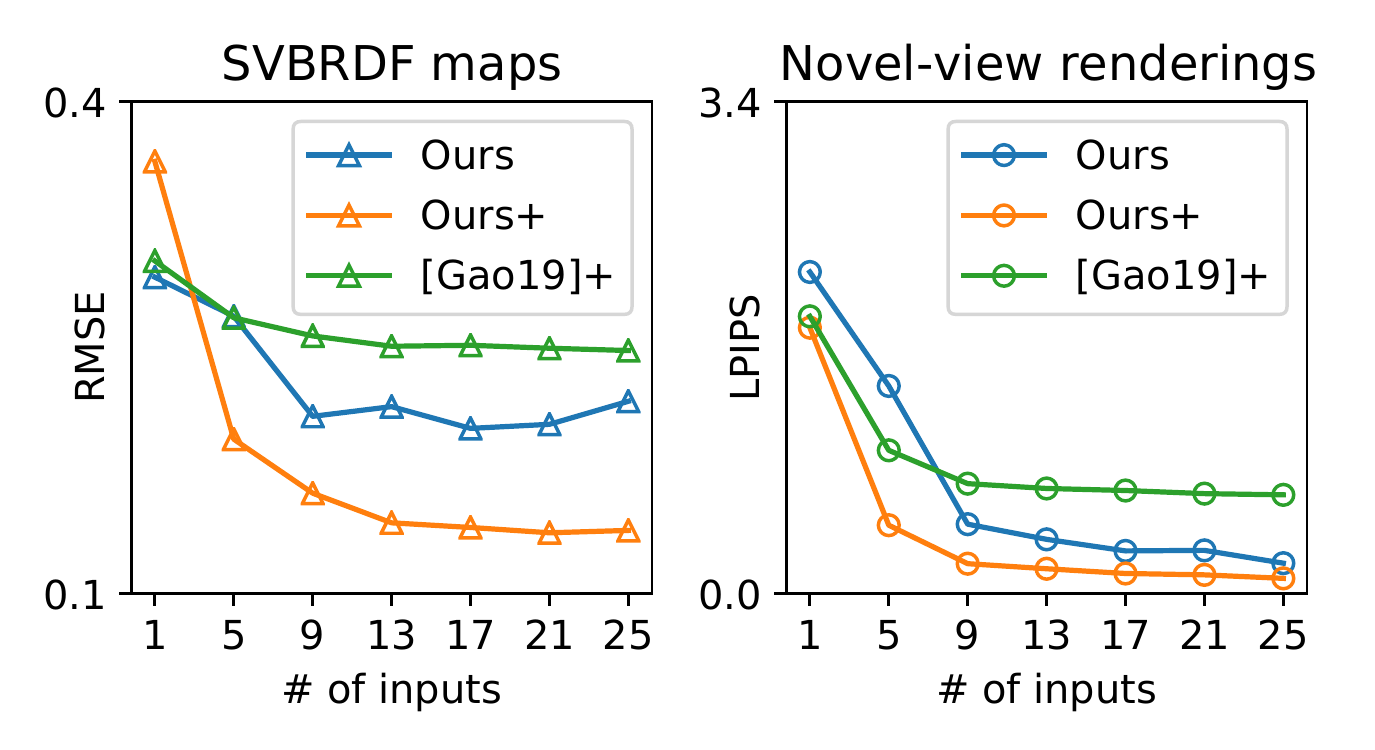}} &
		\raisebox{0.1in}{\includegraphics[height=\resLen]{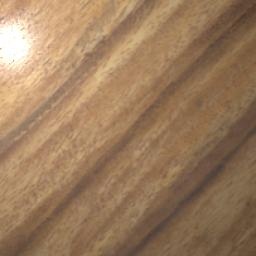}} &
		\multicolumn{3}{l}{\includegraphics[height=1.7\resLen]{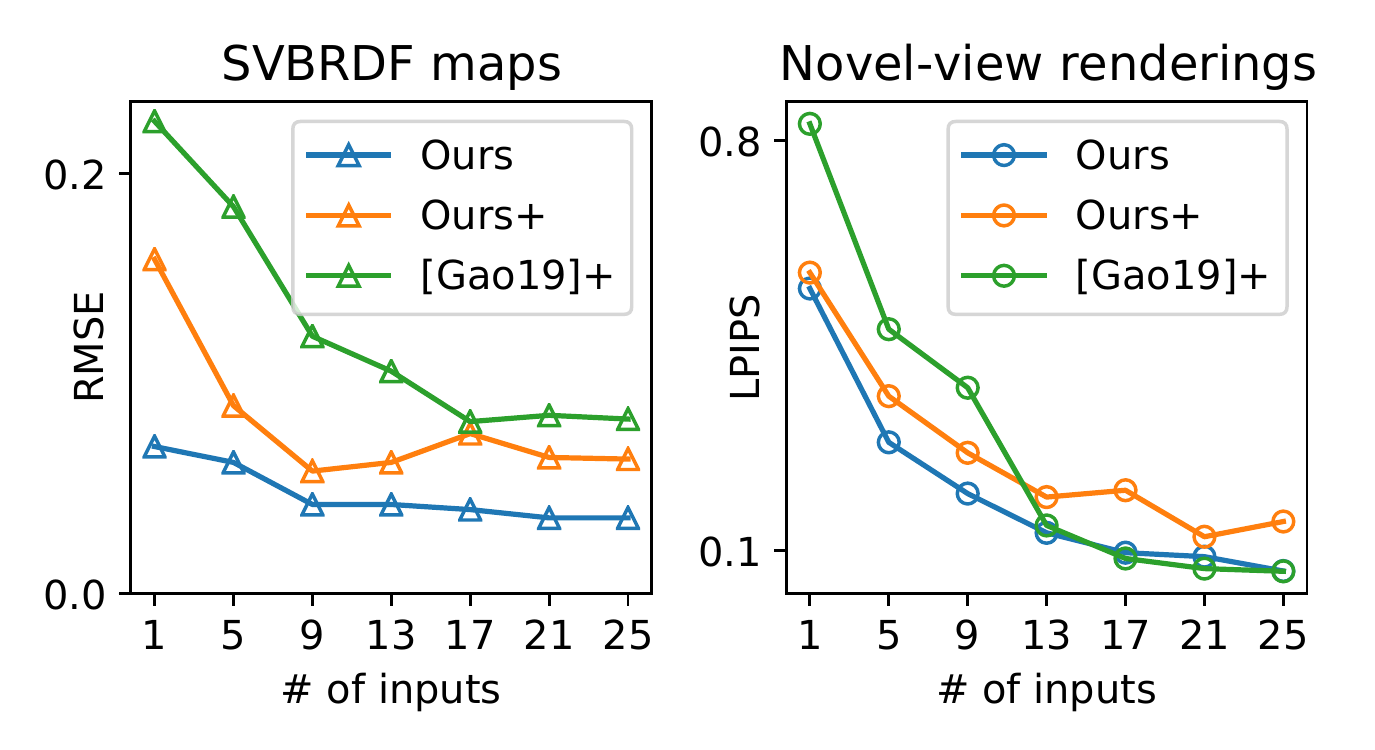}}
		\\
		& \textbf{\small N=1} & \textbf{\small 5} & \textbf{\small 9} & \textbf{\small 25}
		& \textbf{\small N=1} & \textbf{\small 5} & \textbf{\small 9} & \textbf{\small 25}
		& \textbf{\small N=1} & \textbf{\small 5} & \textbf{\small 9} & \textbf{\small 25}
		\\
		\raisebox{.25in}{\rotatebox[origin=c]{90}{\footnotesize{Ours}}} &
		\includegraphics[height=\resLen]{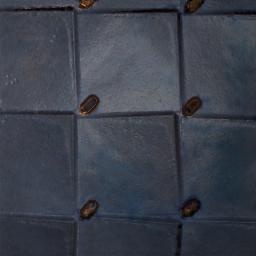} &
		\includegraphics[height=\resLen]{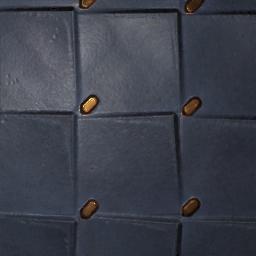} &
		\includegraphics[height=\resLen]{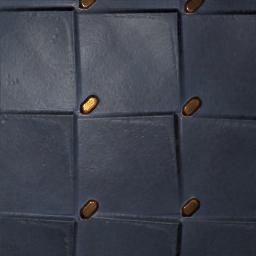} &
		\includegraphics[height=\resLen]{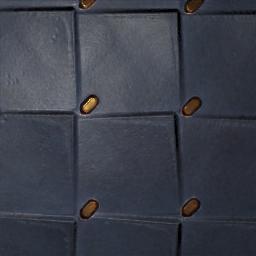} &
		\includegraphics[height=\resLen]{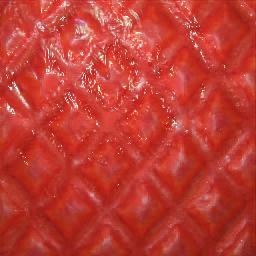} &
		\includegraphics[height=\resLen]{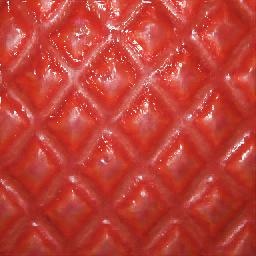} &
		\includegraphics[height=\resLen]{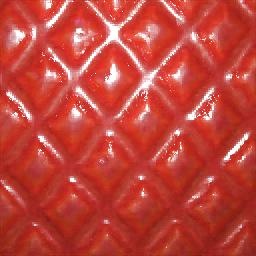} &
		\includegraphics[height=\resLen]{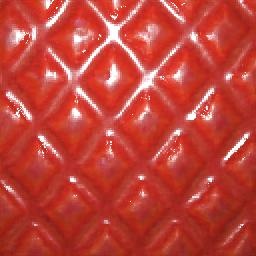} &
		\includegraphics[height=\resLen]{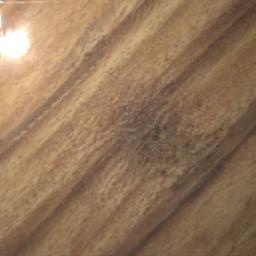} &
		\includegraphics[height=\resLen]{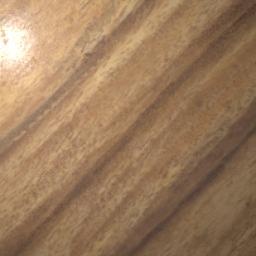} &
		\includegraphics[height=\resLen]{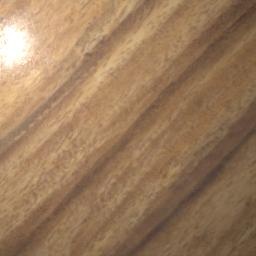} &
		\includegraphics[height=\resLen]{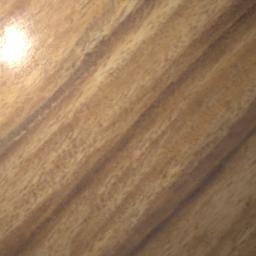}
		\\
		\raisebox{.25in}{\rotatebox[origin=c]{90}{\footnotesize{Ours+}}} &
		\includegraphics[height=\resLen]{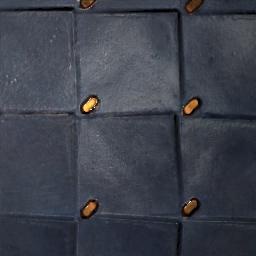} &
		\includegraphics[height=\resLen]{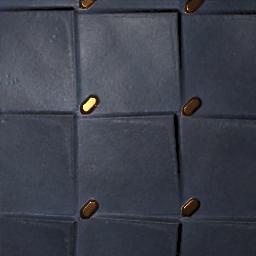} &
		\includegraphics[height=\resLen]{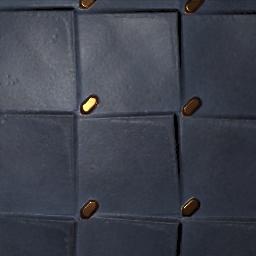} &
		\includegraphics[height=\resLen]{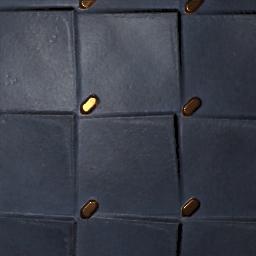} &
		\includegraphics[height=\resLen]{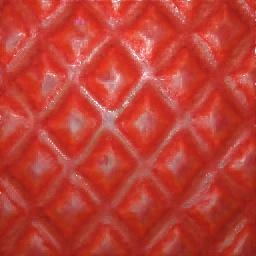} &
		\includegraphics[height=\resLen]{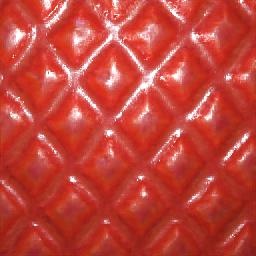} &
		\includegraphics[height=\resLen]{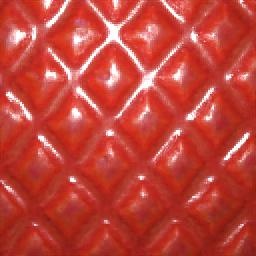} &
		\includegraphics[height=\resLen]{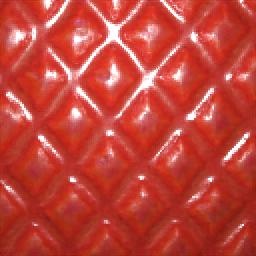} &
		\includegraphics[height=\resLen]{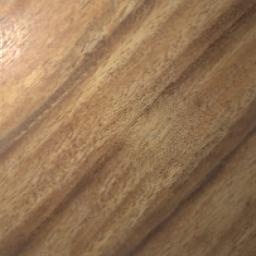} &
		\includegraphics[height=\resLen]{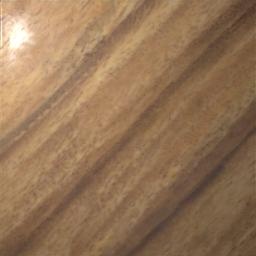} &
		\includegraphics[height=\resLen]{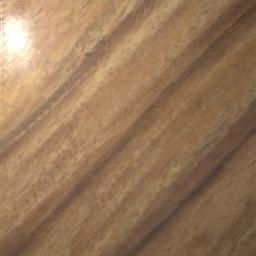} &
		\includegraphics[height=\resLen]{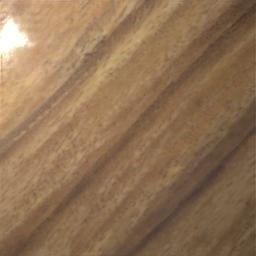}
		\\
		\raisebox{.25in}{\rotatebox[origin=c]{90}{\footnotesize{[Gao19]+}}} &
		\includegraphics[height=\resLen]{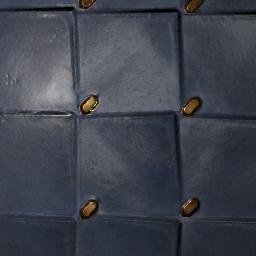} &
		\includegraphics[height=\resLen]{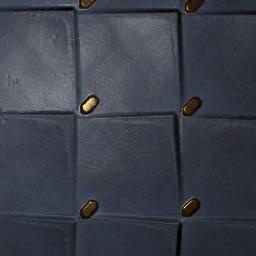} &
		\includegraphics[height=\resLen]{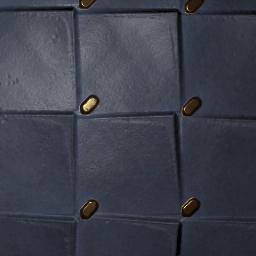} &
		\includegraphics[height=\resLen]{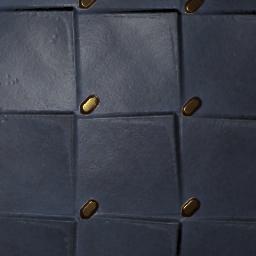} &
		\includegraphics[height=\resLen]{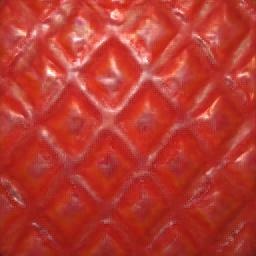} &
		\includegraphics[height=\resLen]{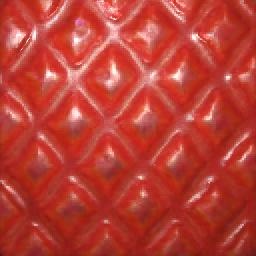} &
		\includegraphics[height=\resLen]{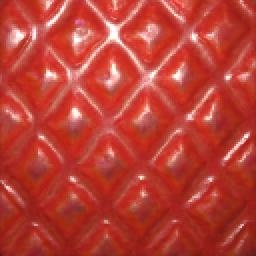} &
		\includegraphics[height=\resLen]{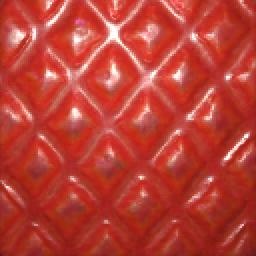} &
		\includegraphics[height=\resLen]{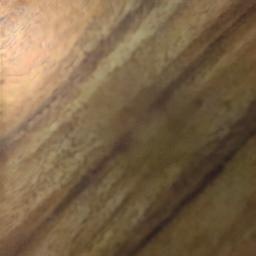} &
		\includegraphics[height=\resLen]{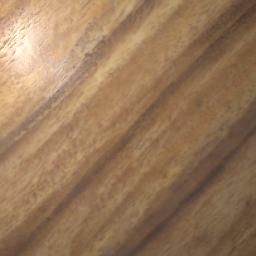} &
		\includegraphics[height=\resLen]{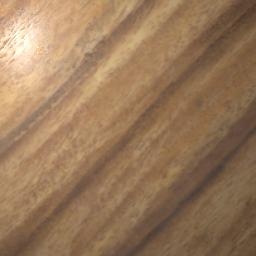} &
		\includegraphics[height=\resLen]{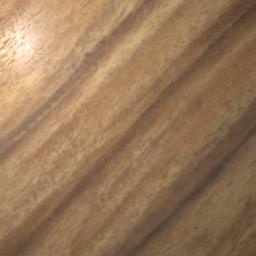}
	\end{tabular}
	\caption{\label{fig:results_multi_inputs2}
		\revision{\textbf{Performance using different numbers of input images (synthetic data).} 
			The quality of recovered SVBRDF maps, as demonstrated by the plots, generally improves with more input images for both our and Gao's~\shortcite{Gao2019} methods.
			Our method with constant (Ours) and neural (Ours+) initializations are comparable or better than Gao's  ([Gao19]+) with neural initialization~\cite{Deschaintre2019} .
			For a highly specular material shown on the right, although the LPIPS metric computed using renderings under 5 novel views of our results is similar to that of Gao's, ours better preserve the specular highlight.
			For each material, all the renderings including the references (GT Novel) are generated using one of the 5 novel views.
		}
	}
\end{figure*}
\renewcommand{\one}{real_plastic-red-carton}
\renewcommand{\two}{real_cards-red}

\setlength{\raiseLen}{0.23in}
\setlength{\resLen}{0.52in}
\begin{figure*}[t]
	\addtolength{\tabcolsep}{-4pt}
	\begin{tabular}{cccc@{\hspace{4\tabcolsep}}ccc}
		& \textbf{\small SVBRDF maps} &
		\multicolumn{2}{c}{\textbf{\small Novel views}} &
		\textbf{\small SVBRDF maps} &
		\multicolumn{2}{c}{\textbf{\small Novel views}}
		\\
		\raisebox{\raiseLen}{\rotatebox[origin=c]{90}{\footnotesize{Ours (1)}}} &
		\includegraphics[height=\resLen]{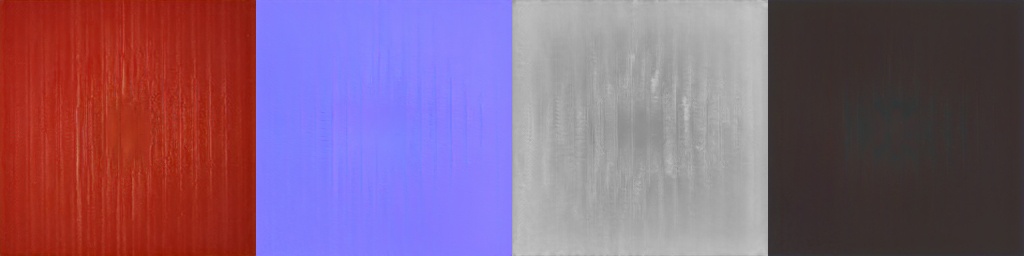} &
		\includegraphics[height=\resLen]{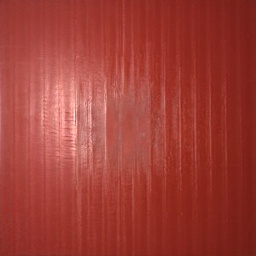} &
		\includegraphics[height=\resLen]{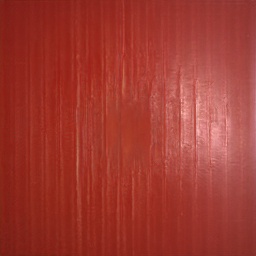} &
		\includegraphics[height=\resLen]{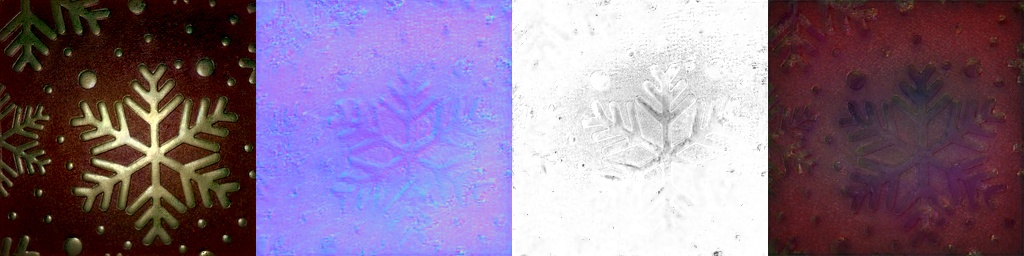} &
		\includegraphics[height=\resLen]{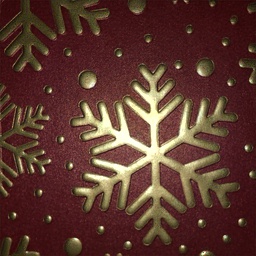} &
		\includegraphics[height=\resLen]{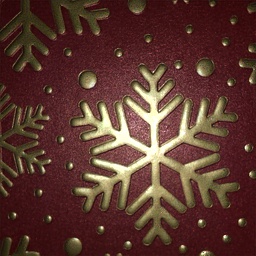}
		\\
		\raisebox{\raiseLen}{\rotatebox[origin=c]{90}{\footnotesize{Ours (3)}}} &
		\includegraphics[height=\resLen]{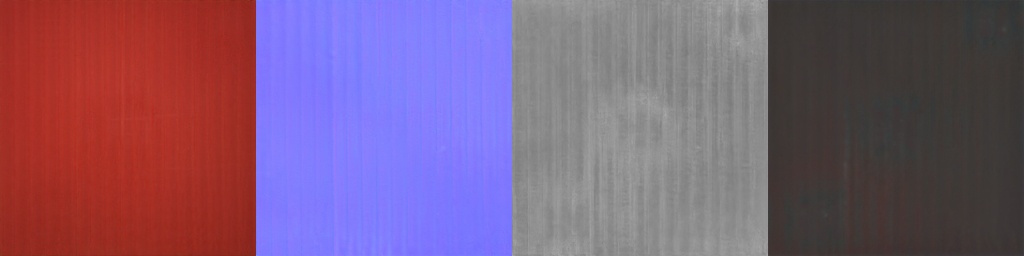} &
		\includegraphics[height=\resLen]{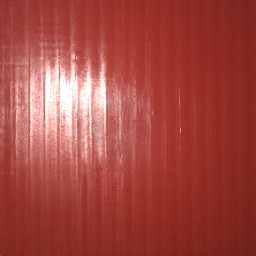} &
		\includegraphics[height=\resLen]{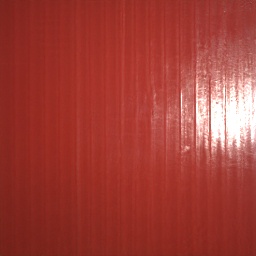} &
		\includegraphics[height=\resLen]{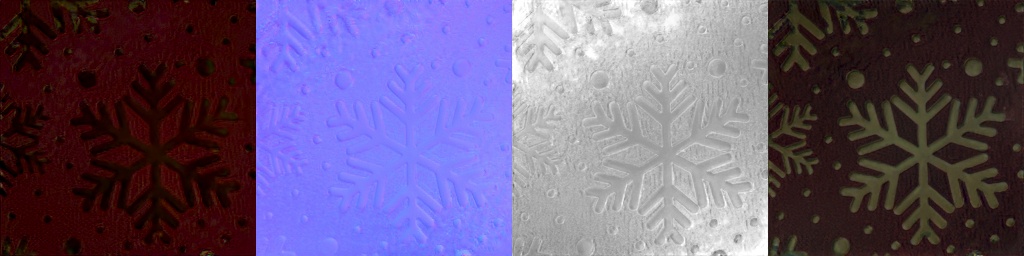} &
		\includegraphics[height=\resLen]{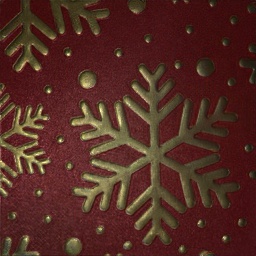} &
		\includegraphics[height=\resLen]{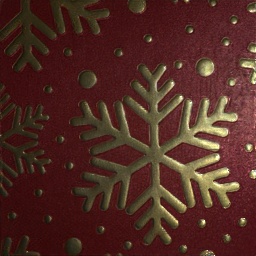}
		\\
		\raisebox{\raiseLen}{\rotatebox[origin=c]{90}{\footnotesize{Ours (7)}}} &
		\includegraphics[height=\resLen]{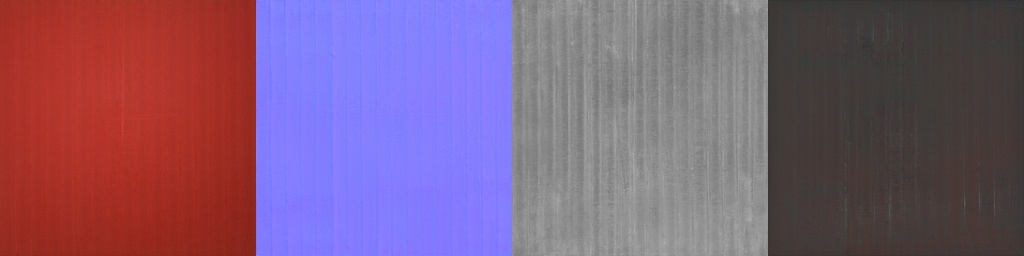} &
		\includegraphics[height=\resLen]{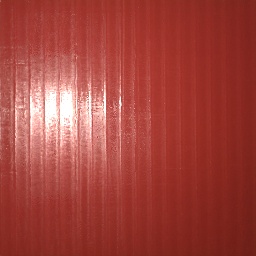} &
		\includegraphics[height=\resLen]{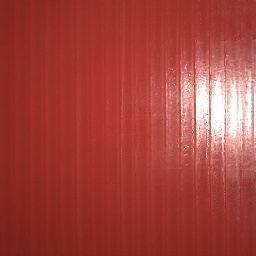} &
		\includegraphics[height=\resLen]{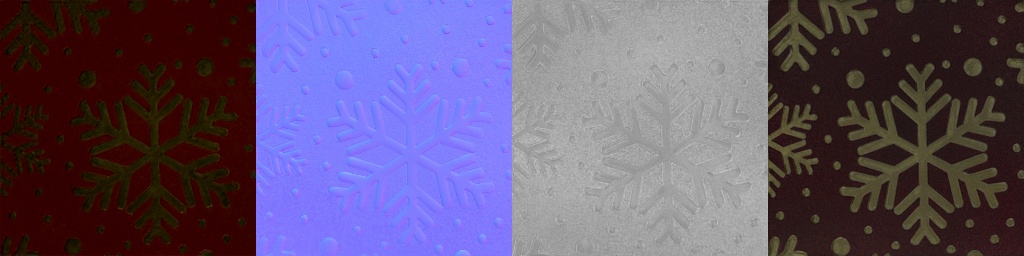} &
		\includegraphics[height=\resLen]{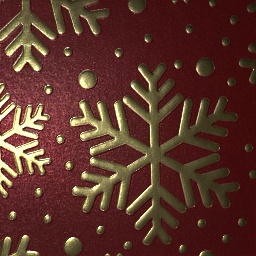} &
		\includegraphics[height=\resLen]{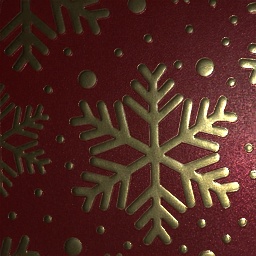}
		\\
		\hline\\[-8pt]
		\raisebox{\raiseLen}{\rotatebox[origin=c]{90}{\footnotesize{Ours+ (1)}}} &
		\includegraphics[height=\resLen]{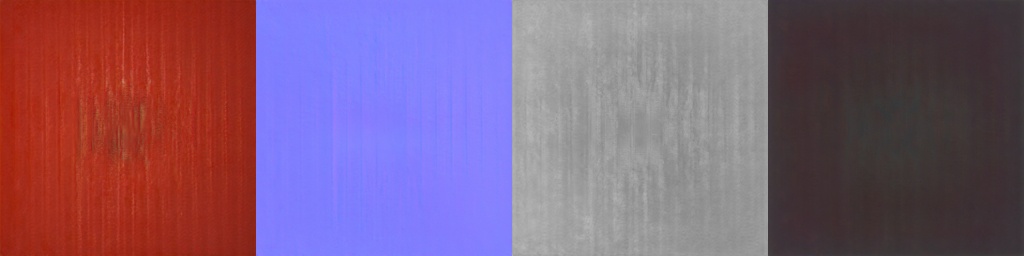} &
		\includegraphics[height=\resLen]{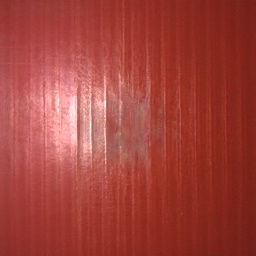} &
		\includegraphics[height=\resLen]{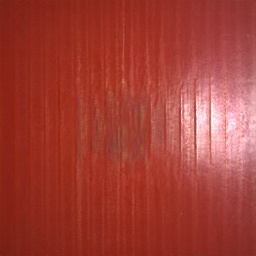} &
		\includegraphics[height=\resLen]{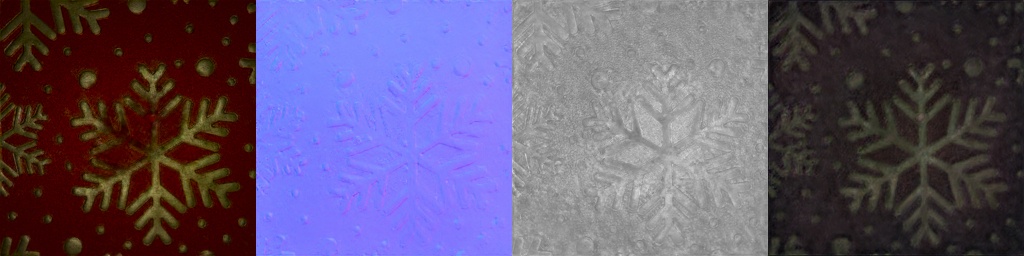} &
		\includegraphics[height=\resLen]{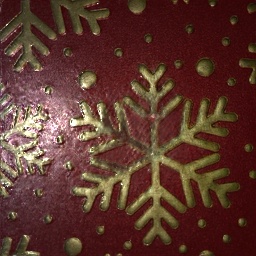} &
		\includegraphics[height=\resLen]{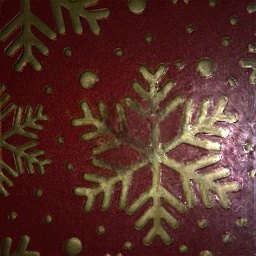}
		\\
		\raisebox{\raiseLen}{\rotatebox[origin=c]{90}{\footnotesize{Ours+ (3)}}} &
		\includegraphics[height=\resLen]{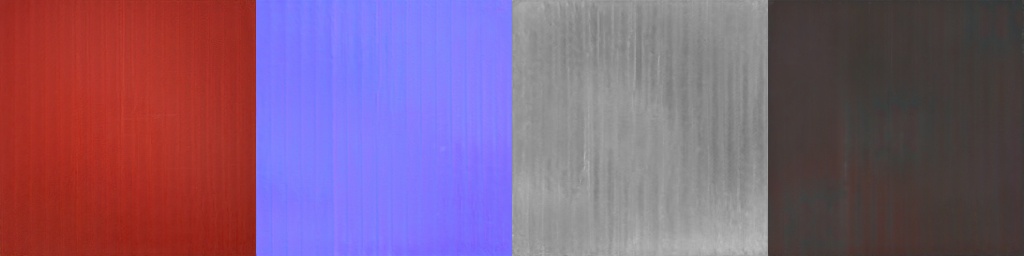} &
		\includegraphics[height=\resLen]{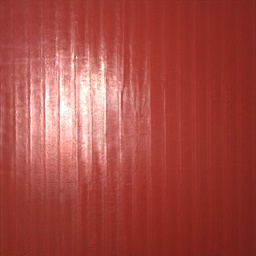} &
		\includegraphics[height=\resLen]{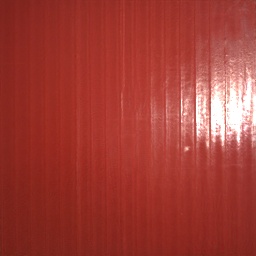} &
		\includegraphics[height=\resLen]{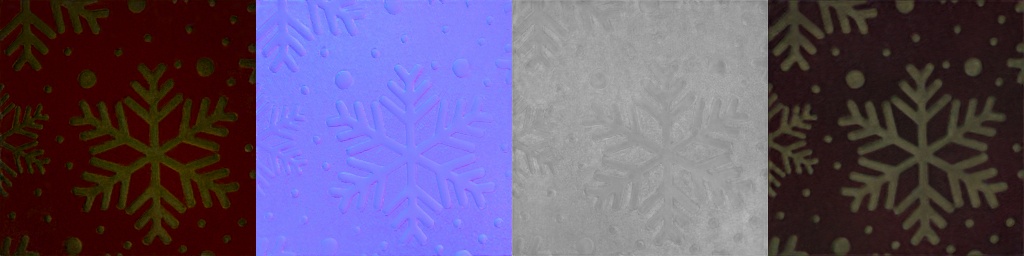} &
		\includegraphics[height=\resLen]{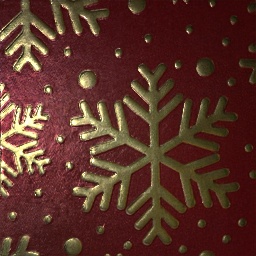} &
		\includegraphics[height=\resLen]{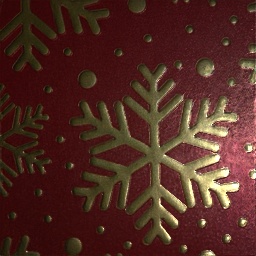}
		\\
		\raisebox{\raiseLen}{\rotatebox[origin=c]{90}{\footnotesize{Ours+ (7)}}} &
		\includegraphics[height=\resLen]{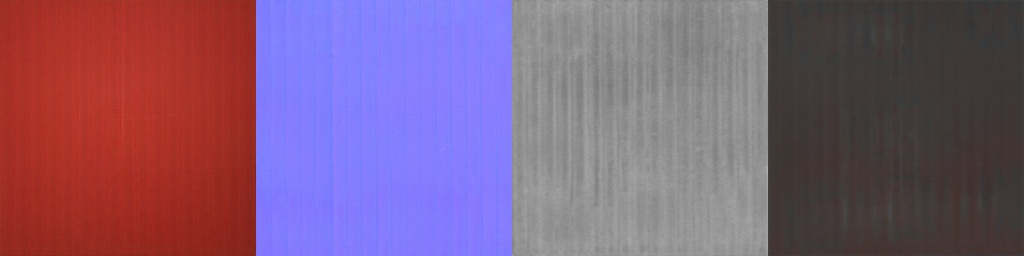} &
		\includegraphics[height=\resLen]{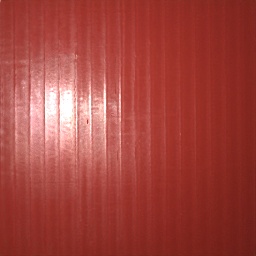} &
		\includegraphics[height=\resLen]{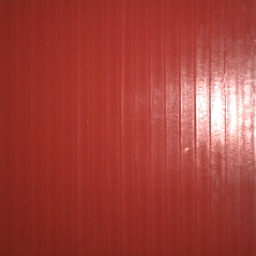} &
		\includegraphics[height=\resLen]{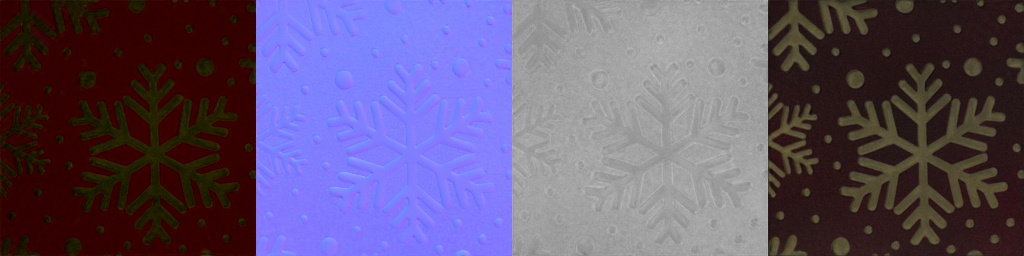} &
		\includegraphics[height=\resLen]{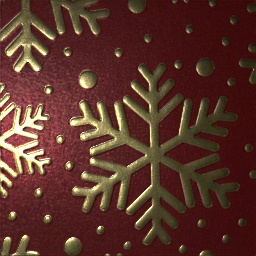} &
		\includegraphics[height=\resLen]{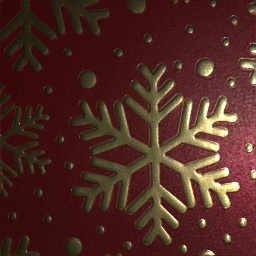}
		\\
		\hline\\[-8pt]
		\raisebox{\raiseLen}{\rotatebox[origin=c]{90}{\footnotesize{[Gao19]+ (1)}}} &
		\includegraphics[height=\resLen]{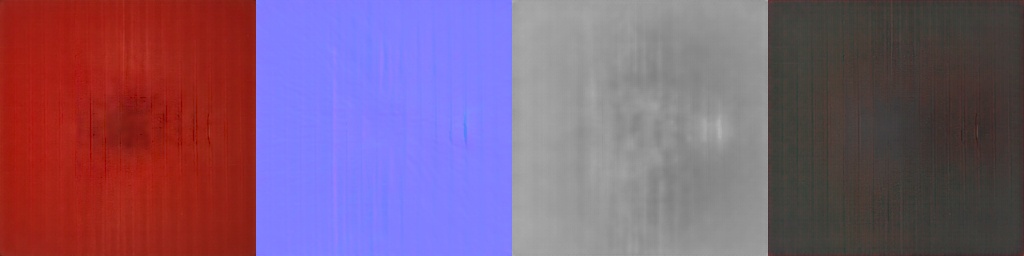} &
		\includegraphics[height=\resLen]{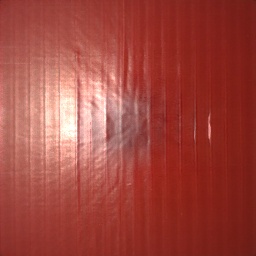} &
		\includegraphics[height=\resLen]{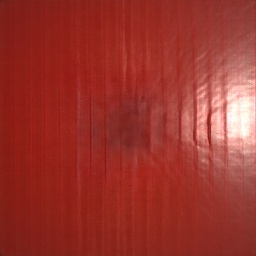} &
		\includegraphics[height=\resLen]{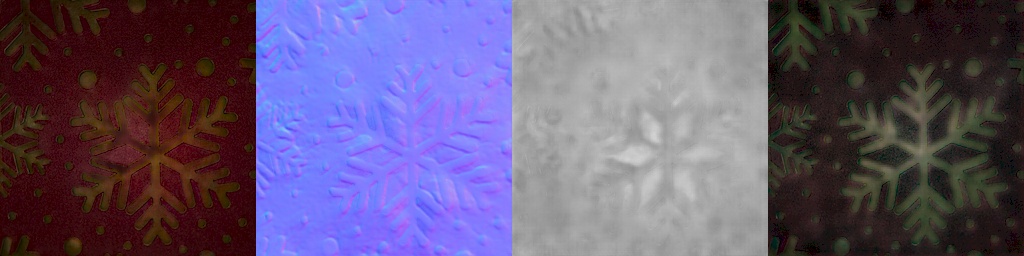} &
		\includegraphics[height=\resLen]{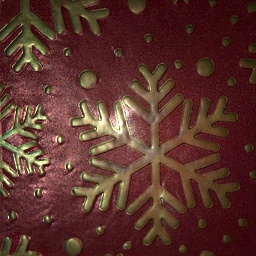} &
		\includegraphics[height=\resLen]{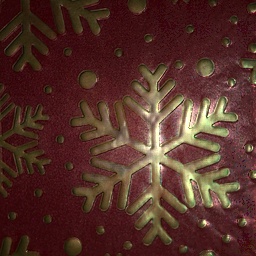}
		\\
		\raisebox{\raiseLen}{\rotatebox[origin=c]{90}{\footnotesize{[Gao19]+ (3)}}} &
		\includegraphics[height=\resLen]{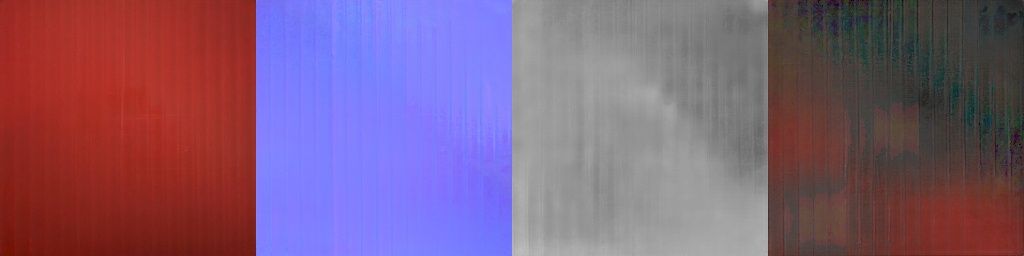} &
		\includegraphics[height=\resLen]{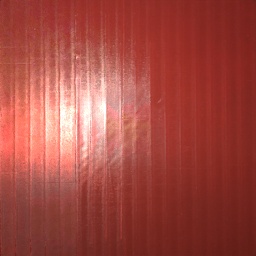} &
		\includegraphics[height=\resLen]{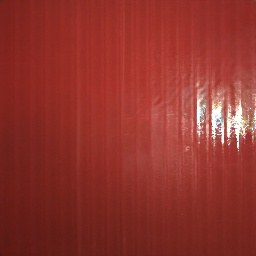} &
		\includegraphics[height=\resLen]{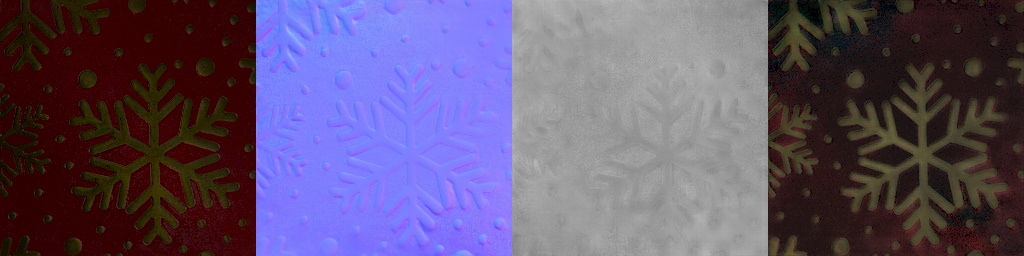} &
		\includegraphics[height=\resLen]{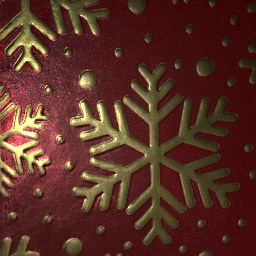} &
		\includegraphics[height=\resLen]{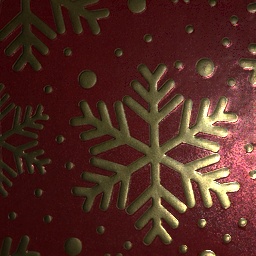}
		\\
		\raisebox{\raiseLen}{\rotatebox[origin=c]{90}{\footnotesize{[Gao19]+ (7)}}} &
		\includegraphics[height=\resLen]{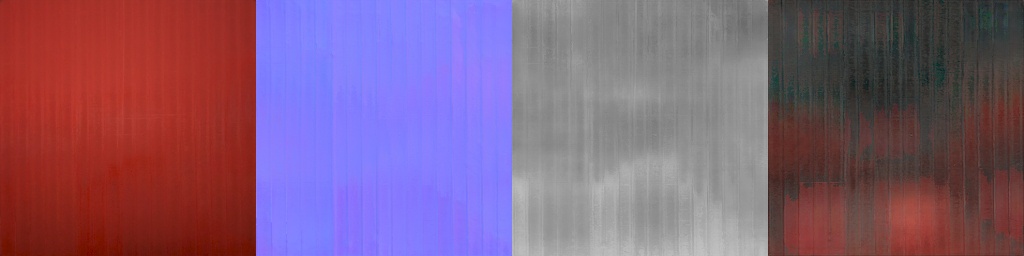} &
		\includegraphics[height=\resLen]{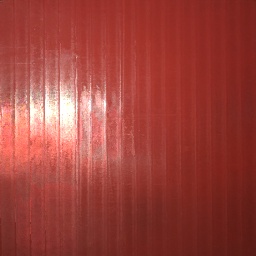} &
		\includegraphics[height=\resLen]{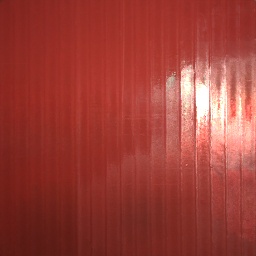} &
		\includegraphics[height=\resLen]{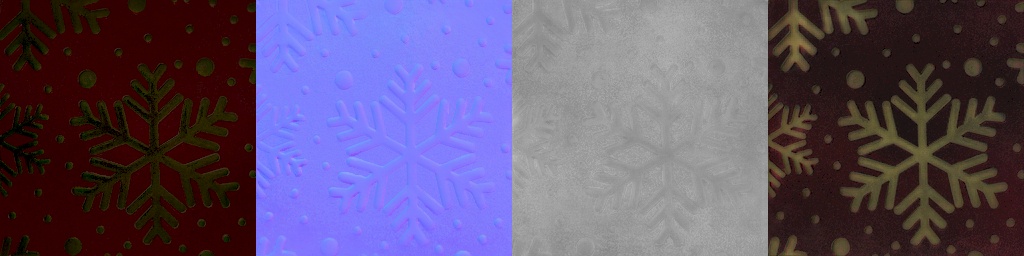} &
		\includegraphics[height=\resLen]{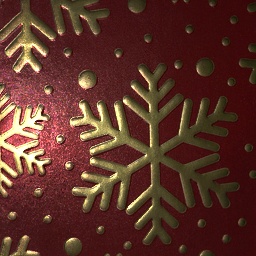} &
		\includegraphics[height=\resLen]{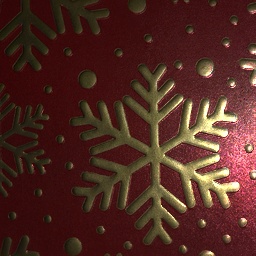}
	\end{tabular}
	\caption{\label{fig:results_multi_inputs}
		\textbf{Performance using different numbers of input images (real data).}
		\revision{The quality of SVBRDF maps recovered by our method generally improves with more input images under both constant initialization (see ``Ours'') and Deschaintre~\shortcite{Deschaintre2019} initialization (see ``Ours+'').}
	}
\end{figure*}
\paragraph{Optimization with different numbers of input images.} While most of our results are shown with 7 inputs, using two additional inputs for novel lighting evaluation, our method does work with various numbers of input images. We show 3 synthetic examples in Figure~\ref{fig:results_multi_inputs2}, with different numbers of inputs from 1 to 25. All the three examples are the same as used in Gao's work. The errors of both reconstructed SVBRDF maps and novel-view renderings generally decrease with more input images, as is expected for an inverse-rendering method. In Figure~\ref{fig:results_multi_inputs}, we compare real capture results with 1, 3, and 7 inputs, with and without initialization by Deschaintre's method, and also include Gao's results for 3 and 7 inputs (with initialization). Our result remains plausible with 3 inputs, though artifacts do get reduced with more inputs. For all numbers of inputs, our result (with or without initialization) tends to be cleaner than Gao's.
\paragraph{Editing operations.} An additional advantage of the StyleGAN-based latent space is the ability to achieve semantically meaningful operations such as morphing, by interpolating two or more parent latent codes to create a hybrid offspring material. Morphing in latent space often preserves semantic features qualitatively better than naive interpolation in pixel space. Figure~\ref{fig:interp}~\revision{and the supplemental video show} morphing of a few real materials using linear interpolation in latent space, compared to the corresponding naive interpolation (linear in pixel space).
\setlength{\resLen}{.96\columnwidth}
\begin{figure*}[t]
	\centering
	\addtolength{\tabcolsep}{-3pt}
	\begin{tabular}{cc}
		{\small \textbf{GAN-based} interpolation of SVBRDF maps} & {\small \textbf{Linear interpolation} of SVBRDF maps}\\
		\includegraphics[width=\resLen]{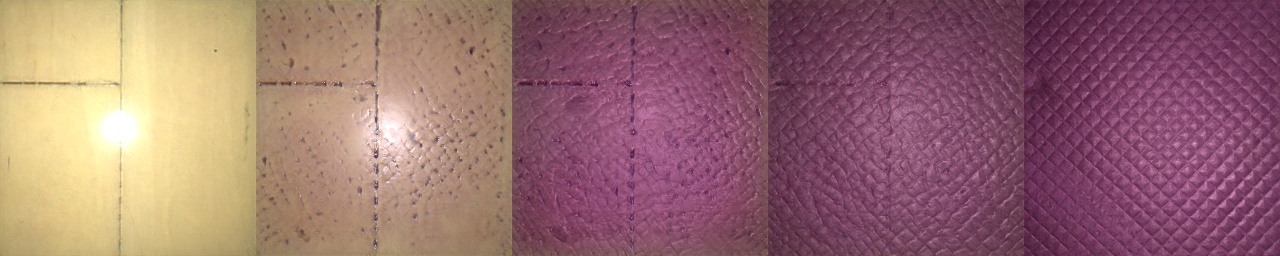} &
		\includegraphics[width=\resLen]{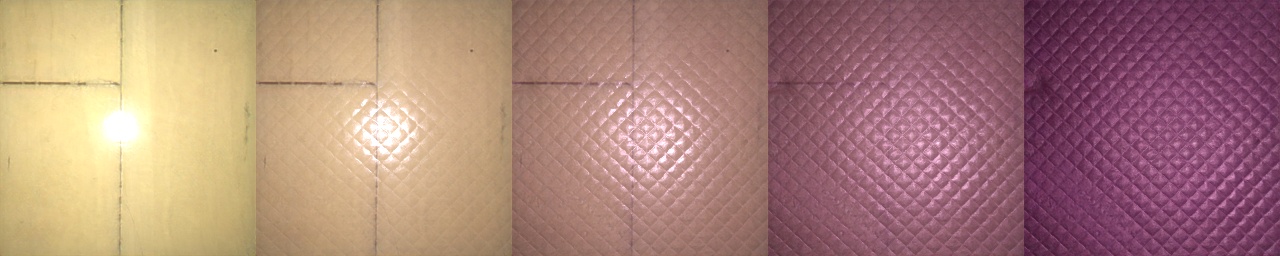}\\
		\includegraphics[width=\resLen]{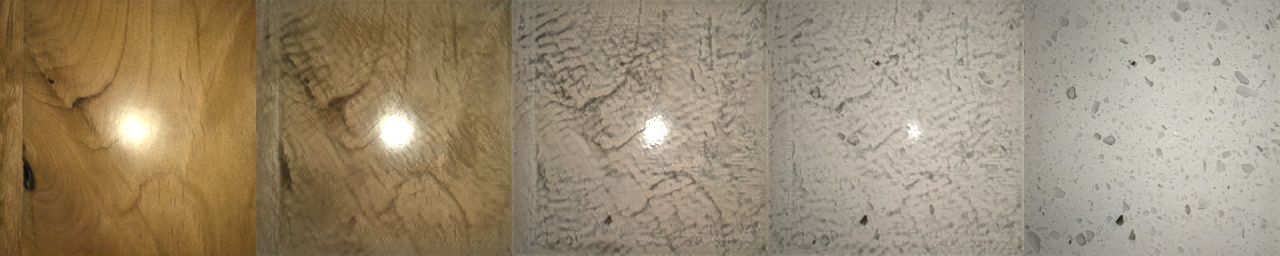} &
		\includegraphics[width=\resLen]{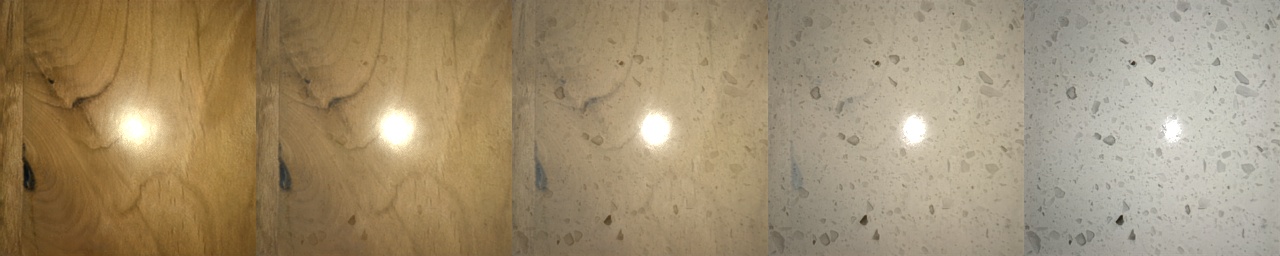}\\
		\includegraphics[width=\resLen]{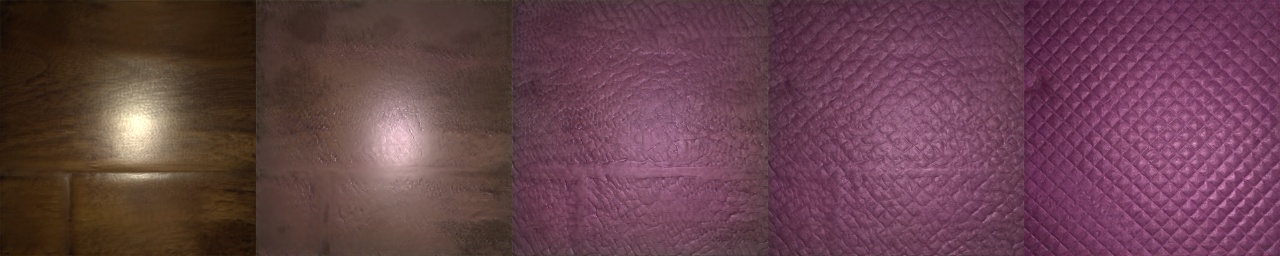} &
		\includegraphics[width=\resLen]{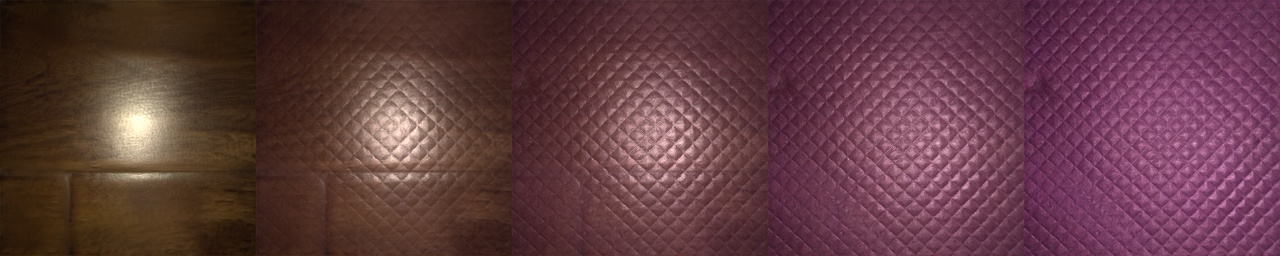}
	\end{tabular}
	\caption{\label{fig:interp}
		\textbf{Material interpolation.} Renderings of interpolations between two SVBRDFs recovered from real images using our method. Results on the left and right columns are obtained, respectively, using our GAN latent space and na\"ive linear interpolation.
	}
\end{figure*}

    \section{Conclusion and Future Work}
\label{sec:conclusion}
\paragraph{Discussion and limitations.}
While we believe our framework improves upon the state of the art, there are also some limitations. Our current BRDF model is shared by previous work, but certain common effects (layering on book covers, subsurface fiber scattering in woods, anisotropy in fabrics) are not correctly captured by it. An extension of our generative model and rendering operator would be possible, though the key challenge is finding high-quality  training data for these effects.

Our assumption of almost flat samples will fail for materials with strong relief patterns, and will produce blurring or ghosting if there are obvious parallax effects in the aligned captured images. Strong self-shadowing or inter-reflections are also not currently handled. Solving for height instead of normal, with a more advanced rendering operator, may be able to resolve parallax effects \revision{and to} correctly predict (and undo) shadowing effects from strong height variations.

Furthermore, more precise calibration may improve our accuracycd. This would likely require knowledge of the cell phone hardware, and/or pre-calibration of its properties (e.g. flash light falloff, lens vignetting, and color processing properties).


\paragraph{Conclusion.}
We propose a novel method for acquiring SVBRDFs from a small number of input images, typically 3 to 7, captured using a hand-held mobile phone.
We use an optimization framework that leverages a powerful material prior, based on a generative network, MaterialGAN, trained to synthesize plausible SVBRDFs.
MaterialGAN learns correlations in SVBRDF parameters and provides local and global regularization to our optimization.
This produces high-quality SVBRDFs that accurately reconstruct the input images, and because of our MaterialGAN prior, lie on a plausible material manifold.
As a result, our reconstructions generalize better to novel views and lighting than previous state-of-the-art methods.

We believe that our work is only a first step toward GAN-based material analysis and synthesis and our experiments suggest many avenues for further exploration including improving material latent spaces and optimization techniques using novel architectures and losses, learning disentangled and editable latent spaces, and expanding beyond our current isotropic BRDF model.

    \begin{acks}
	This research was started during Yu Guo's internship at Adobe Research. We thank TJ Rhodes for help with material capture hardware setup. This work was supported in part by NSF IIS-1813553.
\end{acks}

    \bibliographystyle{ACM-Reference-Format}
    \bibliography{references}
\end{document}